\documentclass[11pt]{article}
\usepackage{setspace,epsfig,amsmath,amssymb,color,stmaryrd,nicefrac}
\usepackage[round]{natbib}
\usepackage{subfig}

\newcommand*{\defeq}{\mathrel{\vcenter{\baselineskip0.5ex \lineskiplimit0pt
			\hbox{\footnotesize.}\hbox{\footnotesize.}}}%
	=}
\renewcommand{\vec}[1]{\boldsymbol{#1}}
\newcommand{\cX}{\mathcal{X}}
\newcommand{\cY}{\mathcal{Y}}
\newcommand{\cH}{\mathcal{H}}
\newcommand{\cD}{\mathcal{D}}

\newcommand{\hath}{\hat{h}}

\newcommand{\haty}{\hat{y}}
\newcommand{\sety}{\widehat{Y}}

\newcommand{\argmin}{\operatorname*{argmin}}
\newcommand{\argmax}{\operatorname*{argmax}}

\newcommand{\with}{\,  | \,}
\newcommand{\given}{\, | \,}

\newcommand{\set}[1]{\mathcal{#1}}
\newcommand{\Prob}{P}
\newcommand{\prob}{p}

\newcommand{\on}{\operatorname}

\newcommand{\fromto}{\longrightarrow}

\newcommand{\evalue}{\mathbf{E}}
\newcommand{\variance}{\mathbf{V}}

\newcommand{\mmp}[1]{}

\begin{document}

\title{\bf\LARGE Aleatoric and Epistemic Uncertainty in\\ Machine Learning: An Introduction\\ to Concepts and Methods}

\author{Eyke H\"ullermeier\,${}^{a}$ and Willem Waegeman\,${}^{b}$\\[3mm]
\small ${}^{a}$\,Paderborn University\\[-1mm]
\small Heinz Nixdorf Institute and Department of Computer Science\\[-1mm]
\small eyke@upb.de\\[2mm]
\small ${}^{b}$\,Ghent University\\[-1mm]
\small Department of Mathematical Modelling, Statistics and Bioinformatics \\[-1mm]
\small willem.waegeman@ugent.be}

\date{}

\maketitle

\begin{abstract}
The notion of uncertainty is of major importance in machine learning and constitutes a key element of machine learning methodology. In line with the statistical tradition, uncertainty has long been perceived as almost synonymous with standard probability and probabilistic predictions. Yet, due to the steadily increasing relevance of machine learning for practical applications and related issues such as safety requirements, new problems and challenges have recently been identified by machine learning scholars, and these problems may call for new methodological developments. In particular, this includes the importance of distinguishing between (at least) two different types of uncertainty, often referred to as \emph{aleatoric} and \emph{epistemic}. In this paper, we provide an introduction to the topic of uncertainty in machine learning as well as an overview of attempts so far at handling uncertainty in general  and formalizing this distinction in particular. 
\end{abstract}

\section{Introduction}

Machine learning is essentially concerned with extracting models from data, often (though not exclusively) using them for the purpose of prediction. As such, it is inseparably connected with uncertainty. Indeed, learning in the sense of generalizing beyond the data seen so far is necessarily based on a process of \emph{induction}, i.e., replacing specific observations by general models of the data-generating process. Such models are never provably correct but only hypothetical and therefore uncertain, and the same holds true for the predictions produced by a model. In addition to the uncertainty inherent in inductive inference, other sources of uncertainty exist, including incorrect model assumptions and noisy or imprecise data.

Needless to say, a trustworthy representation of uncertainty is desirable and should be considered as a key feature of any machine learning method, all the more in safety-critical application domains such as medicine \citep{yang_ur09,lamb_rc11} or socio-technical systems \citep{vars_es16,vars_ot16}. 
Besides, uncertainty is also a major concept within machine learning methodology itself; for example, the principle of uncertainty reduction plays a key role in settings such as active learning \citep{agga_al14,mpub392}, or in concrete learning algorithms such as decision tree induction \citep{mitc_tn80}.

Traditionally, uncertainty is modeled in a probabilistic way, and indeed, in fields like statistics and machine learning, probability theory has always been perceived as the ultimate tool for uncertainty handling. Without questioning the probabilistic approach in general, one may argue that conventional approaches to probabilistic modeling, which are essentially based on capturing knowledge in terms of a single probability distribution, fail to distinguish two inherently different sources of uncertainty, which are often referred to as \emph{aleatoric} and \emph{epistemic} uncertainty \citep{hora_aa96,kiur_ao09}. Roughly speaking, aleatoric (\emph{aka} statistical) uncertainty refers to the notion of randomness, that is, the variability in the outcome of an experiment which is due to inherently random effects. The prototypical example of aleatoric uncertainty is coin flipping: The data-generating process in this type of experiment has a stochastic component that cannot be reduced by any additional source of information (except Laplace's demon). Consequently, even the best model of this process will only be able to provide probabilities for the two possible outcomes, heads and tails, but no definite answer. As opposed to this, epistemic (\emph{aka} systematic) uncertainty refers to uncertainty caused by a lack of knowledge (about the best model). In other words, it refers to the ignorance (cf.\ Sections \ref{sec:rlk} and \ref{sec:aproi}) of the agent or decision maker, and hence to the epistemic state of the agent instead of any underlying random phenomenon. As opposed to uncertainty caused by randomness, uncertainty caused by ignorance can in principle be reduced on the basis of additional information. For example, what does the word ``kichwa'' mean in the Swahili language, head or tail? The possible answers are the same as in coin flipping, and one might be equally uncertain about which one is correct. Yet, the nature of uncertainty is different, as one could easily get rid of it. 
In other words, epistemic uncertainty refers to the \emph{reducible} part of the (total) uncertainty, whereas aleatoric uncertainty refers to the \emph{irreducible} part. 

    \begin{figure}
    \centering
    \includegraphics[width=0.33\linewidth]{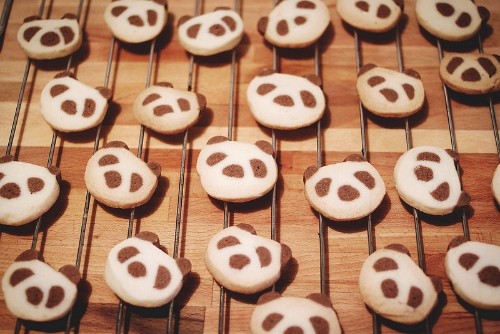}\qquad
     \includegraphics[width=0.3\linewidth]{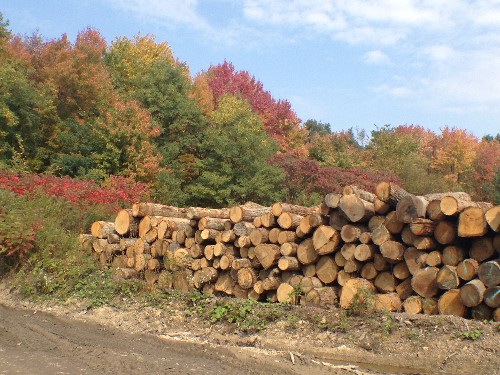}
    \caption{Predictions by EfficientNet \citep{tan_er19} on test images from ImageNet: For the left image, the neural network predicts ``typewriter keyboard'' with certainty  $83.14$\,\%, for the right image ``stone wall'' with certainty $87.63$\,\%.}
    \label{fig:image}
    \end{figure}

In machine learning, where the agent is a learning algorithm, the two sources of uncertainty are usually not distinguished. In some cases, such a distinction may indeed appear unnecessary. For example, if an agent is forced to make a decision or prediction, the source of its uncertainty---aleatoric or epistemic---might actually be irrelevant. This argument is often put forward by Bayesians in favor of a purely probabilistic approach (and classical Bayesian decision theory). One should note, however, that this scenario does not always apply. Instead, the ultimate decision can often be refused or delayed, like in classification with a reject option \citep{chow_oo70,hell_tn70}, or actions can be taken that are specifically directed at reducing uncertainty, like in active learning \citep{agga_al14}.


Motivated by such scenarios, and advocating a trustworthy representation of uncertainty in machine learning, \citet{mpub272} explicitly refer to the distinction between aleatoric and epistemic uncertainty. They propose a quantification of these uncertainties and show the usefulness of their approach in the context of medical decision making. A very similar motivation is given by \citet{kull_rm14} in the context of their work on reliability maps. They distinguish between a predicted probability score and the uncertainty in that prediction, and illustrate this distinction with an example from weather forecasting\footnote{The semantic interpretation of probability is arguably not very clear in examples of that kind: What exactly does it mean to have a 50\% chance of rain for the next day? This is not important for the point we want to make here, however.}: ``... a weather forecaster can be very certain that the chance of rain is 50\,\%; or her best estimate at 20\,\% might be very uncertain due to lack of data.'' Roughly, the 50\,\% chance corresponds to what one may understand as aleatoric uncertainty, whereas the uncertainty in the 20\,\% estimate is akin to the notion of epistemic uncertainty. On the more practical side, \citet{vars_ot16} give an example of a recent accident of a self-driving car, which led to the death of the driver (for the first time after 130 million miles of testing). They explain the car's failure by the extremely rare circumstances, and emphasize ``the importance of epistemic uncertainty or ``uncertainty on uncertainty'' in these AI-assisted systems''. 

More recently, a distinction between aleatoric and epistemic uncertainty has also been advocated in the literature on deep learning \citep{kend_wu17}, where the limited awareness of neural networks of their own competence has been demonstrated quite nicely. For example, experiments on image classification have shown that a trained model does often fail on specific instances, despite being very confident in its prediction (cf.\ Fig.\ \ref{fig:image}). Moreover, such models are often lacking robustness and can easily be fooled by ``adversarial examples'' \citep{pape_dk18}: Drastic changes of a prediction may already be provoked by minor, actually unimportant changes of an object. This problem has not only been observed for images but also for other types of data, such as natural language text (cf.\ Fig.\ \ref{fig:text} for an example).

    \begin{figure}
    \centering
    
\framebox{
\begin{minipage}{6cm}\footnotesize
There is really but one thing to say about \textbf{this} sorry movie It should
never have been made The first one one of my favourites An American
Werewolf in London is a great movie with a good plot good actors and
good FX But this one It stinks to heaven with a cry of helplessness
\end{minipage}
} \qquad
\framebox{
\begin{minipage}{6cm}\footnotesize
There is really but one thing to say about \textbf{that} sorry movie It should
never have been made The first one one of my favourites An American
Werewolf in London is a great movie with a good plot good actors and
good FX But this one It stinks to heaven with a cry of helplessness
\end{minipage}
}
 
\caption{Adversarial example (right) misclassified by a machine learning model trained on textual data: Changing only a single\,---\,and apparently not very important\,---\,word (highlighted in bold font) is enough to turn the correct prediction ``negative sentiment'' into the incorrect prediction ``positive sentiment'' \citep{sato_ia18}.}
\label{fig:text}
\end{figure}


This paper provides an overview of machine learning methods for handling uncertainty, with a specific focus on the distinction between aleatoric and epistemic uncertainty in the common setting of supervised learning. In the next section, we provide a short introduction to this setting and propose a distinction between different sources of uncertainty. 
Section 3 elaborates on modeling epistemic uncertainty. More specifically, it considers set-based modeling and modeling with (probability) distributions as two principled approaches to capturing the learner's  epistemic uncertainty about its main target, that is, the (Bayes) optimal model within the hypothesis space.
Concrete approaches for modeling and handling uncertainty in machine learning are then discussed in Section~4, prior to concluding the paper in Section~5. 
In addition, Appendix A provides some general background on uncertainty modeling (independent of applications in machine learning), specifically focusing on the distinction between set-based and distributional (probabilistic) representations and emphasizing the potential usefulness of combining them.




\begin{table}
\begin{center}
\caption{Notation used throughout the paper.}
\label{tab:notation}
\begin{tabular}{ll}
\hline
Notation  & Meaning \\
\hline
$\Prob$, $\prob$ & probability measure, density or mass function\\
$\mathcal{X}$, $\vec{x}$, $\vec{x}_i$ & instance space, instance \\
$\mathcal{Y}$, $y$, $y_i$ & output space, outcome \\
$\cH$, $h$ & hypothesis space, hypothesis\\
$\mathcal{V} = \mathcal{V}(\cH , \cD)$ & version space\\
$\pi(y)$ & possibility/plausibility of outcome $y$\\
$h(\vec{x})$ & outcome $y \in \cY$ for instance $\vec{x}$ predicted by hypothesis $h$\\
$p_h(y \given \vec{x}) = p(y \given \vec{x}, h)$ & probability of outcome $y$ given $\vec{x}$, predicted by $h$\\
$\ell$ & loss function \\ 
$h^*$, $\hat{h}$ & risk-minimizing hypothesis, empirical risk minimizer\\
$f^*$ & pointwise Bayes predictor\\
$\evalue(X)$ & expected value of random variable $X$\\
$\variance(X)$ & variance of random variable $X$\\
$\llbracket \cdot \rrbracket$ & indicator function \\
\hline
\end{tabular}
\end{center}
\end{table}

Table \ref{tab:notation} summarizes some important notation. Let us note that, for the sake of readability, a somewhat simplified notation will be used for probability measures and associated distribution functions. Most of the time, these will be denoted by $\Prob$ and $\prob$, respectively, even if they refer to different measures and distributions on different spaces (which will be clear from the arguments). For example, we will write $\prob( h )$ and $\prob(y \given \vec{x})$ for the probability (density) of hypotheses $h$ (as elements of the hypothesis space $\cH$) and outcomes $y$ (as elements of the output space $\cY$), instead of using different symbols, such as $\prob_\cH( h )$ and $\prob_\cY(y \given \vec{x})$.     

\section{Sources of uncertainty in supervised learning}
\label{sec:know}

\begin{figure}
\begin{center}
\includegraphics[scale=0.45]{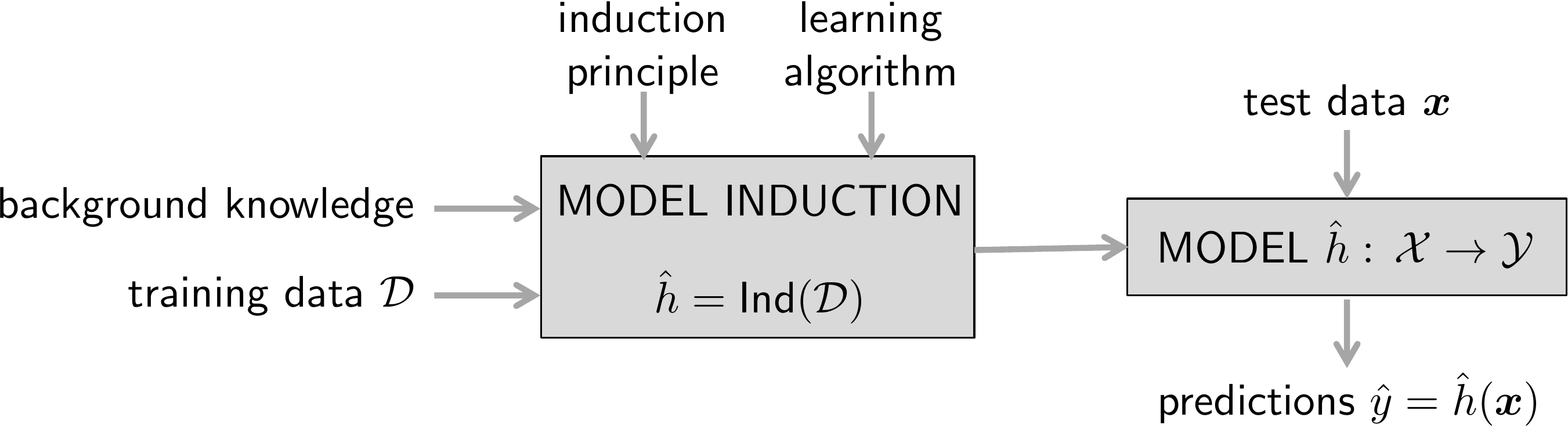}
\caption{The basic setting of supervised learning: A hypothesis $\hath \in \mathcal{H}$ is induced from the training data $\mathcal{D}$ and used to produce predictions for new query instances $\vec{x} \in \mathcal{X}$.}
\label{fig:setting}
\end{center}
\end{figure}

Uncertainty occurs in various facets in machine learning, and different settings and learning problems will usually require a different handling from an uncertainty modeling point of view. In this paper, we focus on the standard setting of \emph{supervised learning} (cf.\ Fig.\ \ref{fig:setting}), which we briefly recall in this section. Moreover, we identify different sources of (predictive) uncertainty in this setting.

\subsection{Supervised learning and predictive uncertainty} 

In supervised learning, a learner is given access to a set of training data 
\begin{equation}\label{eq:td}
\mathcal{D} \defeq \big\{ (\vec{x}_1 , y_1 ), \ldots , (\vec{x}_N , y_N ) \big\} \subset \mathcal{X} \times \mathcal{Y} \enspace ,
\end{equation}
where $\mathcal{X}$ is an instance space and $\mathcal{Y}$ the set of outcomes that can be associated with an instance. Typically, the training examples $(\vec{x}_i , y_i)$ are assumed to be independent and identically distributed (i.i.d.) according to some unknown probability measure $\Prob$ on $\mathcal{X} \times \mathcal{Y}$. Given a \emph{hypothesis space} $\mathcal{H}$ (consisting of hypotheses $h:\, \cX \fromto \cY$ mapping instances $\vec{x}$ to outcomes $y$) and a loss function $\ell: \, \mathcal{Y} \times \mathcal{Y} \longrightarrow \mathbb{R}$, the goal of the learner is to induce a hypothesis $h^* \in \mathcal{H}$ with low risk (expected loss)
\begin{equation}
R(h) \defeq \int_{\cX \times \cY} \ell( h(\vec{x}) , y) \, d \, \Prob(\vec{x} , y) \enspace .
\end{equation}
Thus, given the training data $\cD$, the learner needs to ``guess'' a good hypothesis $h$. This choice is commonly guided by the empirical risk 
\begin{equation}
R_{emp}(h) \defeq  \frac{1}{N} \sum_{i=1}^N \ell(h(\vec{x}_i) , y_i) \enspace ,
\end{equation}
i.e., the performance of a hypothesis on the training data. However, since $R_{emp}(h)$ is only an estimation of the true risk $R(h)$, the hypothesis (empirical risk minimizer) 
\begin{equation}\label{eq:argerm}
\hath \defeq \argmin_{h \in \cH} R_{emp}(h)
\end{equation}
favored by the learner will normally not coincide with the true risk minimizer 
\begin{equation}\label{eq:bayespred}
h^* \defeq \argmin_{h \in \cH} R(h) \, .
\end{equation} 
Correspondingly, there remains uncertainty regarding $h^*$ as well as the approximation quality of $\hath$ (in the sense of its proximity to $h^*$) and its true risk $R(\hath)$.

Eventually, one is often interested in \emph{predictive uncertainty}\mmp{predictive uncertainty}, i.e., the uncertainty related to the prediction $\haty_{q}$ for a concrete query instance $\vec{x}_{q} \in \cX$. In other words, given a partial observation $(\vec{x}_{q} , \cdot)$, we are wondering what can be said about the missing outcome, especially about the uncertainty related to a prediction of that outcome. Indeed, estimating and quantifying uncertainty in a transductive way, in the sense of tailoring it for individual instances, is arguably important and practically more relevant than a kind of average accuracy or confidence, which is often reported in machine learning. In medical diagnosis, for example, a patient will be interested in the reliability of a test result in her particular case, not in the reliability of the test on average. This view is also expressed, for example, by \citet{kull_rm14}: ``Being able to assess the reliability of a probability score for each instance is much more powerful than assigning an aggregate reliability score [...] independent of the instance to be classified.''

Emphasizing the transductive nature of the learning process, the learning task could also be formalized as a problem of \emph{predictive inference} as follows:  Given a set (or sequence) of data points $(X_1, Y_1), \ldots , (X_N, Y_N)$ and a query point $X_{N+1}$, what is the associated outcome $Y_{N+1}$? Here, the data points are considered as (realizations of) random variables\footnote{whence they are capitalized here}, which are commonly assumed to be independent and identically distributed (\emph{i.i.d.}). A prediction could be given in the form of a point prediction $\hat{Y}_{N+1} \in \cY$, but also (and perhaps preferably)  in the form of a \emph{predictive set} $\hat{C}(X_{N+1}) \subseteq \cY$ that is likely to cover the true outcome, for example an interval in the case of regression (cf.\ Sections \ref{sec:cp} and \ref{sec:sbus}). Regarding the aspect of uncertainty, different properties of $\hat{C}(X_{N+1})$ might then be of interest. For example, coming back to the discussion from the previous paragraph, a basic distinction can be made between a statistical guarantee for \emph{marginal coverage}, namely $\Prob( Y_{N+1} \in \hat{C}(X_{N+1}) ) \geq 1 - \delta$ for a (small) threshold $\delta > 0$, and \emph{conditional coverage}, namely 
$$
\Prob \Big( Y_{N+1} \in \hat{C}(X_{N+1}) \given X_{N+1} = \vec{x} \Big) \geq 1 - \delta \quad \text{ for (almost) all } \vec{x} \in \cX \, .
$$
Roughly speaking, in the case of marginal coverage, one averages over both $X_{N+1}$ and $Y_{N+1}$ (i.e., the probability $\Prob$ refers to a joint measure on $\cX \times \cY$), while in the case of conditional coverage, $X_{N+1}$ is fixed and the average in taken over $Y_{N+1}$ only \citep{barb_tl20}. Note that predictive inference as defined here does not necessarily require the induction of a hypothesis $\hath$ in the form of a (global) map $\cX \fromto \cY$, i.e., the solution of an induction problem. While it is true that transductive inference can be realized via inductive inference (i.e., by inducing a hypothesis $\hath$ first and then producing a prediction $\hat{Y}_{N+1} = \hath(X_{N+1})$ by applying this hypothesis to the query $X_{N+1}$), one should keep in mind that induction is a more difficult problem than transduction \citep{vapn_sl98}.


\subsection{Sources of uncertainty}

\begin{figure}
\begin{center}
\includegraphics[scale=0.45]{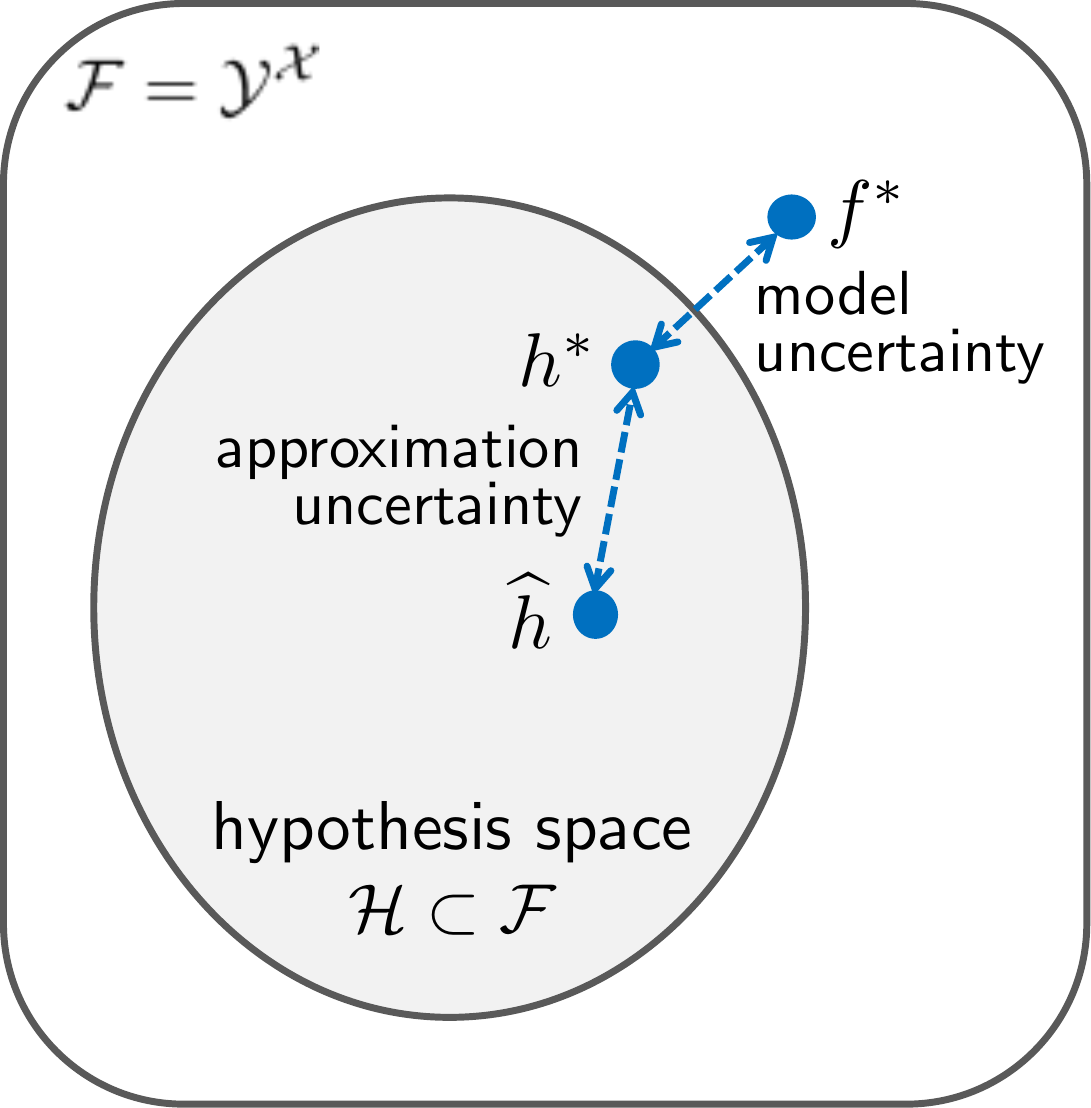} \qquad
\small
\begin{tabular}[b]{lcc}
\hline
 & point prediction & probability \\
\hline
ground truth & $f^*(\vec{x})$ & $\prob( \cdot \given \vec{x})$ \\
best possible & $h^*(\vec{x})$ & $\prob( \cdot \given \vec{x}, h^*)$ \\
induced predictor & $\hath(\vec{x})$ & $\prob( \cdot \given \vec{x}, \hath)$\\
\hline
\end{tabular}
\caption{Different types of uncertainties related to different types of discrepancies and approximation errors:  $f^*$ is the pointwise Bayes predictor, $h^*$ is the best predictor within the hypothesis space, and $\hath$ the predictor produced by the learning algorithm.}
\label{fig:approx}
\end{center}
\end{figure}

As the prediction $\haty_{q}$ constitutes the end of a process that consists of different learning and approximation steps, all errors and uncertainties related to these steps may also contribute to the uncertainty about $\haty_{q}$ (cf.\ Fig.\ \ref{fig:approx}):
\begin{itemize}
\item Since the dependency between $\cX$ and $\cY$ is typically non-deterministic, the description of a new prediction problem in the form of an instance $\vec{x}_{q}$ gives rise to a conditional probability distribution
\begin{equation}\label{eq:ccp}
\prob( y \given \vec{x}_{q}) = \frac{\prob(\vec{x}_{q} , y)}{\prob(\vec{x}_q)} 
\end{equation}
on $\cY$, but it does normally not identify a single outcome $y$ in a unique way. Thus, even given full information in the form of the measure $\Prob$ (and its density $\prob$), uncertainty about the actual outcome $y$ remains. This uncertainty is of an \emph{aleatoric} nature. In some cases, the distribution (\ref{eq:ccp}) itself (called the predictive posterior distribution in Bayesian inference) might be delivered as a prediction. Yet, when being forced to commit to point estimates, the best predictions (in the sense of minimizing the expected loss) are prescribed by the pointwise Bayes predictor $f^*$, which is defined by
\begin{equation}\label{eq:pointbayespred}
f^*(\vec{x}) \defeq \argmin_{\haty \in \cY} \int_\cY \ell(y , \haty) \, d \Prob( y \given \vec{x} )
\end{equation} 
for each $\vec{x} \in \cX$.

\item The Bayes predictor (\ref{eq:bayespred}) does not necessarily coincide with the pointwise Bayes predictor (\ref{eq:pointbayespred}). This discrepancy between $h^*$ and $f^*$ is connected to the uncertainty regarding the right type of model to be fit, and hence the choice of the hypothesis space $\cH$ (which is part of what is called ``background knowledge'' in Fig.\ \ref{fig:setting}). We shall refer to this uncertainty as \emph{model uncertainty}\footnote{The notion of a model is not used in a unique way in the literature. Here, as already said, we essentially refer to the choice of the hypothesis space $\cH$, so a model could be a linear model or a Gaussian process. Sometimes, a single hypothesis $h \in \cH$ is also called a model. Then, ``model uncertainty'' might be incorrectly understood as uncertainty about this hypothesis, for which we shall introduce the term ``approximation uncertainty''.}. Thus, due to this uncertainty, one can not guarantee that $h^*(\vec{x}) = f^*(\vec{x})$, or, in case the hypothesis $h^*$ (e.g., a probabilistic classifier) delivers probabilistic predictions $\prob(y \given \vec{x}, h^*)$ instead of point predictions, that $\prob( \cdot \given \vec{x}, h^*) = \prob( \cdot \given \vec{x})$.

\item The hypothesis $\hath$ produced by the learning algorithm, for example the empirical risk minimizer (\ref{eq:argerm}), is only an estimate of $h^*$, and the quality of this estimate strongly depends on the quality and the amount of training data. We shall refer to the discrepancy between $\hath$ and $h^*$, i.e., the uncertainty about how well the former approximates the latter, as \emph{approximation uncertainty}\mmp{approximation uncertainty}. 

\end{itemize}

\subsection{Reducible versus irreducible uncertainty}

As already said, one way to characterize uncertainty as aleatoric or epistemic is to ask whether or not the uncertainty can be reduced through additional information: Aleatoric uncertainty refers to the irreducible part of the uncertainty, which is due to the non-deterministic nature of the sought input/output dependency, that is, to the stochastic dependency between instances $\vec{x}$ and outcomes $y$, as expressed by the conditional probability (\ref{eq:ccp}). Model uncertainty and approximation uncertainty, on the other hand, are subsumed under the notion of epistemic uncertainty, that is, uncertainty due to a lack of knowledge about the perfect predictor (\ref{eq:pointbayespred}), for example caused by uncertainty about the parameters of a model. In principle, this uncertainty can be reduced. 

\begin{figure}
\captionsetup[subfigure]{labelformat=empty}
	\centering
	\subfloat{
		\includegraphics[width=0.4\textwidth]{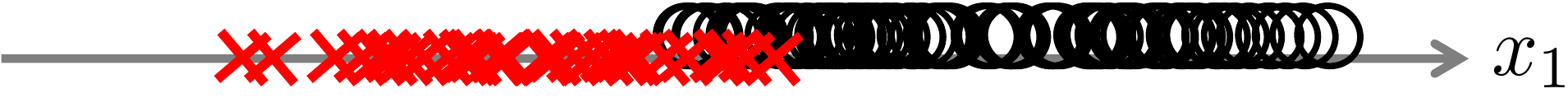}
		}
		\quad
		\subfloat{
		\includegraphics[width=0.4\textwidth]{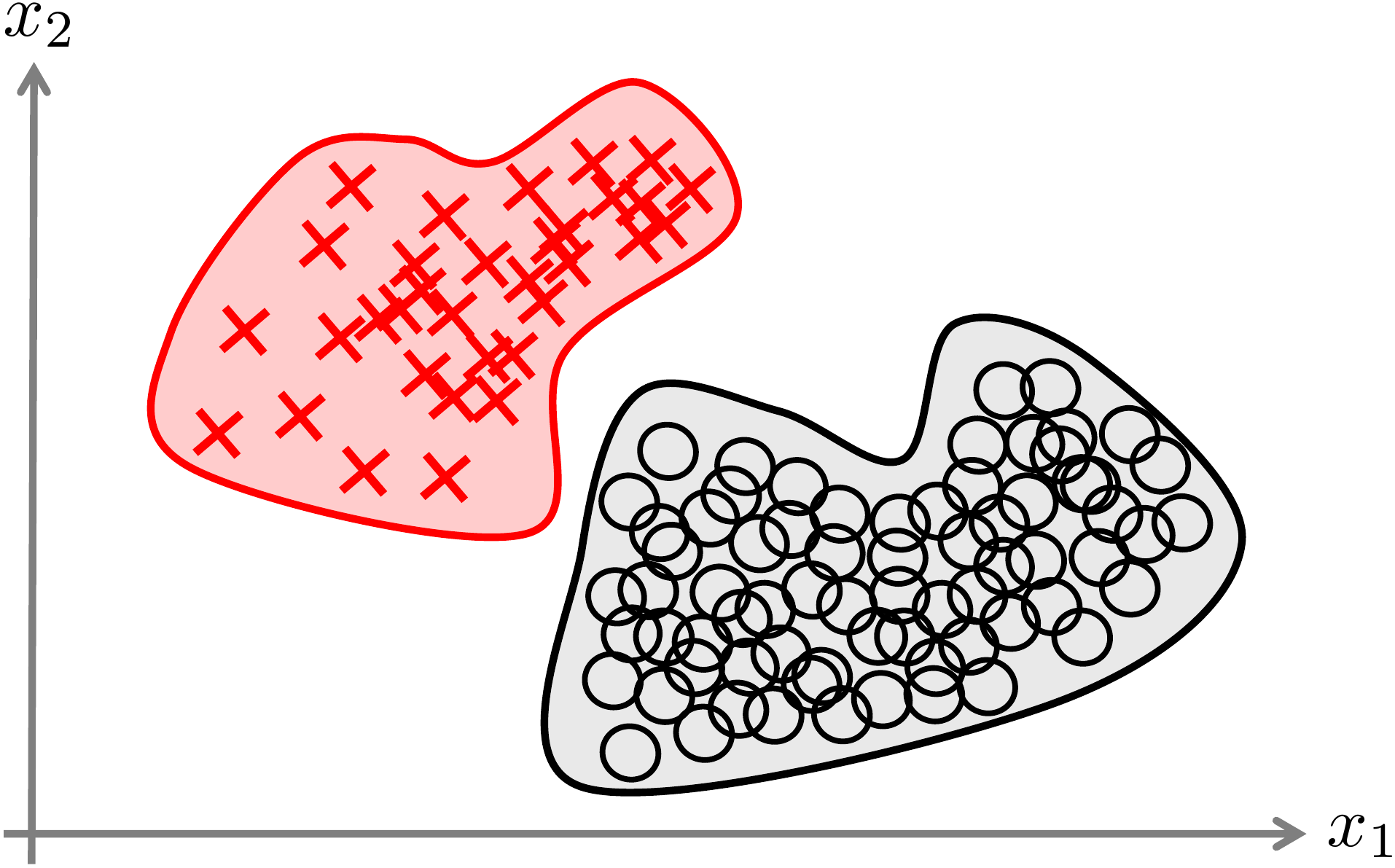}
		}
		\caption{Left: The two classes are overlapping, which causes (aleatoric) uncertainty in a certain region of the instance space. Right: By adding a second feature, and hence embedding the data in a higher-dimensional space, the two classes become separable, and the uncertainty can be resolved.}
		\label{fig:dimen}
\end{figure}

This characterization, while evident at first sight, may appear somewhat blurry upon closer inspection.
What does ``reducible'' actually mean? An obvious source of additional information is the training data $\mathcal{D}$: The learner's uncertainty can be reduced by observing more data, while the setting of the learning problem\,---\,the instance space $\mathcal{X}$, output space $\mathcal{Y}$, hypothesis space $\mathcal{H}$, joint probability $\Prob$ on $\mathcal{X} \times \mathcal{Y}$\,---\,remains fixed. In practice, this is of course not always the case. Imagine, for example, that a learner can decide to extend the description of instances by additional features, which essentially means replacing the current instance space $\cX$ by another space $\cX'$. This change of the setting may have an influence on uncertainty. An example is shown in Fig.\ \ref{fig:dimen}: In a low-dimensional space (here defined by a single feature $x_1$), two class distributions are overlapping, which causes (aleatoric) uncertainty in a certain region of the instance space. By embedding the data in a higher-dimensional space (here accomplished by adding a second feature $x_2$), the two classes become separable, and the uncertainty can be resolved. More generally, embedding data in a higher-dimensional space will reduce aleatoric and increase epistemic uncertainty, because fitting a model will become more difficult and require more data.

What this example shows is that aleatoric and epistemic uncertainty should not be seen as absolute notions. Instead, they are context-dependent in the sense of depending on the setting $(\cX, \cY, \cH, \Prob)$. Changing the context will also change the sources of uncertainty: aleatoric may turn into epistemic uncertainty and vice versa. Consequently, by allowing the learner to change the setting, the distinction between these two types of uncertainty will be somewhat blurred (and their quantification will become even more difficult). This view of the distinction between aleatoric and epistemic uncertainty is also shared by \cite{kiur_ao09}, who note that ``these concepts only make unambiguous sense if they are defined within the confines of a model of analysis'', and that ``In one model an addressed uncertainty may be aleatory, in another model it may be epistemic.''


\subsection{Approximation and model uncertainty}
\label{sec:aamu}

\begin{figure}
\begin{center}
\includegraphics[scale=0.30]{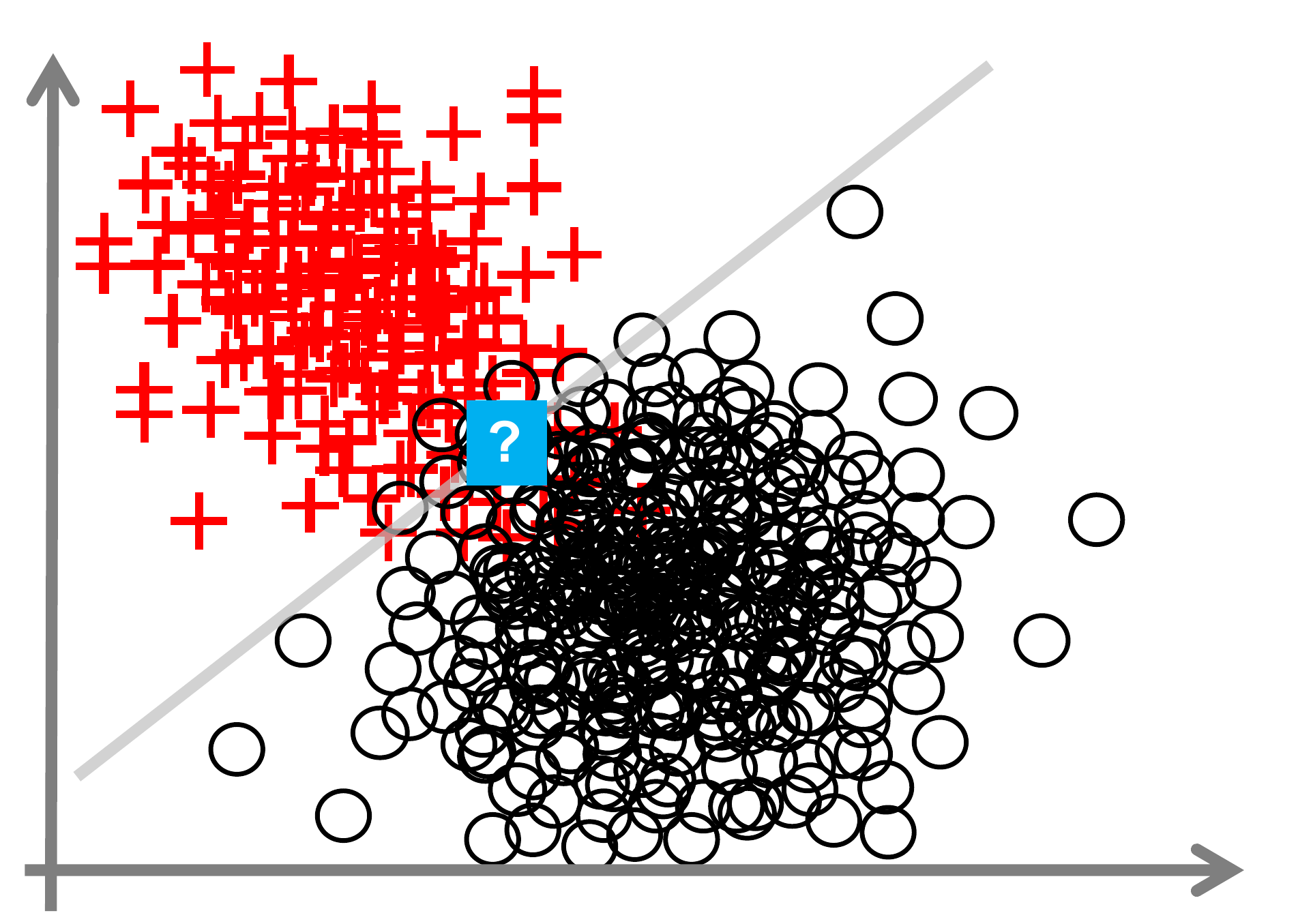} \quad 
\includegraphics[scale=0.30]{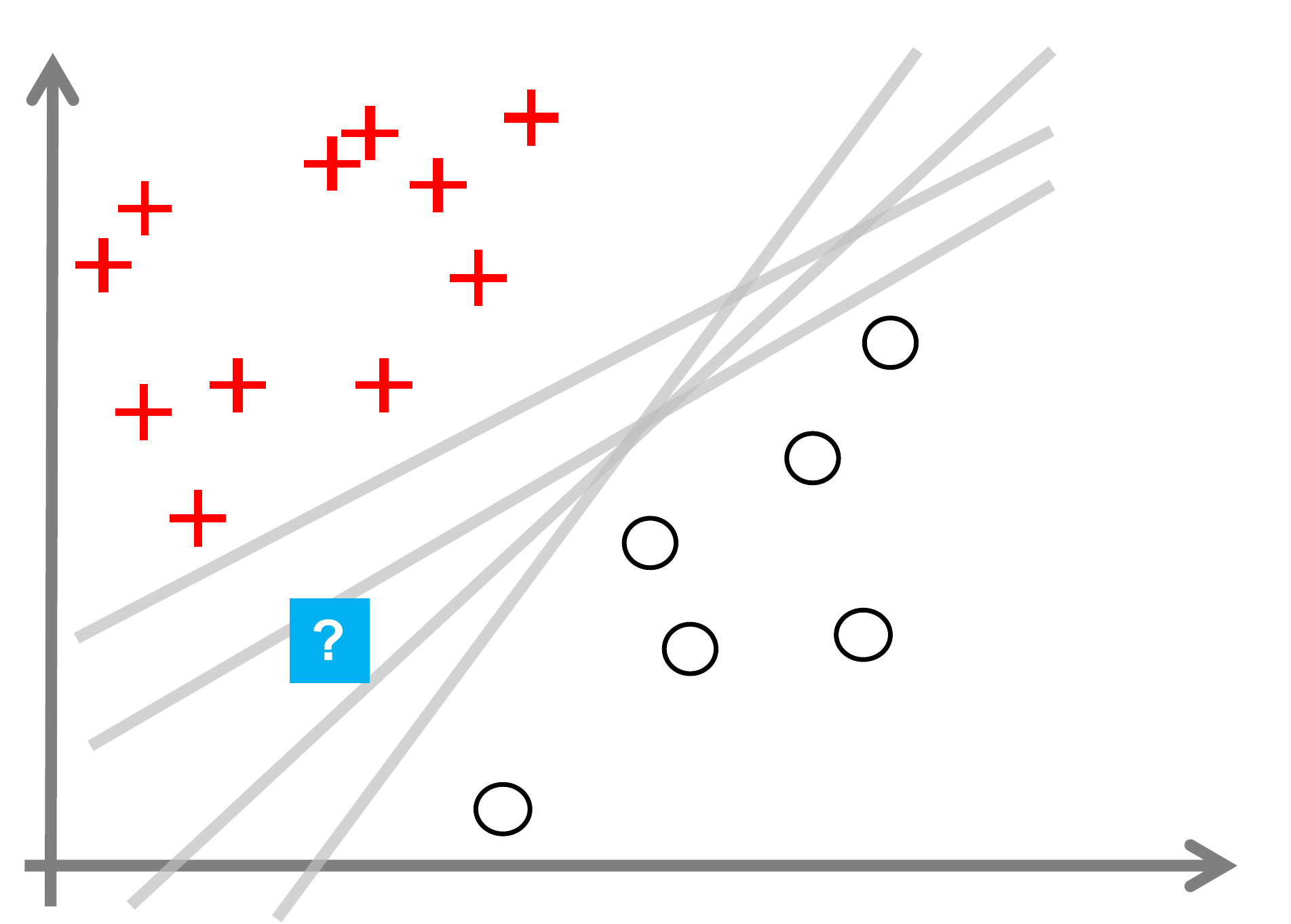}
\caption{Left: Even with precise knowledge about the optimal hypothesis, the prediction at the query point (indicated by a question mark) is aleatorically uncertain, because the two classes are overlapping in that region. Right: A case of epistemic uncertainty due to a lack of knowledge about the right hypothesis, which is in turn caused by a lack of data.}
\label{fig:u}
\end{center}
\end{figure}

Assuming the setting $(\cX, \cY, \cH, \Prob)$ to be fixed, the learner's lack of knowledge will essentially depend on the amount of data it has seen so far: The larger the number $N = | \mathcal{D}|$ of observations, the less ignorant the learner will be when having to make a new prediction. In the limit, when $N \rightarrow \infty$, a consistent learner will be able to identify $h^*$  (see Fig.\ \ref{fig:u} for an illustration), i.e., it will get rid of its approximation uncertainty. 

What is (implicitly) assumed here is a correctly specified hypothesis space $\mathcal{H}$, such that $f^* \in \mathcal{H}$. In other words, model uncertainty is simply ignored.  
For obvious reasons, this uncertainty is very difficult to capture, let alone quantify. In a sense, a kind of meta-analysis would be required: Instead of expressing uncertainty about the ground-truth hypothesis $h$ within a hypothesis space $\mathcal{H}$, one has to express uncertainty about which $\mathcal{H}$ among a set $\mathbb{H}$ of candidate hypothesis spaces might be the right one.   
Practically, such kind of analysis does not appear to be feasible. On the other side, simply assuming a correctly specified hypothesis space $\mathcal{H}$ actually means neglecting the risk of model misspecification. To some extent, this appears to be unavoidable, however. In fact, the learning itself as well as all sorts of inference from the data are normally done under the assumption that the model is valid. Otherwise, since some assumptions are indeed always needed, it will be difficult to derive any useful conclusions .

As an aside, let us note that these assumptions, in addition to the nature of the ground truth $f^*$, also include other (perhaps more implicit) assumptions about the setting and the data-generating process. For example, imagine that a sudden change of the distribution cannot be excluded, or a strong discrepancy between training and test data \citep{mali_pu18}. Not only prediction but also the assessment of uncertainty would then become difficult, if not impossible. Indeed, if one cannot exclude something completely unpredictable to happen, there is hardly any way to reduce predictive uncertainty. To take a simple example, (epistemic) uncertainty about the bias of a coin can be estimated and quantified, for example in the form of a confidence interval, from an i.i.d.\ sequence of coin tosses. But what if the bias may change from one moment to the other (e.g., because the coin is replaced by another one), or another outcome becomes possible (e.g., ``invalid'' if the coin has not been tossed in agreement with a new execution rule)? While this example may appear a bit artificial, there are indeed practical classification problems in which certain classes $y \in \cY$ may ``disappear'' while new classes emerge, i.e., in which $\cY$ may change in the course of time. Likewise, non-stationarity of the data-generating process (the measure $\Prob$), including the possibility of drift or shift of the distribution, is a common assumption in learning on data streams \citep{gama_as12}.




Coming back to the assumptions about the hypothesis space $\mathcal{H}$, the latter is actually very large for some learning methods, such as nearest neighbor classification or (deep) neural networks. Thus, the learner has a high capacity (or ``universal approximation'' capability) and can express hypotheses in a very flexible way. In such cases, $h^* = f^*$ or at least $h^* \approx f^*$ can safely be assumed. In other words, since the model assumptions are so weak, model uncertainty essentially disappears (at least when disregarding or taking for granted other assumptions as discussed above). Yet, the approximation uncertainty still remains a source of epistemic uncertainty. In fact, this uncertainty tends to be high for methods like nearest neighbors or neural networks, especially if data is sparse. 

In the next section, we recall two specific though arguably natural and important approaches for capturing this uncertainty, namely version space learning\mmp{version space learning} and Bayesian inference. In version space learning, uncertainty about $h^*$ is represented in terms of a \emph{set} of possible candidates, whereas in Bayesian learning, this uncertainty is modeled in terms of a probability \emph{distribution} on $\mathcal{H}$. In both cases, an explicit distinction is made between the uncertainty about $h^*$, and how this uncertainty translates into uncertainty about the outcome for a query $\vec{x}_q$.

\section{Modeling approximation uncertainty: Set-based versus distributional representations}
\label{sec:sbvd}

Bayesian inference can be seen as the main representative of probabilistic methods and provides a coherent framework for statistical reasoning that is well-established in machine learning (and beyond). Version space learning can be seen as a ``logical'' (and in a sense simplified)  counterpart of Bayesian inference, in which hypotheses and predictions are not assessed numerically in terms of probabilities, but only qualified (deterministically) as being possible or impossible. In spite of its limited practical usefulness, version space learning is interesting for various reasons. In particular, in light of our discussion about uncertainty, it constitutes an interesting case: By construction, version space learning  is free of aleatoric uncertainty, i.e., all uncertainty is epistemic.

\subsection{Version space learning}
\label{sec:vsl}

In the idealized setting of version space learning, we assume a deterministic dependency $f^*:\, \cX \longrightarrow \cY$, i.e., the distribution (\ref{eq:ccp}) degenerates to
\begin{equation}\label{eq:ccpvs}
\prob( y \given \vec{x}_{q}) = \left\{ \begin{array}{ll}
1 & \text{ if } y = f^*(\vec{x}_{q}) \\
0 & \text{ if } y \neq f^*(\vec{x}_{q}) \\
\end{array} \right.
\end{equation}
Moreover, the training data (\ref{eq:td}) is free of noise. Correspondingly, we also assume that classifiers produce deterministic predictions $h(\vec{x}) \in \{ 0, 1 \}$ in the form of probabilities 0 or 1. Finally, we assume that $f^* \in \cH$, and therefore $h^* = f^*$ (which means there is no model uncertainty). 

Under these assumptions, a hypothesis $h \in \cH$ can be eliminated as a candidate as soon as it makes at least one mistake on the training data: in that case, the risk of $h$ is necessarily higher than the risk of $h^*$ (which is 0). The idea of the candidate elimination algorithm \citep{mitc_vs77} is to maintain the \emph{version space} $\mathcal{V} \subseteq \cH$ that consists of the set of all hypotheses consistent with the data seen so far:
\begin{equation}
\mathcal{V} = \mathcal{V}(\cH , \cD) \defeq \Big\{ h \in \cH \with h(\vec{x}_i) = y_i \text{ for } i = 1, \ldots , N \Big\}
\end{equation}
Obviously, the version space is shrinking with an increasing amount of training data, i.e., $\mathcal{V}(\cH , \cD') \subseteq \mathcal{V}(\cH , \cD)$ for $\cD \subseteq \cD'$. 

\begin{figure}
\begin{center}
\includegraphics[scale=0.45]{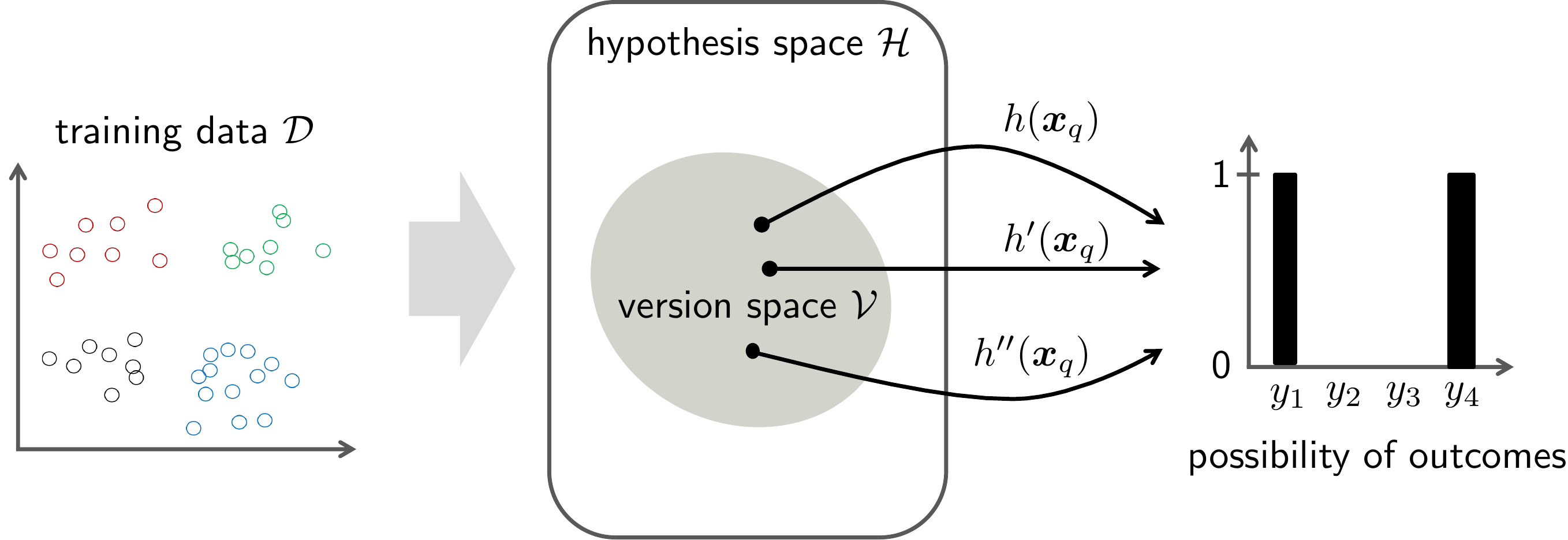} 
\caption{Illustration of version space learning inference. The version space $\mathcal{V}$, which is the subset of hypotheses consistent with the data seen so far, represents the state of knowledge of the learner. For a query $\vec{x}_{q} \in \mathcal{X}$, the set of possible predictions is given by the set $Y = \{ h(\vec{x}_q) \given h \in \mathcal{V}\}$. The distribution on the right is the characteristic function $\pi:\, \mathcal{Y} \longrightarrow \{ 0,1 \}$ of this set, which can be interpreted as a $\{0,1\}$-valued possibility distribution (indicating whether $y$ is a possible outcome or not, i.e., $\pi(y) = \llbracket y \in Y \rrbracket$).}
\label{fig:vs}
\end{center}
\end{figure}

If a prediction $\haty_{q}$ for a query instance $\vec{x}_{q}$ is sought, this query is submitted to all members $h \in \mathcal{V}$ of the version space. Obviously, a unique prediction can only be made if all members agree on the outome of $\vec{x}_{q}$. Otherwise, several outcomes $y \in \mathcal{Y}$ may still appear possible. Formally, mimicking the logical conjunction with the minimum operator and the existential quantification with a maximum, we can express the degree of possibility or plausibility of an outcome $y \in \cY$ as follows ($\llbracket \cdot \rrbracket$ denotes the indicator function):
\begin{equation}\label{eq:ee1}
\pi( y) \defeq \max_{h \in \cH} \min \Big( \llbracket h \in \mathcal{V} \rrbracket , \llbracket h(\vec{x}_q) = y \rrbracket \Big)
\end{equation}
Thus, $\pi(y)=1$ if there exists a candidate hypothesis $h \in \mathcal{V}$ such that $h(\vec{x}_{q}) = y$, and $\pi(y)=0$ otherwise. In other words, the prediction produced in version space learning is a subset
\begin{equation}\label{eq:vss}
Y = Y(\vec{x}_q) \defeq  \{ h(\vec{x}_q) \given h \in \mathcal{V}\}  = \{ y \given \pi(y) = 1 \}  \subseteq \mathcal{Y} 
\end{equation} 
See Fig.\ \ref{fig:vs} for an illustration.

\begin{figure}
\captionsetup[subfigure]{labelformat=empty}
	\centering
	\subfloat{
		\includegraphics[width=0.44\textwidth]{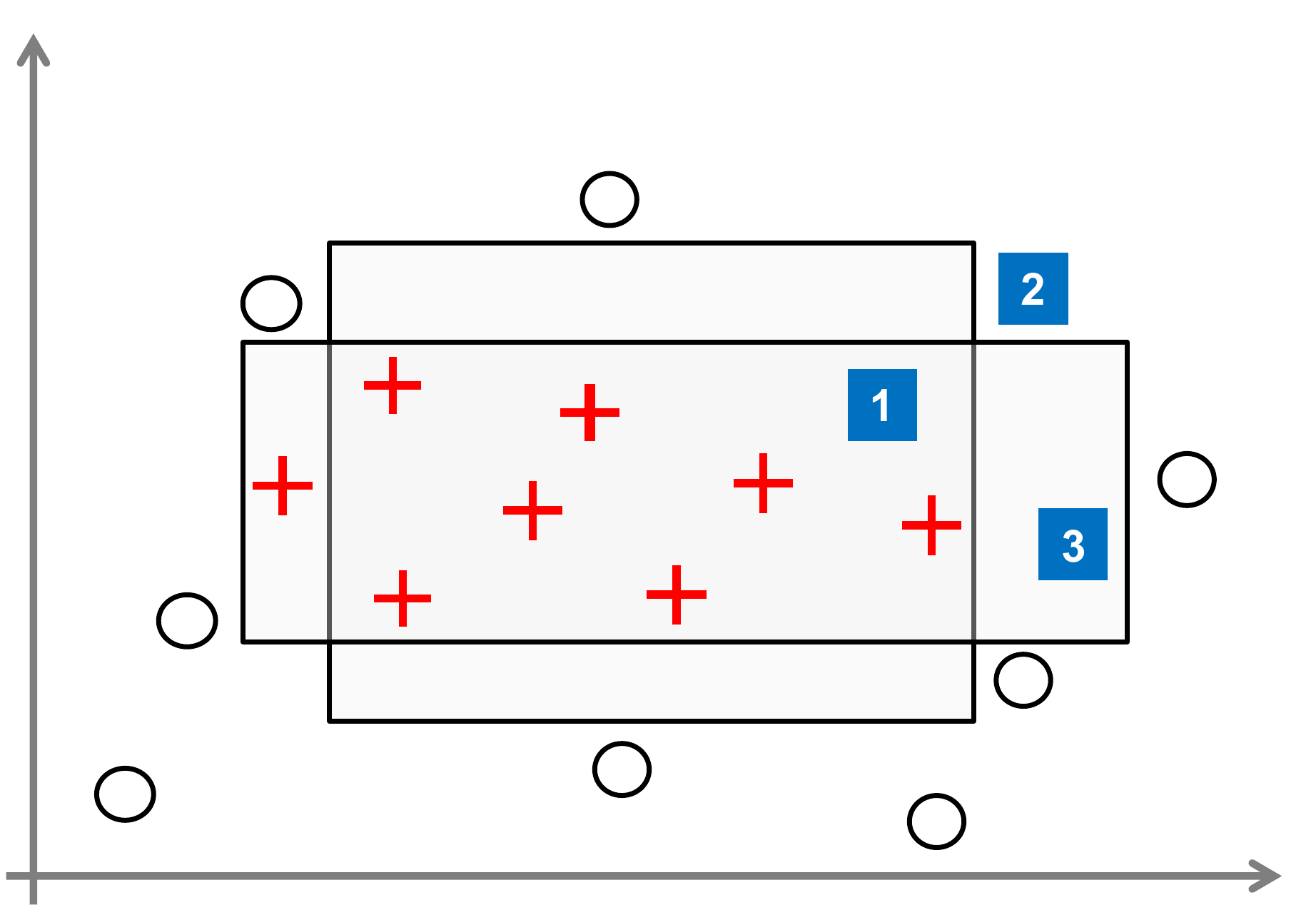}
		}
		\subfloat{
		\includegraphics[width=0.44\textwidth]{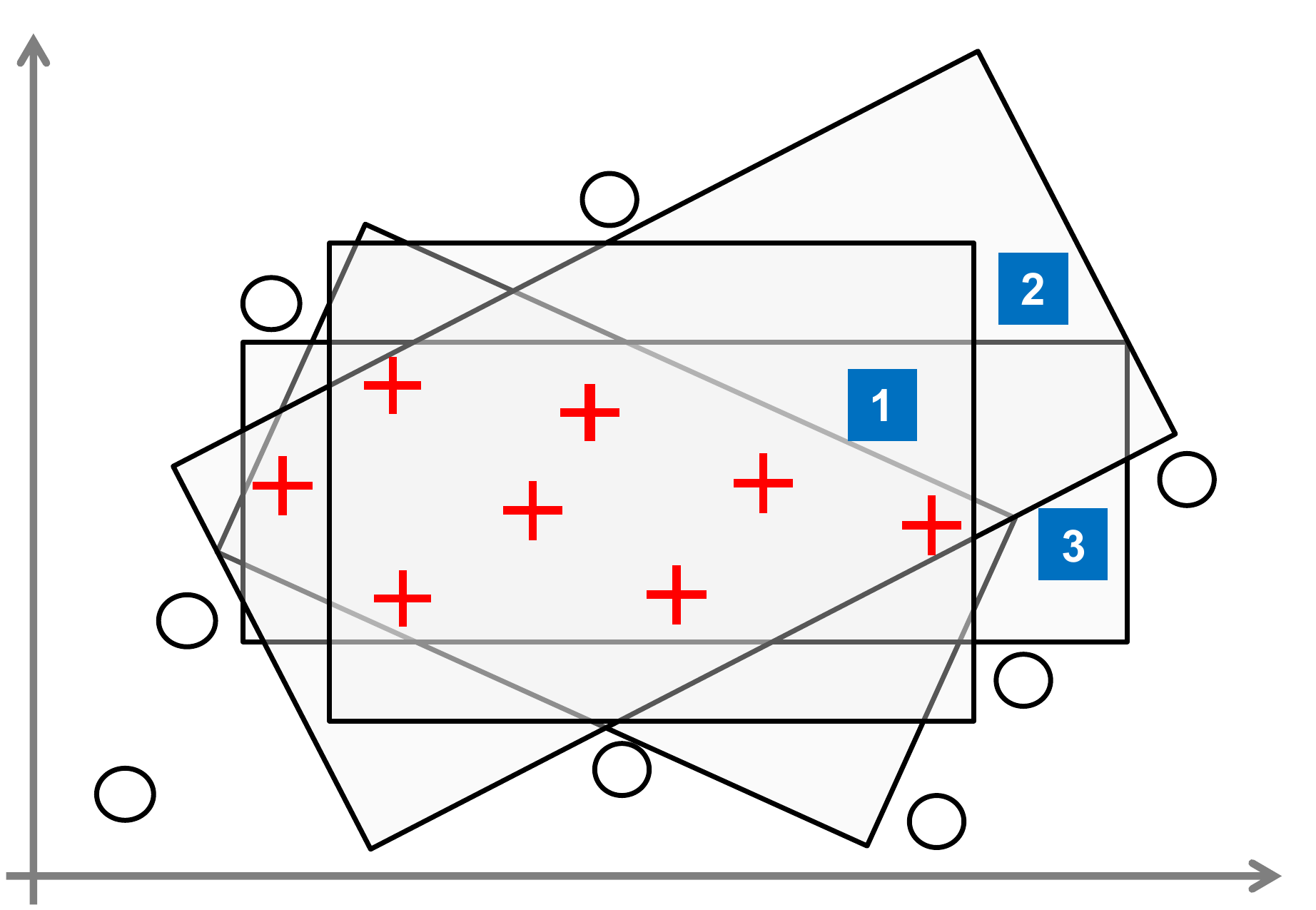}
		}\qquad
		\subfloat{
		\includegraphics[width=0.44\textwidth]{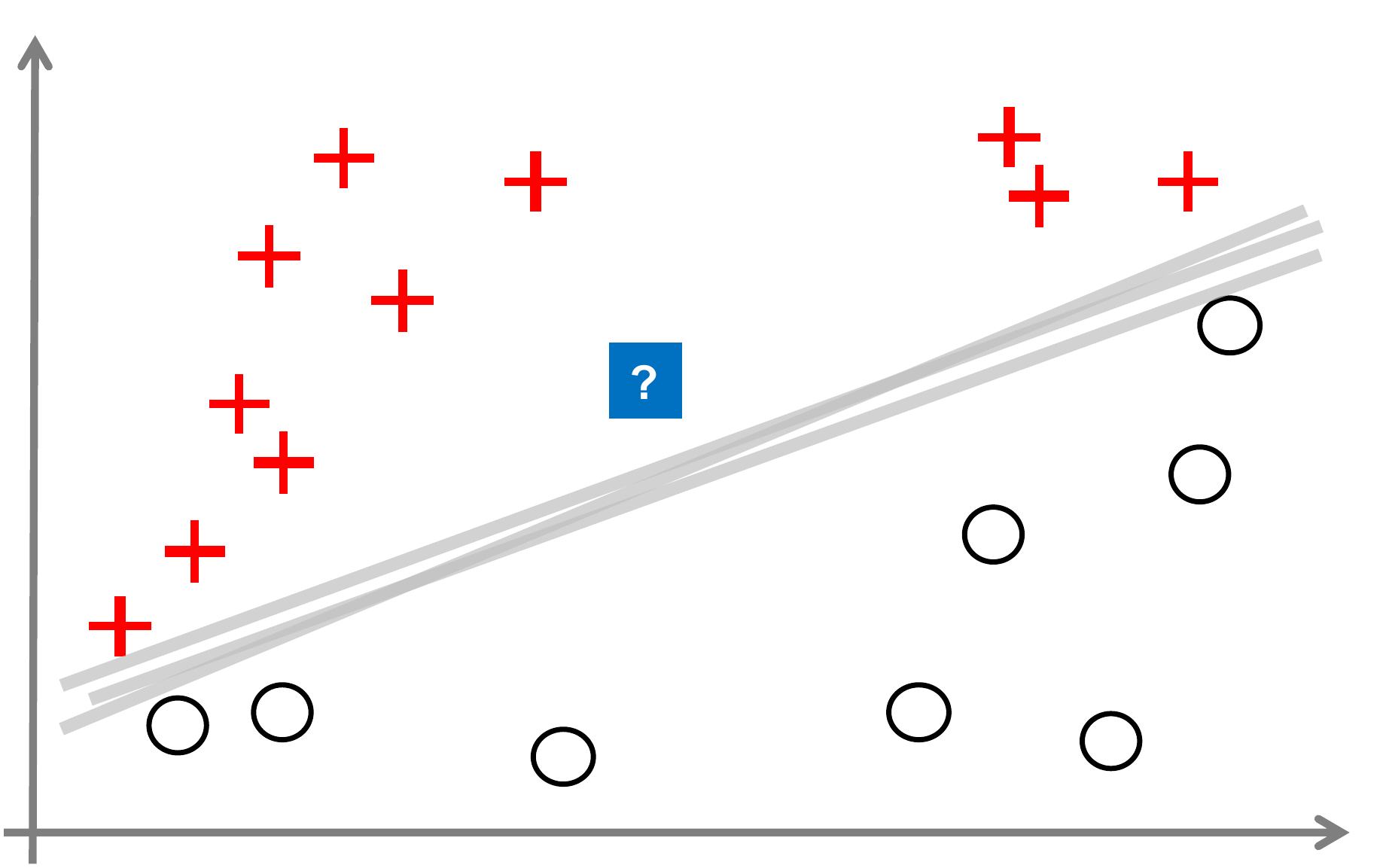}
	    }
		\subfloat{
		\includegraphics[width=0.44\textwidth]{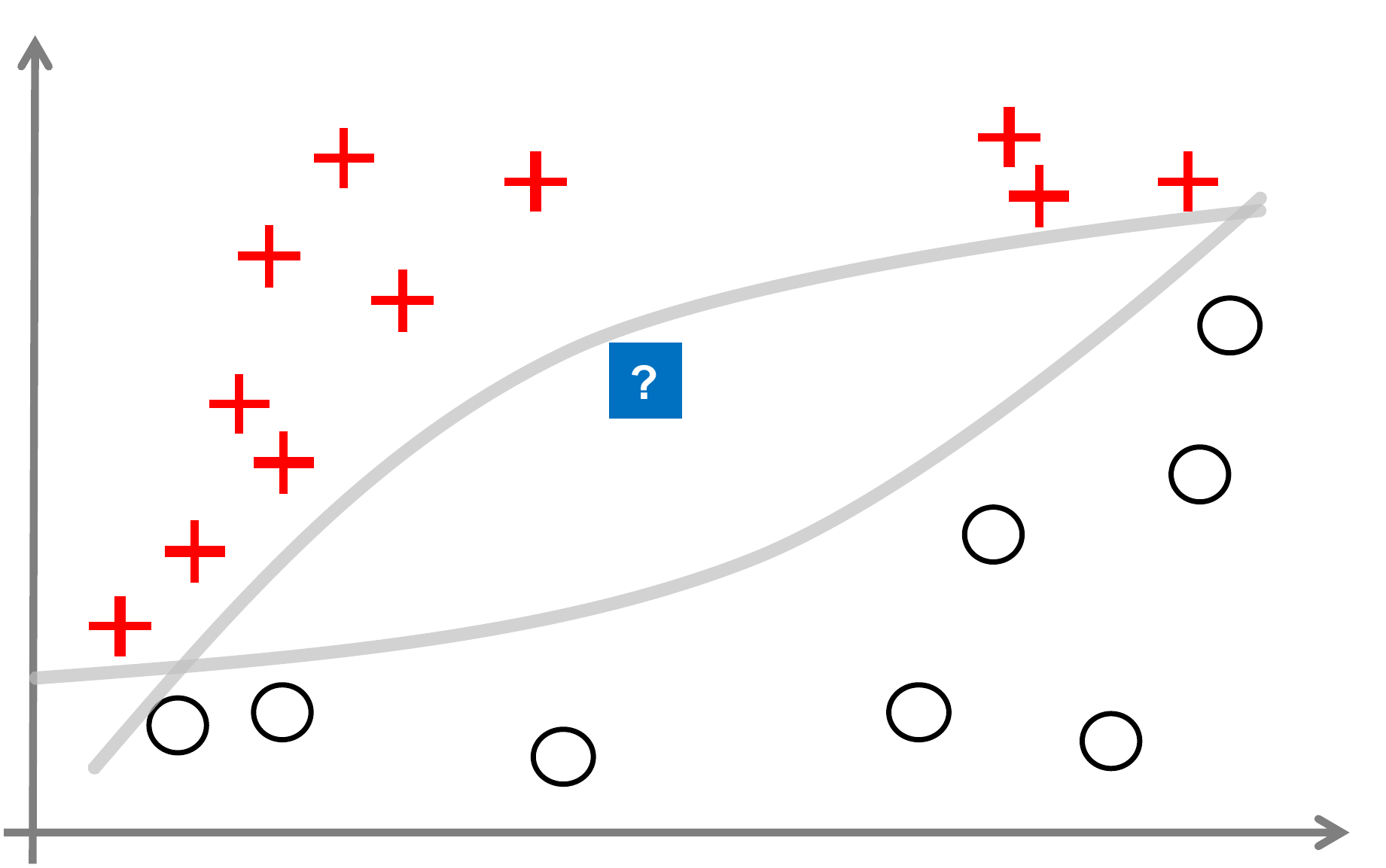}
		}
		\caption{Two examples illustrating predictive uncertainty in version space learning for the case of binary classification.
In the first example (above), the hypothesis space is given in terms of rectangles (assigning interior instances to the positive class and instances outside to the negative class). Both pictures show some of the (infinitely many) elements of the version space. Top left: Restricting $\mathcal{H}$ to axis-parallel rectangles, the first query point is necessarily positive, the second one necessarily negative, because these predictions are produced by all $h \in \mathcal{V}$. As opposed to this, both positive and negative predictions can be produced for the third query. Top right: Increasing flexibility (or weakening prior knowledge) by enriching the hypothesis space and also allowing for non-axis-parallel rectangles, none of the queries can be classified with certainty anymore. In the second example (below), hypotheses are linear and quadratic discriminant functions, respectively. Bottom left: If the hypothesis space is given by linear discriminant functions, all hypothesis consistent with the data will predict the query instance as positive. Bottom right: Enriching the hypothesis space with quadratic discriminants, the members of the version space will no longer vote unanimously: some of them predict the positive and others the negative class.		
}
		\label{fig:vsli}
\end{figure}

Note that the inference (\ref{eq:ee1}) can be seen as a kind of constraint propagation, in which the constraint $h \in \mathcal{V}$ on the hypothesis space $\cH$ is propagated to a constraint on $\cY$, expressed in the form of the subset (\ref{eq:vss}) of possible outcomes; or symbolically:
\begin{equation}\label{eq:cbi}
\cH , \cD , \vec{x}_{q}  \models Y
\end{equation}
This view highlights the interaction between prior knowledge and data: It shows that what can be said about the possible outcomes $y_{q}$ not only depends on the data $\cD$ but also on the hypothesis space $\cH$, i.e., the \emph{model assumptions} the learner starts with.
The specification of $\mathcal{H}$ always comes with an \emph{inductive bias}, which is indeed essential for learning from data \citep{mitc_tn80}. 
In general, both aleatoric and epistemic uncertainty (ignorance) depend on the way in which prior knowledge and data interact with each other. Roughly speaking, the stronger the knowledge the learning process starts with, the less data is needed to resolve uncertainty. In the extreme case, the true model is already known, and data is completely superfluous. Normally, however, prior knowledge is specified by assuming a certain type of model, for example a linear relationship between inputs $\vec{x}$ and outputs $y$. Then, all else (namely the data) being equal, the degree of predictive uncertainty depends on how flexible the corresponding model class is. Informally speaking, the more restrictive the model assumptions are, the smaller the uncertainty will be. This is illustrated in Fig.~\ref{fig:vsli} for the case of binary classification.


Coming back to our discussion about uncertainty, it is clear that version space learning as outlined above does not involve any kind of aleatoric uncertainty. Instead, the only source of uncertainty is a lack of knowledge about $h^*$, and hence of epistemic nature. On the model level, the amount of uncertainty is in direct correspondence with the size of the version space $\mathcal{V}$ and reduces with an increasing sample size. Likewise, the predictive uncertainty could be measured in terms of the size of the set (\ref{eq:vss}) of candidate outcomes. Obviously, this uncertainty may differ from instance to instance, or, stated differently, approximation uncertainty may translate into prediction uncertainty in different ways.

In version space learning, uncertainty is represented in a purely set-based manner: the version space $\mathcal{V}$ and prediction set $Y(\vec{x}_q)$ are subsets of $\mathcal{H}$ and $\mathcal{Y}$, respectively. In other words, hypotheses $h \in \mathcal{H}$ and outcomes $y \in \mathcal{Y}$ are only qualified in terms of being possible or not. In the following, we discuss the Bayesian approach, in which hypotheses and predictions are qualified more gradually in terms of probabilities.  




\subsection{Bayesian inference}
\label{sec:bi}

\begin{figure}
\begin{center}
\includegraphics[scale=0.45]{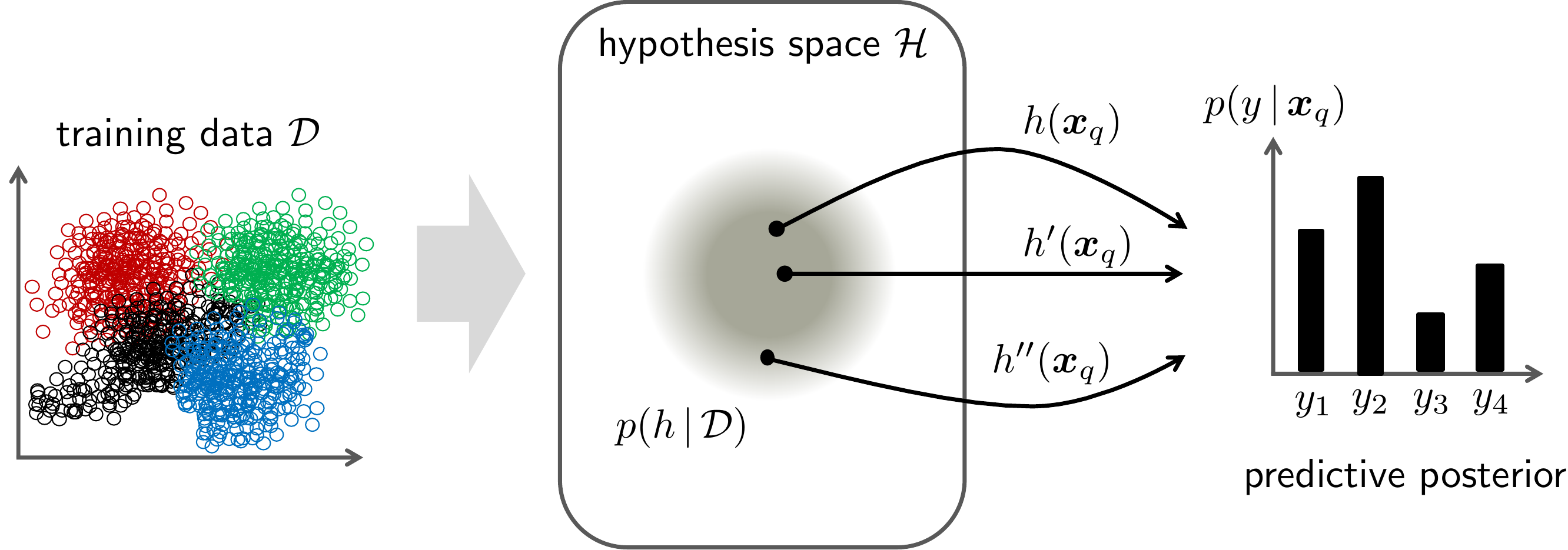} 
\caption{Illustration of Bayesian inference. Updating a prior distribution on $\mathcal{H}$ based on training data $\mathcal{D}$, a posterior distribution $\prob(h \given \mathcal{D})$ is obtained, which represents the state of knowledge of the learner (middle). Given a query $\vec{x}_{q} \in \mathcal{X}$, the predictive posterior distribution on $\mathcal{Y}$ is then obtained through Bayesian model averaging: Different hypotheses $h \in \mathcal{H}$ provide predictions, which are aggregated in terms of a weighted average.}
\label{fig:bayesian}
\end{center}
\end{figure}

Consider a hypothesis space $\cH$ consisting of probabilistic predictors, that is, hypotheses $h$ that deliver probabilistic predictions $\prob_h(y \given \vec{x}) = \prob(y \given \vec{x}, h)$ of outcomes $y$ given an instance $\vec{x}$. In the Bayesian approach, $\cH$ is supposed to be equipped with a prior distribution $\prob(\cdot)$, and learning essentially consists of replacing the prior by the posterior distribution:
\begin{equation}\label{eq:bpost}
\prob(h \given \cD) \, = \frac{\prob(h) \cdot \prob(\cD \given h)}{\prob(\cD)} \,  \propto \, \prob(h) \cdot \prob(\cD \given h) \, ,
\end{equation}
where $\prob(\cD \given h)$ is the probability of the data given $h$ (the likelihood of $h$). 
Intuitively, $\prob( \cdot \given \cD)$ captures the learner's state of knowledge, and hence its epistemic uncertainty: The more ``peaked'' this distribution, i.e., the more concentrated the probability mass in a small region in $\mathcal{H}$, the less uncertain the learner is. Just like the version space $\mathcal{V}$ in version space learning, the posterior distribution on $\mathcal{H}$ provides global (averaged/aggregated over the entire instance space) instead of local, \emph{per-instance} information. For a given query instance $\vec{x}_{q}$, this information may translate into quite different representations of the uncertainty regarding the prediction $\haty_{q}$ (cf.\ Fig.\ \ref{fig:bayesian}).

More specifically, the representation of uncertainty about a prediction $\haty_{q}$ is given by the image of the posterior $\prob( h \given \cD)$ under the mapping $h \mapsto \prob(y \given \vec{x}_{q}, h)$ from hypotheses to probabilities of outcomes. This yields the predictive posterior distribution
\begin{equation}\label{eq:pd}
\prob(y \given \vec{x}_{q}) = 
\int_\cH \prob(y \given \vec{x}_{q} , h) \, d\, \Prob(h \given \set{D}) \enspace .
\end{equation}
In this type of (proper) Bayesian inference, a final prediction is thus produced by \emph{model averaging}: The predicted probability of an outcome $y \in \cY$ is the \emph{expected} probability $\prob(y \given \vec{x}_{q}, h)$, where the expectation over the hypotheses is taken with respect to the posterior distribution on $\cH$; or, stated differently, the probability of an outcome is a weighted average over its probabilities under all hypotheses in $\cH$, with the weight of each hypothesis $h$ given by its posterior probability $\prob( h \given \cD)$.
Since model averaging is often difficult and computationally costly, sometimes only the single hypothesis
\begin{equation}\label{eq:map}
h^{map} \defeq \argmax_{h \in \cH} \prob( h \given \cD) 
\end{equation}
with the highest posterior probability is adopted, and predictions on outcomes are derived from this hypothesis. 

In (\ref{eq:pd}), aleatoric and epistemic uncertainty are not distinguished any more, because epistemic uncertainty is ``averaged out.'' Consider the example of coin flipping, and let the hypothesis space be given by $\mathcal{H} \defeq \{ h_\alpha \with 0 \leq \alpha \leq 1 \}$, where $h_\alpha$ is modeling a biased coin landing heads up with a probability of $\alpha$ and tails up with a probability of $1- \alpha$. According to (\ref{eq:pd}), we derive a probability of $1/2$ for heads and tails, regardless of whether the (posterior) distribution on $\mathcal{H}$ is given by the uniform distribution (all coins are equally probable, i.e., the case of complete ignorance) or the one-point measure assigning probability 1 to $h_{1/2}$ (the coin is known to be fair with complete certainty):
\begin{equation*} 
\prob(y ) = 
\int_\cH \alpha \, d\, \Prob  = \frac{1}{2} = \int_\cH \alpha \, d\, \Prob'  
\end{equation*}
for the uniform measure $\Prob$ (with probability density function $\prob(\alpha) \equiv 1$) as well as the measure $\Prob'$ with probability mass function $\prob'(\alpha) = 1$ if $\alpha = \nicefrac{1}{2}$ and $= 0$ for $\alpha \neq \nicefrac{1}{2}$. 
Obviously, MAP inference (\ref{eq:map}) does not capture epistemic uncertainty either. 

More generally, consider the case of binary classification with $\cY \defeq \{ -1, +1 \}$ and $p_h(+1 \given \vec{x}_{q})$ the probability predicted by the hypothesis $h$ for the positive class. Instead of deriving a distribution on $\mathcal{Y}$ according to (\ref{eq:pd}), one could also derive a predictive distribution for the (unknown) probability $q \defeq \prob(+1 \given \vec{x}_q)$ of the positive class:
\begin{equation}\label{eq:sop}
\prob(q \given \vec{x}_{q}) = 
\int_{\mathcal{H}} \, \llbracket \, \prob(+1 \given \vec{x}_{q} , h) = q \, \rrbracket \, d\, \Prob(h \given \mathcal{D}) \enspace .
\end{equation}
This is a second-order probability, which still contains both aleatoric and epistemic uncertainty. 
The question of how to quantify the epistemic part of the uncertainty is not at all obvious, however. Intuitively, epistemic uncertainty should be reflected by the variability of the distribution (\ref{eq:sop}): the more spread the probability mass over the unit interval, the higher the uncertainty seems to be. But how to put this into quantitative terms? Entropy is arguably not a good choice, for example, because this measure is invariant against redistribution of probability mass. For instance, the distributions $\prob$ and $\prob'$ with
$\prob(q \given \vec{x}_{q}) \defeq 12(q-\nicefrac{1}{2})^2$ and $\prob'(q \given \vec{x}_{q}) \defeq 3 - \prob(q \given \vec{x}_{q})$ both have the same entropy, although they correspond to quite different states of information. 
From this point of view, the variance of the distribution would be better suited, but this measure has other deficiencies (for example, it is not maximized by the uniform distribution, which could be considered as a case of minimal informedness).  

The difficulty of specifying epistemic uncertainty in the Bayesian approach is rooted in the more general difficulty of representing a lack of knowledge in probability theory. This issue will be discussed next.


\subsection{Representing a lack of knowledge}
\label{sec:rlk}



As already said, uncertainty is captured in a purely \emph{set-based} way in version space learning: $\mathcal{V} \subseteq \mathcal{H}$ is a set of candidate hypotheses, which translates into a set of candidate outcomes $Y \subseteq \mathcal{Y}$ for a query $\vec{x}_{q}$. In the case of sets, there is a rather obvious correspondence between the degree of uncertainty in the sense of a lack of knowledge and the size of the set of candidates: Proceeding from a reference set $\Omega$ of alternatives, assuming some ground-truth $\omega^* \in \Omega$, and expressing knowledge about the latter in terms of a subset $C \subseteq \Omega$ of possible candidates, we can clearly say that the bigger $C$, the larger the lack of knowledge. More specifically, for finite $\Omega$, a common uncertainty measure in information theory is $\log(|C|)$. Consequently, knowledge gets weaker by adding additional elements to $C$ and stronger by removing candidates. 

In standard probabilistic modeling and Bayesian inference, where knowledge is conveyed by a distribution $p$ on $\Omega$, it is much less obvious how to ``weaken'' this knowledge. This is mainly because the total amount of belief is fixed in terms of a unit mass that can be distributed among the elements $\omega \in \Omega$. Unlike for sets, where an additional candidate $\omega$ can be added or removed without changing the plausibility of all other candidates, increasing the weight of one alternative $\omega$ requires decreasing the weight of another alternative $\omega'$ by exactly the same amount. 

Of course, there are also measures of uncertainty for probability distributions, most notably the (Shannon) entropy, which, for finite $\Omega$, is given as follows:
\begin{equation}\label{eq:shannon}
H(\prob) \defeq - \sum_{\omega \in \Omega} p(\omega) \, \log \prob(\omega)  
\end{equation}
However, these measures are primarily capturing the shape of the distribution, namely its ``peakedness'' or non-uniformity \citep{mpub128}, and hence inform about the predictability of the outcome of a random experiment. Seen from this point of view, they are more akin to aleatoric uncertainty, whereas the set-based approach (i.e., representing knowledge in terms of a subset $C \subseteq \Omega$ of candidates) is arguably better suited for capturing epistemic uncertainty. 

For these reasons, it has been argued that probability distributions are less suitable for representing \emph{ignorance}\mmp{ignorance} in the sense of a lack of knowledge \citep{dubo_rp96}. 
For example, the case of \emph{complete ignorance} is typically modeled in terms of the uniform distribution $\prob \equiv 1/|\Omega|$ in probability theory; this is justified by the ``principle of indifference'' invoked by Laplace, or by referring to the principle of maximum entropy\footnote{Obviously, there is a technical problem in defining the uniform distribution in the case where $\Omega$ is not finite.}. Then, however, it is not possible to distinguish between (i) precise (probabilistic) knowledge about a random event, such as tossing a fair coin, and (ii) a complete lack of knowledge due to an incomplete description of the experiment.
This was already pointed out by the famous Ronald Fisher, who noted that ``\emph{not knowing the chance of mutually exclusive events and knowing the chance to be equal are two quite different states of knowledge}.''  

Another problem in this regard is caused by the measure-theoretic grounding of probability and its additive nature. For example, the uniform distribution is not invariant under reparametrization (a uniform distribution on a parameter $\omega$ does not translate into a uniform distribution on $1/\omega$, although ignorance about $\omega$ implies ignorance about $1/\omega$). Likewise, expressing ignorance about the length $x$ of a cube in terms of a uniform distribution on an interval $[l,u]$ does not yield a uniform distribution of $x^3$ on $[l^3 , u^3]$, thereby suggesting some degree of informedness about its volume. Problems of this kind render the use of a uniform prior distribution, often interpreted as representing epistemic uncertainty in Bayesian inference, at least debatable\footnote{This problem is inherited by hierarchical Bayesian modeling. See work on ``non-informative'' priors, however \citep{jeff_46,bern_rp79}.}.

\begin{figure}
    \centering
    \includegraphics[width=0.9\linewidth]{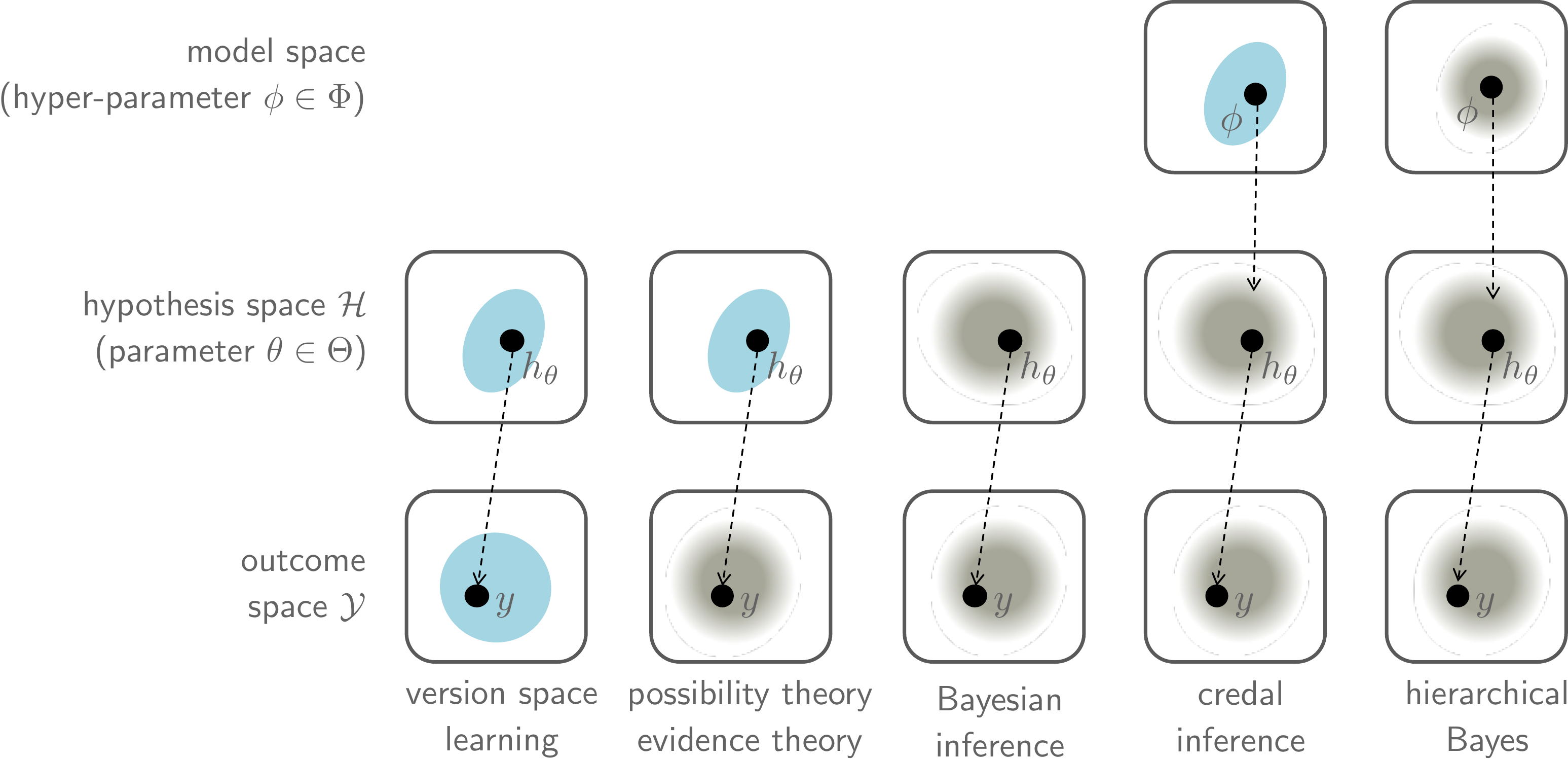}
        \caption{Set-based versus distributional knowledge representation on the level of predictions, hypotheses, and models. In version space learning, the model class is fixed, and knowledge about hypotheses and outcomes is represented in terms of sets (blue color). In Bayesian inference, sets are replaced by probability distributions (gray shading). Theories like possibility and evidence theory are in-between, essentially working with sets of distributions. Credal inference (cf.\ Section \ref{sec:credal}) generalizes Bayesian inference by replacing a single prior with a set of prior distributions (here identified by a set of hyper-parameters). In hierarchical Bayesian modeling, this set is again replaced by a distribution, i.e., a second-order probability.}
    \label{fig:setting2}
    \end{figure}

The argument that a single (probability) distribution is not enough for representing uncertain knowledge is quite prominent in the literature. Correspondingly, various generalizations of standard probability theory have been proposed, including imprecise probability \citep{wall_sr}, evidence theory \citep{shaf_am,smet_tt94}, and possibility theory \citep{dubo_pt}. These formalisms are essentially working with sets of probability distributions, i.e., they seek to take advantage of the complementary nature of sets and distributions, and to combine both representations in a meaningful way (cf.\ Fig.\ \ref{fig:setting2}). We also refer to Appendix \ref{app:unc} for a brief overview.

\section{Machine learning methods for representing uncertainty}
\label{sec:methods}

This section presents several important machine learning methods that allow for representing the learner's uncertainty in a prediction. They differ with regard to the type of prediction produced and the way in which uncertainty is represented. Another interesting question is whether they allow for distinguishing between aleatoric and epistemic uncertainty, and perhaps even for quantifying the amount of uncertainty in terms of degrees of aleatoric and epistemic (and total) uncertainty.

\begin{figure}
    \centering
    \includegraphics[width=0.95\linewidth]{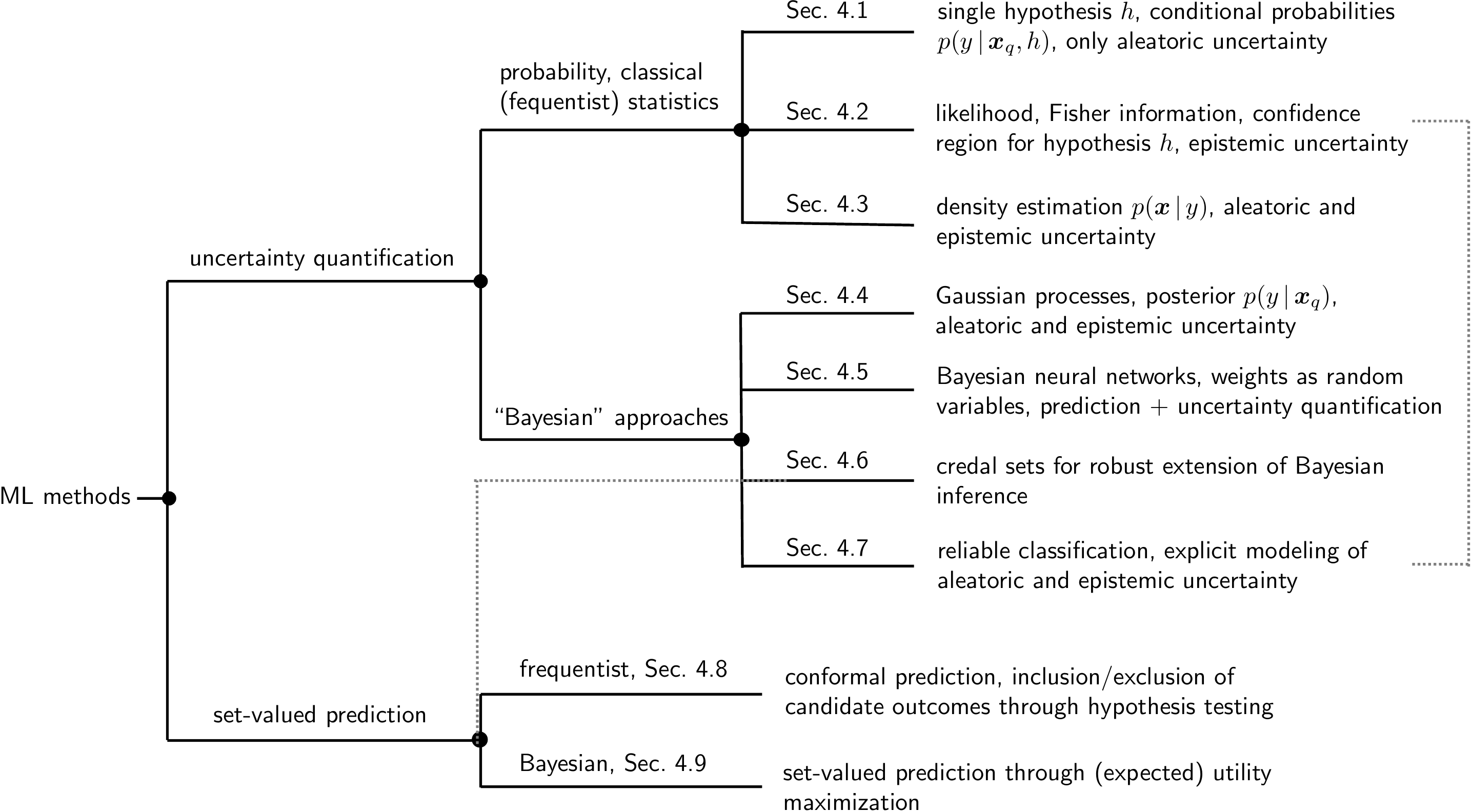}
        \caption{Overview of the methods presented in Section \ref{sec:methods}.}
    \label{fig:overview}
    \end{figure}

Structuring the methods or arranging them in the form of a taxonomy is not an easy task, and there are various possibilities one may think of. As shown in the overview in Fig.\ \ref{fig:overview}, a possible distinction one may think of is whether a method is rooted in classical, frequentist statistics (building on methods for maximum likelihood inference and parameter estimation like in Section \ref{sec:fisher}, density estimation like in Section \ref{sec:generative}, or hypothesis testing like in Section \ref{sec:cp}), or whether it is more inspired by Bayesian inference. Another distinction that can be made is between uncertainty quantification and set-valued prediction: Most of the methods produce a prediction and equip this prediction with additional information about the certainty or uncertainty in that prediction. The other way around, one may of course also pre-define a desired level of certainty, and then produce a prediction that complies with this requirement\,---\,naturally, such a prediction appears in the form of a set of candidates that is sufficiently likely to contain the ground truth.   

The first three sections are related to classical statistics and probability estimation. Machine learning methods for probability estimation (Section \ref{sec:prob}), i.e., for training a probabilistic predictor, often commit to a single hypothesis learned on the data, thereby ignoring epistemic uncertainty. Instead, predictions obtained by this hypothesis essentially represent aleatoric uncertainty. As opposed to this, Section \ref{sec:fisher} discusses likelihood-based inference and Fisher information, which allows for deriving confidence regions that reflect epistemic uncertainty about the hypothesis. The focus here (like in frequentist statistics in general) is more on parameter estimation and less on prediction. Section \ref{sec:generative} presents methods for learning generative models, which essentially comes down to the problem of density estimation. Here, predictions and uncertainty quantification are produced on the basis of class-conditional density functions.

The second block of methods is more in the spirit of Bayesian machine learning. Here, knowledge and uncertainty about the sought (Bayes-optimal) hypothesis is maintained explicitly in the form of a distribution on the hypothesis space (Sections \ref{sec:gp} and \ref{sec:m1}), a set of distributions (Section \ref{sec:credal}), or a graded set of distributions (Section \ref{sec:uqnl}), and this knowledge is updated in the light of additional observations (see also Fig.\ \ref{fig:setting2}). Correspondingly, all these methods provide information about both aleatoric and epistemic uncertainty. Perhaps the most well-known approach in the realm of Bayesian machine learning is Gaussian processes, which are discussed in Section \ref{sec:gp}. Another popular approach is deep learning and neural networks, and especially interesting from the point of view of uncertainty quantification is recent work on ``Bayesian'' extensions of such networks. Section \ref{sec:credal} addresses the concept of credal sets, which is closely connected to the theory of imprecise probabilities, and which can be seen as a generalization of Bayesian inference where a single prior distribution is replaced by a set of priors. 
Section \ref{sec:uqnl} presents an approach to reliable prediction that combines set-based and distributional (probabilistic) inference, and which, to the best of our knowledge, was the first to explicitly motivate the distinction between aleatoric and epistemic uncertainty in a machine learning context.

Finally, methods focusing on set-valued prediction are discussed in Sections \ref{sec:cp} and \ref{sec:sbus}. Conformal prediction is rooted in frequentist statistics and hypothesis testing. Roughly speaking, to decide whether to include or exclude a candidate prediction, the correctness of that prediction is tested as a hypothesis. The second approach makes use of concepts from statistical (Bayesian) decision theory and constructs set-valued predictions based on the principle of expected utility maximization.  

As already said, the categorization of methods is certainly not unique. For example, as the credal approach is of a ``set-valued'' nature (starting with a set of priors, a set of posterior distributions is derived, and from these in turn a set of predictions), it has a strong link to set-valued prediction (as indicated by the dashed line in Fig.\ \ref{fig:overview}). Likewise, the approach to reliable classification in Section \ref{sec:uqnl} makes use of the concept of normalized likelihood, and hence has a strong connection to likelihood-based inference.

\subsection{Probability estimation via scoring, calibration, and ensembling}
\label{sec:prob}

There are various methods in machine learning for inducing probabilistic predictors. These are hypotheses $h$ that do not merely output point predictions $h(\vec{x}) \in \mathcal{Y}$, i.e., elements of the output space $\cY$, but probability estimates $\prob_h(\cdot \given \vec{x}) =  \prob(\cdot \given \vec{x}, h)$, i.e., complete probability distributions on $\cY$. In the case of classification, this means predicting a single (conditional) probability $\prob_h(y \given \vec{x}) = \prob(y \given \vec{x} , h)$ for each class $y \in \mathcal{Y}$, whereas in regression, $\prob( \cdot \given \vec{x}, h)$ is a density function on $\mathbb{R}$. Such predictors can be learned in a discriminative way, i.e., in the form of a mapping $\vec{x} \mapsto \prob( \cdot \given \vec{x})$, or in a generative way, which essentially means learning a joint distribution on $\mathcal{X} \times \mathcal{Y}$. Moreover, the approaches can be parametric (assuming specific parametric families of probability distributions) or non-parametric. Well-known examples include classical statistical methods such as logistic and linear regression, Bayesian approaches such as Bayesian networks and Gaussian processes (cf.\ Section \ref{sec:gp}), as well as various techniques in the realm of (deep) neural networks (cf.\ Section \ref{sec:m1}). 

Training probabilistic predictors is typically accomplished by minimizing suitable loss functions, i.e., loss functions that enforce ``correct'' (conditional) probabilities as predictions. In this regard, proper scoring rules \citep{gnei_sp05} play an important role, including the log-loss as a well-known special case. Sometimes, however, estimates are also obtained in a very simple way, following basic frequentist techniques for probability estimation, like in Na\"ive Bayes or nearest neighbor classification. 

The predictions delivered by corresponding methods are at best ``pseudo-probabilities'' that are often not very accurate. Besides, there are many methods that deliver natural scores, intuitively expressing a degree of confidence (like the distance from the separating hyperplane in support vector machines), but which do not immediately qualify as probabilities either. The idea of \emph{scaling} or \emph{calibration methods} is to turn such scores into proper, well-calibrated probabilities, that is, to learn a mapping from scores to the unit interval that can be applied to the output of a predictor as a kind of post-processing step \citep{flac_cc17}. Examples of such methods include binning \citep{zadr_oc01}, isotonic regression \citep{zadr_tc02}, logistic scaling \citep{Pla00} and improvements thereof \citep{kull_bc17}, as well as the use of Venn predictors \citep{joha_vp18}. Calibration is still a topic of ongoing research. 

Another import class of methods is \emph{ensemble learning}, such as bagging or boosting, which are especially popular in machine learning due to their ability to improve accuracy of (point) predictions. 
Since such methods produce a (large) set of predictors $h_1, \ldots, h_M$ instead of a single hypothesis, it is tempting to produce probability estimates following basic frequentist principles. In the simplest case (of classification), each prediction $h_i(\vec{x})$ can be interpreted as a ``vote'' in favor of a class $y \in \mathcal{Y}$, and probabilities can be estimated by relative frequencies\,---\,needless to say, probabilities constructed in this way tend to be biased and are not necessarily well calibrated. Especially important in this field are tree-based methods such as random forests \citep{brei_rf01,krup_pe14}.  

Obviously, while standard probability estimation is a viable approach to representing uncertainty in a prediction, there is no explicit distinction between different types of uncertainty. Methods falling into this category are mostly concerned with the aleatoric part of the overall uncertainty.\footnote{Yet, as will be seen later on, one way to go beyond mere aleatoric uncertainty is to combine the above methods, for example learning ensembles of probabilistic predictors (cf.\ Section \ref{sec:m1}).}

\subsection{Maximum likelihood estimation and Fisher information}
\label{sec:fisher}


The notion of likelihood is one of the key elements of statistical inference in general, and maximum likelihood estimation is an omnipresent principle in machine learning as well. Indeed, many learning algorithms, including those used to train deep neural networks, realize model induction as likelihood maximization. Besides, there is a close connection between likelihood-based and Bayesian inference, as in many cases, the former is effectively equivalent to the latter with a uniform prior\,---\,disregarding, of course, important conceptual differences and philosophical foundations.  


Adopting the common perspective of classical (frequentist) statistics, consider a data-generating process specified by a parametrized family of probability measures $\Prob_{\vec{\theta}}$, where $\vec{\theta} \in \Theta$ is a parameter vector. Let $f_{\vec{\theta}}( \cdot )$ denote the density (or mass) function of $\Prob_{\vec{\theta}}$, i.e., $f_{\vec{\theta}}( X)$ is the probability to observe the data $X$ when sampling from $\Prob_{\vec{\theta}}$. An important problem in statistical inference is to estimate the parameter $\vec{\theta}$, i.e., to identify the underlying data-generating process $\Prob_{\vec{\theta}}$, on the basis of a set of observations $\mathcal{D} = \{ X_1, \ldots , X_N \}$, and maximum likelihood estimation (MLE) is a general principle for tackling this problem. More specifically, the MLE principle prescribes to estimate $\vec{\theta}$ by the maximizer of the likelihood function, or, equivalently, the log-likelihood function. Assuming that $X_1, \ldots , X_N$ are independent, and hence that $\mathcal{D}$ is distributed according to $( \Prob_{\vec{\theta}})^N$, the log-likelihood function is given by
$$
\ell_N(\vec{\theta}) \defeq \sum_{n=1}^N \log f_{\vec{\theta}}(X_n ) \, .
$$
An important result of mathematical statistics states that the distribution of the maximum likelihood estimate $\hat{\vec{\theta}}$ is asymptotically normal, i.e., converges to a normal distribution as $N \rightarrow \infty$. More specifically, $\sqrt{N}(\hat{\vec{\theta}} - \vec{\theta})$ converges to a normal distribution with mean 0 and covariance matrix $\mathcal{I}_N^{-1}(\vec{\theta})$, where 
$$
\mathcal{I}_N(\vec{\theta}) =  - \left[
\evalue_{\vec{\theta}} \left( \frac{\partial^2 \ell_N}{\partial \theta_i \, \partial \theta_j} \right) 
\right]_{1 \leq i,j \leq N}
$$
is the \emph{Fisher information matrix} (the negative Hessian of the log-likelihood function at $\vec{\theta})$. This result has many implications, both theoretically and practically. For example, the Fisher information plays an important role in the Cram\'er-Rao bound and provides a lower bound to the variance of any unbiased estimator of $\vec{\theta}$ \citep{frie_sf}. 

Moreover, the Fisher information matrix allows for constructing (approximate) \emph{confidence regions} for the sought parameter $\vec{\theta}$ around the estimate $\hat{\vec{\theta}}$ (approximating\footnote{Of course, the validity of this ``plug-in'' estimate requires some formal assumptions.} $\mathcal{I}_N(\vec{\theta})$ by $\mathcal{I}_N(\hat{\vec{\theta}})$). Obviously, the larger this region, the higher the (epistemic) uncertainty about the true model $\vec{\theta}$. Roughly speaking, the size of the confidence region is in direct correspondence with the ``peakedness'' of the likelihood function around its maximum. If the likelihood function is peaked, small changes of the data do not change the estimation $\hat{\vec{\theta}}$ too much, and the learner is relatively sure about the true model (data-generating process). As opposed to this, a flat likelihood function reflects a high level of uncertainty about $\vec{\theta}$, because there are many parameters that are close to  $\hat{\vec{\theta}}$ and have a similar likelihood. 

In a machine learning context, where parameters $\vec{\theta}$ may identify hypotheses $h = h_{\vec{\theta}}$, a confidence region for the former can be seen as a representation of epistemic uncertainty about $h^*$, or, more specifically, approximation uncertainty. To obtain a quantitative measure of uncertainty, the Fisher information matrix can be summarized in a scalar statistic, for example the trace (of the inverse) or the smallest eigenvalue. Based on corresponding measures, Fisher information has been used, among others, for optimal statistical design \citep{puke_od} and active machine learning \citep{sour_ap18}.

\subsection{Generative models} 
\label{sec:generative}

\begin{figure}
\begin{center}
\includegraphics[scale=0.25]{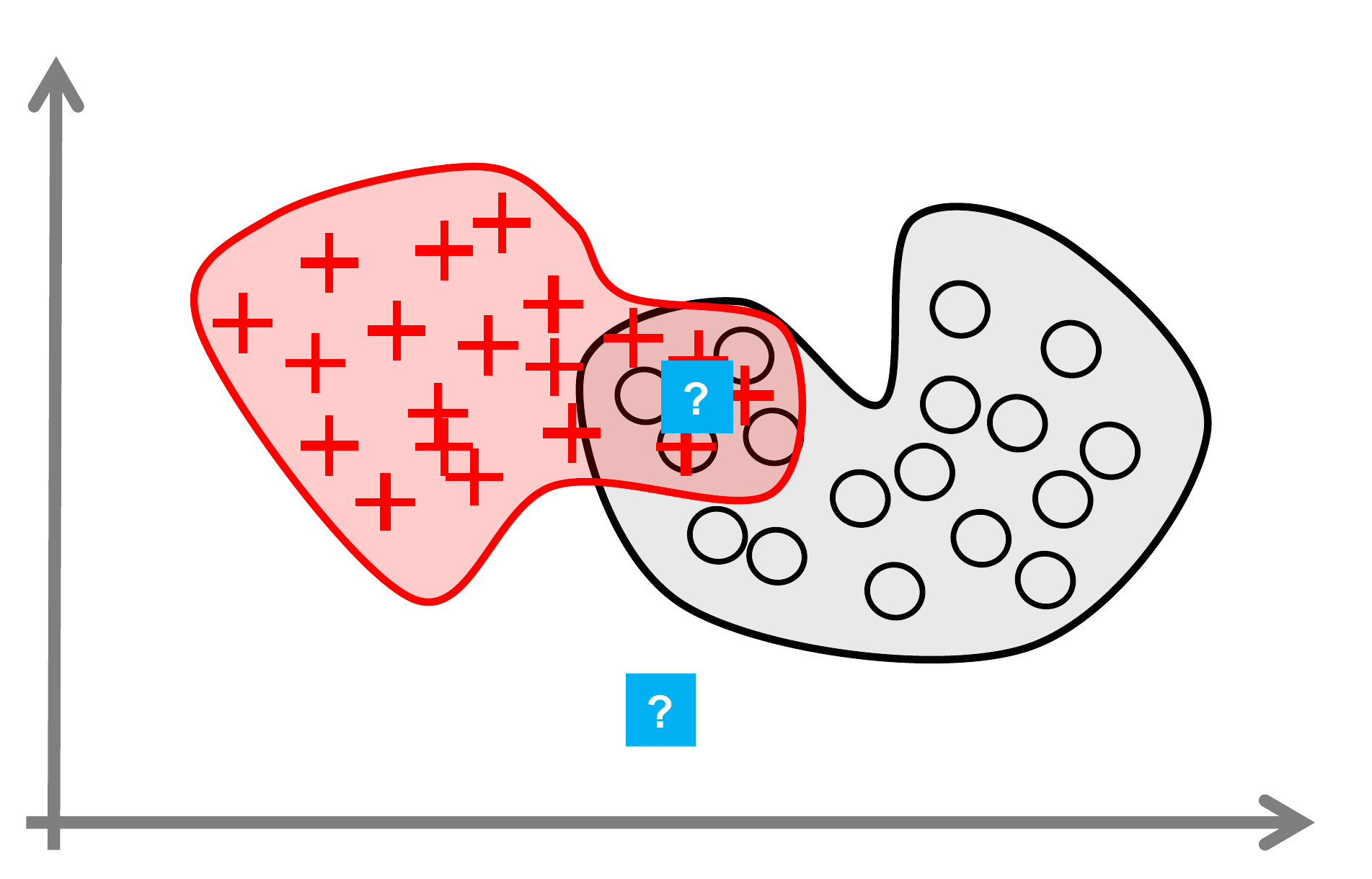} 
\includegraphics[scale=0.25]{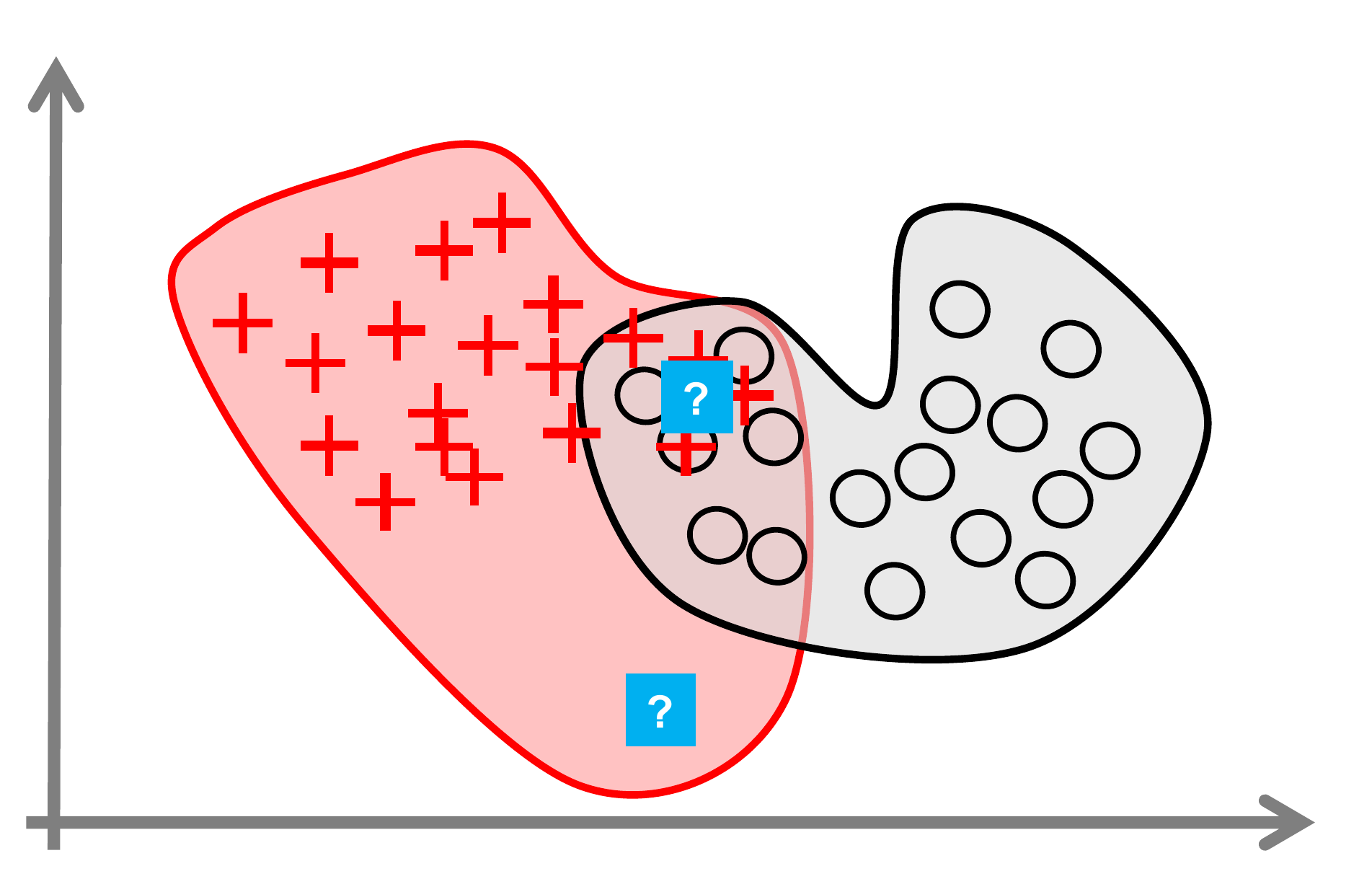} 
\includegraphics[scale=0.25]{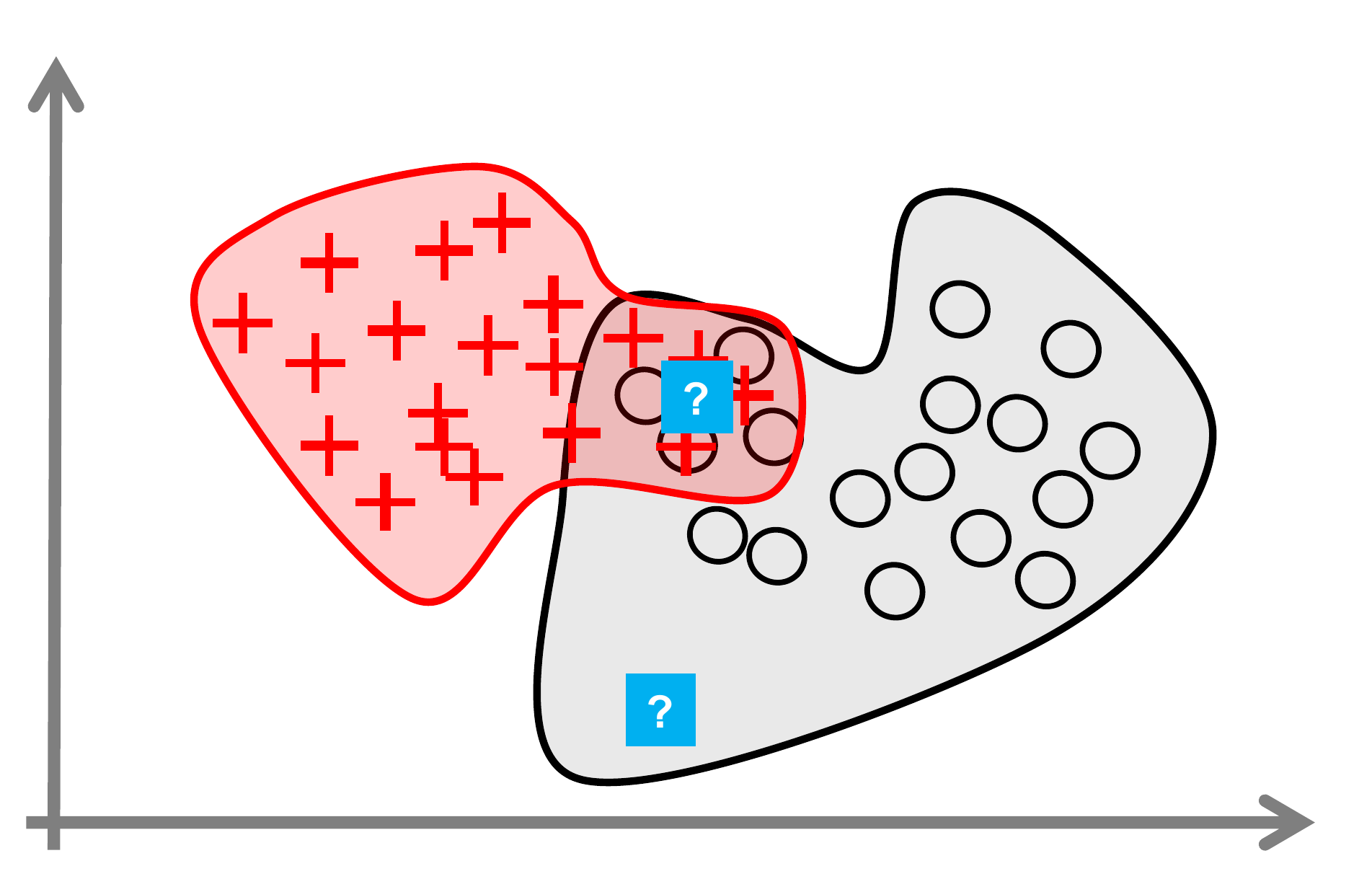}
\caption{If a hypothesis space is very rich, so that it can essentially produce every decision boundary, the epistemic uncertainty is high in sparsely populated regions without training examples. The lower query point in the left picture, for example, could easily be classified as red (middle picture) but also as black (right picture). In contrast to this, the second query is in a region where the two class distributions are overlapping, and where the aleatoric uncertainty is high.}
\label{fig:local}
\end{center}
\end{figure}

As already explained, for learning methods with a very large hypothesis space $\mathcal{H}$, model uncertainty essentially disappears, while approximation uncertainty might be high (cf.\ Section \ref{sec:aamu}). In particular, this includes learning methods with high flexibility, such as (deep) neural networks and nearest neighbors. In methods of that kind, there is no explicit assumption about a \emph{global} structure of the dependency between inputs $\vec{x}$ and outcomes $y$. Therefore, inductive inference will essentially be of a \emph{local} nature: A class $y$ is approximated by the region in the instance space in which examples from that class have been seen, aleatoric uncertainty occurs where such regions are overlapping, and epistemic uncertainty where no examples have been encountered so far (cf.\ Fig.\ \ref{fig:local}).  

An intuitive idea, then, is to consider generative models to quantify epistemic uncertainty. Such approaches typically look at the densities $\prob(\vec{x})$ to decide whether input points are located in regions with high or low density, in which the latter acts as a proxy for a high epistemic uncertainty. The density $\prob(\vec{x})$ can be estimated with traditional methods, such as kernel density estimation or Gaussian mixtures. Yet, novel density estimation methods still appear in the machine learning literature. Some more recent approaches in this area include isolation forests \citep{Liu_isolationforest}, auto-encoders \citep{Goodfellow2016}, and radial basis function networks \citep{Bazargami2019}.

As shown by these methods, density estimation is also a central building block in anomaly and outlier detection. Often, a threshold is applied on top of the density $\prob(\vec{x})$ to decide whether a data point is an outlier or not. For example, in auto-encoders, a low-dimensional representation of the original input in constructed, and the reconstruction error can be used as a measure of support of a data point within the underlying distribution.  Methods of that kind can be classified as semi-supervised outlier detection methods, in contrast to supervised methods, which use annotated outliers during training. Many semi-supervised outlier detection methods are inspired by one-class support vector machines (SVMs), which fit a hyperplane that separates outliers from ``normal'' data points in a high-dimensional space \citep{Khan_2014}. Some variations exist, where for example a hypersphere instead of a hyperplane is fitted to the data \citep{Tax2004}.

Outlier detection is also somewhat related to the setting of classification with a reject option, e.g., when the classifier refuses to predict a label in low-density regions \citep{pere_bc16}. For example, \citet{ziyin2019deep} adopt this viewpoint in an optimization framework where an outlier detector and a standard classifier are jointly optimized. Since a data point is rejected if it is an outlier, the focus is here on epistemic uncertainty. In contrast, most papers on classification with reject option employ a reasoning on conditional class probabilities $\prob(y \given \vec{x})$, using specific utility scores. These approaches rather capture aleatoric uncertainty (see also Section~\ref{sec:sbus}), showing that seemingly related papers adopting the same terminology can have different notions of uncertainty in mind. 

In recent papers, outlier detection is often combined with the idea of set-valued prediction (to be further discussed in Sections \ref{sec:credal} and \ref{sec:cp}). Here, three scenarios can occur for a multi-class classifier: 
(i) A single class is predicted. In this case, the epistemic and aleatoric uncertainty are both assumed to be sufficiently low. 
(ii) The ``null set'' (empty set) is predicted, i.e., the classifier abstains when there is not enough support for making a prediction. Here, the epistemic uncertainty is too high to make a prediction.  
(iii) A set of cardinality bigger than one is predicted. Here, the aleatoric uncertainty is too high, while the epistemic uncertainty is assumed to be sufficiently low.   

\citet{hech_cd19} implement this idea by fitting a generative model $\prob(\vec{x}\given y)$ per class, and predicting null sets for data points that have a too low density for any of those generative models. A set of cardinality one is predicted when the data point has a sufficiently high density for exactly one of the models $\prob(\vec{x} \given y)$, and a set of higher cardinality is returned when this happens for more than one of the models. 
\citet{feng_sp19} propose a different framework with the same reasoning in mind. Here, a pair of models is fitted in a joint optimization problem. The first model acts as an outlier detector that intends to reject as few instances as possible, whereas the second model optimizes a set-based utility score. The joint optimization problem is therefore a linear combination of two objectives that capture, respectively, components of epistemic and aleatoric uncertainty. 

Generative models often deliver intuitive solutions to the quantification of epistemic and aleatoric uncertainty. In particular, the distinction between the two types of uncertainty explained above corresponds to what \cite{mpub169} called ``conflict'' (overlapping distributions, evidence in favor of more than one class) and ``ignorance'' (sparsely populated region, lack of evidence in favor of any class). Yet, like essentially all other methods discussed in this section, they also have disadvantages. An inherent problem with semi-supervised outlier detection methods is how to define the threshold that decides whether a data point is an outlier or not. Another issue  is how to choose the model class for the generative model. Density estimation is a difficult problem, which, to be successful, typically requires a lot of data. When the sample size is small, specifying the right model class is not an easy task, so the model uncertainty will typically be high.


\subsection{Gaussian processes}
\label{sec:gp}

Just like in likelihood-based inference, the hypothesis space in Bayesian inference (cf.\ Section \ref{sec:bi}) is often parametrized in the sense that each hypothesis $h = h_{\vec{\theta}} \in \cH$ is (uniquely) identified by a parameter (vector) $\vec{\theta} \in \Theta \subseteq \mathbb{R}^d$ of fixed dimensionality $d$. Thus, computation of the posterior (\ref{eq:bpost}) essentially comes down to updating beliefs about the true (or best) parameter, which is treated as a multivariate random variable:
\begin{equation}\label{eq:theta}
\prob(\vec{\theta} \given \cD) \,  \propto \, \prob(\vec{\theta}) \cdot \prob(\cD \given \vec{\theta}) .
\end{equation}
Gaussian processes \citep{seeg_gp04} generalize the Bayesian approach from inference about multivariate (but finite-dimensional) random variables to inference about (infinite-di\-men\-si\-onal) functions. Thus, they can be thought of as
distributions not just over random vectors but over random functions. 

More specifically, a stochastic process in the form of a collection of random variables $\{ f(\vec{x}) \given \vec{x} \in \mathcal{X} \}$ with index set $\cX$ is said to be drawn from a Gaussian process with mean function $m$ and covariance function $k$, denoted $f \sim \mathcal{GP}(m, k)$, if for any finite set of elements $\vec{x}_1, \ldots , \vec{x}_m \in \mathcal{X}$, the associated finite set of random variables $f(\vec{x}_1), \ldots , f(\vec{x}_m)$ has the following multivariate normal distribution:
$$
\left[ \begin{matrix} f(\vec{x}_1) \\ f(\vec{x}_2) \\ \vdots \\ f(\vec{x}_m) \end{matrix} \right] \sim \mathcal{N} \left( \left[ \begin{matrix} m(\vec{x}_1) \\ m(\vec{x}_2) \\ \vdots \\ m(\vec{x}_m) \end{matrix} \right] , 
\left[ \begin{matrix} k(\vec{x}_1, \vec{x}_1) & \cdots & k(\vec{x}_1, \vec{x}_m) \\
\vdots & \ddots & \vdots \\
k(\vec{x}_m, \vec{x}_1) & \cdots & k(\vec{x}_m, \vec{x}_m) \end{matrix} \right]
\right)
$$
Note that the above properties imply
\begin{align*}
m(\vec{x}) & = \evalue( f(\vec{x}) ) \, , \\
k(\vec{x}, \vec{x}') & = \evalue \big( (f(\vec{x}) - m( \vec{x} )) (f(\vec{x}') - m( \vec{x}' )) \big) \, ,
\end{align*}
and that $k$ needs to obey the properties of a kernel function (so as to guarantee proper covariance matrices). 
Intuitively, a function $f$ drawn from a Gaussian process prior can be thought of as a (very) high-dimensional vector drawn from a (very) high-dimensional multivariate
Gaussian. Here, each dimension of the Gaussian corresponds to an element $\vec{x}$ from the index set $\mathcal{X}$, and the corresponding component of the random vector represents the value of $f(\vec{x})$.

Gaussian processes allow for doing proper Bayesian inference (\ref{eq:bpost}) in a non-parametric way: Starting with a prior distribution on functions $h \in \cH$, specified by a mean function $m$ (often the zero function) and kernel $k$, this distribution can be replaced by a posterior in light of observed data $\cD = \{ (\vec{x}_i , y_i ) \}_{i=1}^N$, where an observation $y_i = f(\vec{x}_i) + \epsilon_i$ could be corrupted by an additional (additive) noise component $\epsilon_i$. Likewise, a posterior predictive distribution can be obtained on outcomes $y \in \cY$ for a new query $\vec{x}_{q} \in \cX$. In the general case, the computations involved are intractable, though turn out to be rather simple in the case of regression ($\cY = \mathbb{R}$) with Gaussian noise. In this case, and assuming a zero-mean prior, the posterior predictive distribution is again a Gaussian with mean $\mu$ and variance $\sigma^2$ as follows:
\begin{align*}
\mu & = K(\vec{x}_{q} , X) (K(X,X) + \sigma_\epsilon^2 I)^{-1} \vec{y} \, , \\
\sigma^2 & = K(\vec{x}_{q} , \vec{x}_{q}) + \sigma_\epsilon^2 - K(\vec{x}_{q} , X) (K(X,X) + \sigma_\epsilon^2 I)^{-1} K(X, \vec{x}_{q}) \, ,
\end{align*}
where $X$ is an $N \times d$ matrix summarizing the training inputs (the $i^{th}$ row of $X$ is given by $(\vec{x}_i)^\top$), $K(X,X)$ is the kernel matrix with entries $(K(X,X))_{i,j} = k(\vec{x}_i , \vec{x}_j)$, $\vec{y} = (y_1 , \ldots , y_N) \in \mathbb{R}^N$ is the vector of observed training outcomes, and $\sigma_\epsilon^2$ is the variance of the (additive) error term for these observations (i.e., $y_i$ is normally distributed with expected value $f(\vec{x}_i)$ and variance $\sigma_\epsilon^2$).  

Problems with discrete outcomes $y$, such as binary classification with $\cY = \{ -1, +1 \}$, are made amenable to Gaussian processes by suitable link functions, linking these outcomes with the real values $h = h(\vec{x})$ as underlying (latent) variables. For example, using the logistic link function  
$$
\prob( y \given h) = s(h) = \frac{1}{1 + \exp( - y \, h)} \, ,
$$
the following posterior predictive distribution is obtained:
$$
\prob( y = +1 \given X , \vec{y} , \vec{x}_{q}) = \int \sigma(h') \, \prob( h' \given X , \vec{y} , \vec{x}_{q}) \, d \, h' \, ,
$$
where
$$
\prob( h' \given X , \vec{y} , \vec{x}_{q}) = \int \prob( h' \given X ,  \vec{x}_{q}, \vec{h})\, \prob( \vec{h} \given X, \vec{y}) \, d \, \vec{h} \, .
$$
However, since the likelihood will no longer be Gaussian, approximate inference techniques (e.g., Laplace, expectation propagation, MCMC) will be needed.

\begin{figure}
\captionsetup[subfigure]{labelformat=empty}
	\centering
	\includegraphics[width=0.95\textwidth]{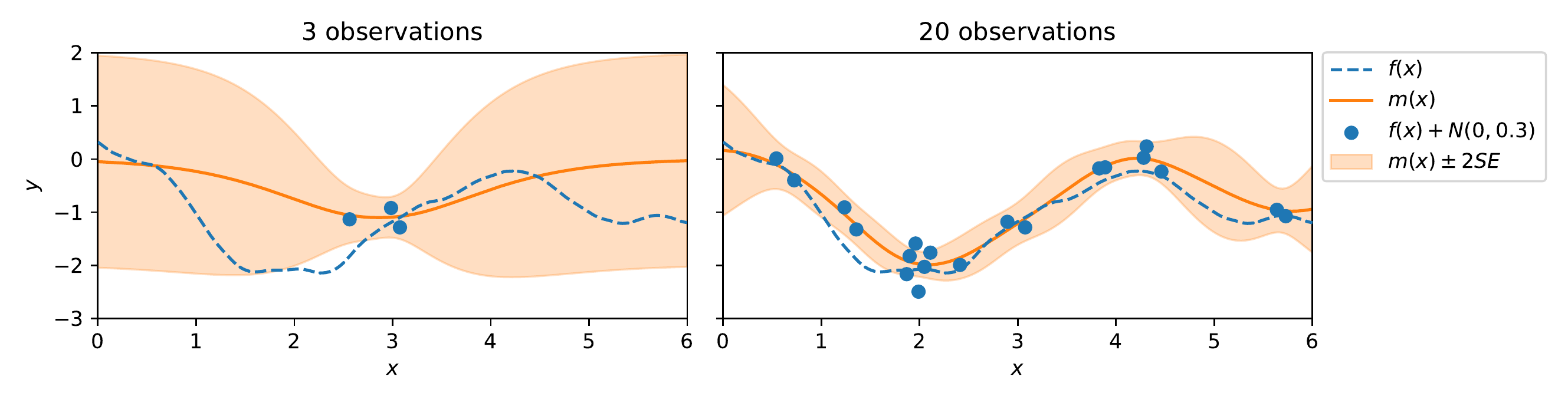}
		\caption{Simple one-dimensional illustration of Gaussian processes ($\cX = [0,6]$), with very few examples on the left and more examples on the right. The predictive uncertainty, as reflected by the width of the confidence band around the mean function (orange line) reduces with an increasing number of observations.}
		\label{fig:gp}
\end{figure}

As for uncertainty representation, our general remarks on Bayesian inference (cf.\ Section \ref{sec:bi}) obviously apply to Gaussian processes as well. In the case of regression, the variance $\sigma^2$ of the posterior predictive distribution for a query $\vec{x}_{q}$ is a meaningful indicator of (total) uncertainty. The variance of the error term, $\sigma_\epsilon^2$, corresponds to the aleatoric uncertainty, so that the difference could be considered as epistemic uncertainty. The latter is largely determined by the (parameters of the) kernel function, e.g., the characteristic length-scale of a squared exponential covariance function \citep{Blum13}. Both have an influence on the width of confidence intervals that are typically obtained for per-instance predictions, and which reflect the total amount of uncertainty (cf.\ Fig.\ \ref{fig:gp}). By determining all (hyper-) parameters of a GP, these sources of uncertainty can in principle be separated, although the estimation itself is of course not an easy task. 

\subsection{Deep neural networks}
\label{sec:m1}

Work on uncertainty in deep learning fits the general framework we introduced so far, especially the Bayesian approach, quite well, although the methods proposed are specifically tailored for neural networks as a model class. A standard neural network can be seen as a probabilistic classifier $h$: in the case of classification, given a query $\vec{x} \in \mathcal{X}$, the final layer of the network typically outputs a probability distribution (using transformations such as softmax) on the set of classes $\mathcal{Y}$, and in the case of regression, a distribution (e.g., a Gaussian\footnote{\cite{lee_dn18} offer an interesting discussion of the connection between deep networks and Gaussian processes.}) is placed over a point prediction $h(\vec{x}) \in \mathbb{R}$ (typically conceived as the expectation of the distribution). Training a neural network can essentially be seen as doing maximum likelihood inference. As such, it yields probabilistic predictions, but no information about the confidence in these probabilities. In other words, it captures aleatoric but no epistemic uncertainty.

In the context of neural networks, epistemic uncertainty is commonly understood as uncertainty about the model parameters, that is, the weights $\vec{w}$ of the neural network (which correspond to the parameter $\theta$ in (\ref{eq:theta})). Obviously, this is a special case of what we called approximation uncertainty. To capture this kind of epistemic uncertainty, Bayesian neural networks (BNNs) have been proposed as a Bayesian extension of deep neural networks \citep{denk_tn91,mack_ap92,neal_bl12}. In BNNs, each weight is represented by a probability distribution (again, typically a Gaussian) instead of a real number, and learning comes down to Bayesian inference, i.e., computing the posterior $\prob( \vec{w} \given \mathcal{D})$. The predictive distribution of an outcome given a query instance $\vec{x}_q$ is then given by
$$
\prob(y \given \vec{x}_q , \mathcal{D} ) = \int \prob( y \given \vec{x}_q , \vec{w}) \, \prob(\vec{w} \given \mathcal{D}) \, d \vec{w}  \enspace .
$$
Since the posteriors on the weights cannot be obtained analytically, approximate variational techniques are used \citep{jord_ai99,grav_pv11}, seeking a variational distribution $q_{\vec{\theta}}$ on the weights that minimizes the Kullback-Leibler divergence $\operatorname{KL}(q_{\vec{\theta}} \| \prob(\vec{w} \given \mathcal{D}))$ between $q_{\vec{\theta}}$ and the true posterior. An important example is Dropout variational inference \citep{gal_bc16}, which establishes an interesting connection between (variational) Bayesian inference and the idea of using Dropout as a learning technique for neural networks. Likewise, given a query $\vec{x}_{q}$, the predictive distribution for the outcome $y$ is approximated using techniques like Monte Carlo sampling, i.e., drawing model weights from the approximate posterior distributions (again, a connection to Dropout can be established). The total uncertainty can then be quantified in terms of common measures such as entropy (in the case of classification) or variance (in the case of regression) of this distribution. 

The total uncertainty, say the variance $\variance(\hat{y})$ in regression, still contains both, the residual error, i.e., the variance of the observation error (aleatoric uncertainty), $\sigma^2_\epsilon$, and the variance due to parameter uncertainty (epistemic uncertainty). The former can be heteroscedastic, which means that $\sigma^2_\epsilon$ is not constant but a function of $\vec{x} \in \mathcal{X}$. \cite{kend_wu17} propose a way to learn heteroscedastic aleatoric uncertainty as loss attenuation. 
Roughly speaking, the idea is to let the neural net not only predict the conditional mean of $y$ given $\vec{x}_{q}$, but also the residual error. The corresponding loss function to be minimized is constructed in such a way (or actually derived from the probabilistic
interpretation of the model) that prediction errors for points with a high residual variance are penalized less, but at the same time, a penalty is also incurred for predicting a high variance.

An explicit attempt at measuring and separating aleatoric and epistemic uncertainty (in the context of regression) is made by \cite{depe_du18}. Their idea is as follows: They quantify the total uncertainty as well as the aleatoric uncertainty, and then obtain the epistemic uncertainty as the difference. More specifically, they propose to measure the total uncertainty in terms of the entropy of the predictive posterior distribution, which, in the case of discrete $\cY$, is given by\footnote{\cite{depe_du18} actually consider the continuous case, in which case the entropy needs to be replaced by the differential entropy.}   
\begin{equation}\label{eq:eto}
H \big[ \, \prob(y \given \vec{x}) \, \big] = 
- \sum_{y \in \cY} \prob(y \given \vec{x}) \log_2 \prob(y \given \vec{x}) \, .
\end{equation}
This uncertainty also includes the (epistemic) uncertainty about the network weights $\vec{w}$. Fixing a set of weights, i.e., considering a distribution $\prob(y \given \vec{w}, \vec{x})$, thus removes the epistemic uncertainty. Therefore, the expectation over the entropies of these distributions, 
\begin{equation}\label{eq:eal}
\evalue_{p(\vec{w} \given \mathcal{D})} H \big[ \prob(y \given \vec{w} , \vec{x}) \big] = 
- \int  p( \vec{w} \given \mathcal{D}) \left( \sum_{y \in \cY} \prob(y \given \vec{w}, \vec{x}) \log_2 \prob(y \given \vec{w}, \vec{x}) \right) \, d \, \vec{w}  \enspace ,  
\end{equation} 
is a measure of the aleatoric uncertainty. Finally, the epistemic uncertainty is obtained as the difference
\begin{equation}\label{eq:mi}
u_e(\vec{x}) \defeq H \big[ \prob(y \given \vec{x}) \big] - \evalue_{p( \vec{w}\given \mathcal{D})} H \big[ \prob(y \given \vec{w}, \vec{x}) \big]  \enspace ,
\end{equation}
which equals the mutual information $I(y, \vec{w})$ between $y$ and $\vec{w}$. Intuitively, epistemic uncertainty thus captures the amount of information about the model parameters $\vec{w}$ that would be gained through knowledge of the true outcome $y$. A similar approach was recently adopted by \cite{mobi_dc19}, who also propose a technique to compute (\ref{eq:mi}) approximately.


Bayesian model averaging establishes a natural connection between Bayesian inference and ensemble learning, and indeed, as already mentioned in Section~\ref{sec:m1}, the variance of the predictions produced by an ensemble is a good indicator of the (epistemic) uncertainty in a prediction. In fact, the variance is inversely related to the ``peakedness'' of a posterior distribution $\prob(h \given \mathcal{D})$ and can be seen as a measure of the discrepancy between the (most probable) candidate hypotheses in the hypothesis space. Based on this idea, \cite{laks_sa17} propose a simple ensemble approach as an alternative to Bayesian NNs, which is easy to implement, readily parallelizable, and requires little hyperparameter tuning.

Inspired by \cite{mobi_dc19}, an ensemble approach is also proposed by \cite{mpub416}. Since this approach does not necessarily need to be implemented with neural networks as base learners\footnote{Indeed, the authors use decision trees, i.e., a random forest as an ensemble.}, we switch notation and replace weight vectors $\vec{w}$ by hypotheses $h$. 
To approximate the measures (\ref{eq:mi}) and (\ref{eq:eal}), the posterior distribution $p( h \given \mathcal{D})$ is represented by a finite ensemble of hypotheses $h_1, \ldots , h_M$. An approximation of (\ref{eq:eal}) can then be obtained by
\begin{equation}\label{eq:approxua}
u_a(\vec{x}) \defeq
- \frac{1}{M} \sum_{i=1}^M 
\sum_{y \in \cY} \prob(y \given h_i, \vec{x}) \log_2 \prob(y \given h_i, \vec{x})   \, ,
\end{equation}
an approximation of (\ref{eq:eto}) by
\begin{equation}\label{eq:approxut}
u_t(\vec{x}) \defeq  - \sum_{y \in \cY} \left( \frac{1}{M} \sum_{i=1}^M \prob(y \given h_i, \vec{x}) \right) \log_2 \left( \frac{1}{M} \sum_{i=1}^M \prob(y \given h_i, \vec{x}) \right) \, ,
\end{equation} 
and finally an approximation of (\ref{eq:mi}) by $u_e(\vec{x}) \defeq u_t(\vec{x}) - u_a(\vec{x})$. The latter corresponds to the Jensen-Shannon divergence \citep{endr_an03} of the distributions $\prob(y \given h_i, \vec{x})$, $i \in [M]$ \citep{mali_pu18}.

\subsection{Credal sets and classifiers}
\label{sec:credal}

The theory of imprecise probabilities \citep{wall_sr} is largely built on the idea to specify knowledge in terms of a set $Q$ of probability distributions instead of a single distribution. 
In the literature, such sets are also referred to as \emph{credal sets}, and often assumed to be convex (i.e., $q , q' \in Q$ implies $\alpha q + (1-\alpha) q' \in Q$ for all $\alpha \in [0,1]$). 
The concept of credal sets and related notions provide the basis of a generalization of Bayesian inference, which is especially motivated by the criticism of non-informative priors as models of ignorance in standard Bayesian inference (cf.\ Section \ref{sec:rlk}). The basic idea is to replace a single prior distribution on the model space, as used in Bayesian inference, by a (credal) set of candidate priors. Given a set of observed data, Bayesian inference can then be applied to each candidate prior, thereby producing a (credal) set of posteriors. Correspondingly, any value or quantity derived from a posterior (e.g., a prediction, an expected value, etc.) is replaced by a set of such values or quantities. An important example of such an approach is robust inference for categorical data with the \emph{imprecise Dirichlet model}, which is an extension of inference with the Dirichlet distribution as a conjugate prior for the multinomial distribution \citep{bern_ai05}. 

Methods of that kind have also been used in the context of machine learning, for example in the framework of the Na\"ive Credal Classifier \citep{zaffalon2002,cora_lr08} or more recently as an extension of sum-product networks \citep{dera_cs17}. As for the former, we can think of a Na\"ive Bayes (NB) classifier as a hypothesis $h = h_{\vec{\theta}}$ specified by a parameter vector ${\vec{\theta}}$ that comprises a prior probability $\theta_k$ for each class $y_k \in \set{Y}$ and a conditional probability $\theta_{i,j,k}$ for observing the $i^{th}$ value $a_{i,j}$ of the $j^{th}$ attribute given class $y_k$. For a given query instance specified by attributes $(a_{i_1,1}, \ldots , a_{i_J,J})$, the posterior class probabilities are then given by
\begin{equation}\label{eq:spo}
\prob(y_k \given a_{i_1,1}, \ldots , a_{i_J,J}) \propto \theta_k \prod_{j=1}^J \theta_{i_j,j,k} \, .
\end{equation}
In the Na\"ive Credal Classifier, the $\theta_k$ and $\theta_{i_j,j,k}$ are specified in terms of (local) credal sets\footnote{These could be derived from empirical data, for example, by doing generalized Bayesian inference with the imprecise Dirichlet model.}, i.e., there is a class of probability distributions $Q$ on $\set{Y}$ playing the role of candidate priors, and a class of probability distributions $Q_{j,k}$ specifying possible distributions on the attribute values of the $j^{th}$ attribute given class $y_k$. A single posterior class probability (\ref{eq:spo}) is then replaced by the set (interval) of such probabilities that can be produced by taking $\theta_k$ from a distribution in $Q$ and $\theta_{i_j,j,k}$ from a distribution in $Q_{j,k}$. As an aside, let us note that the computations involved may of course become costly.

\subsubsection{Uncertainty measures for credal sets}
\label{sec:umcs}

Interestingly, there has been quite some work on defining uncertainty measures for credal sets and related representation, such as Dempster-Shafer evidence theory \citep{klir_mo94}. Even more interestingly, a basic distinction between two types of uncertainty contained in a credal set, referred to as \emph{conflict} (randomness, discord) and \emph{non-specificity}, respectively, has been made by \cite{yage_ea83}. The importance of this distinction was already emphasized by \cite{kolm_ta65}. These notions are in direct correspondence with what we call aleatoric and epistemic uncertainty. 

The standard uncertainty measure in classical possibility theory (where uncertain information is simply represented in the form of subsets $A \subseteq \cY$ of possible alternatives) is the Hartley measure\footnote{Be aware that we denote both the Hartley measure and the Shannon entropy by $H$, which is common in the literature. The meaning should be clear from the context.}  \citep{hart_to28}
\begin{equation}\label{eq:hartley}
H(A) = \log( |A|)  \, ,
\end{equation}
Just like the Shannon entropy (\ref{eq:shannon}), this measure can be justified axiomatically\footnote{For example, see Chapter IX, pages 540--616, in the book by \cite{reny_pt}.}. 

Given the insight that conflict and non-specificity are two different, complementary sources of uncertainty, and standard (Shannon) entropy and (\ref{eq:hartley}) as well-established measures of these two types of uncertainty, a natural question in the context of credal sets is to ask for a generalized representation 
\begin{equation}\label{eq:aggregate}
\on{U}(Q) = \on{AU}(Q) + \on{EU}(Q) \, ,
\end{equation}
where $\on{U}$ is a measure of total (aggregate) uncertainty, $\on{AU}$ a measure of aleatoric uncertainty (conflict, a generalization of the Shannon entropy), and $\on{EU}$ a measure of epistemic uncertainty (non-specificity, a generalization of the Hartely measure). 

As for the non-specificity part in (\ref{eq:aggregate}), the following generalization of the Hartley measure to the case of graded possibilities has been proposed by various authors \citep{abel_an00}:
\begin{equation}\label{eq:gh}
\on{GH}(Q) \defeq  \sum_{A \subseteq \cY} \on{m}_Q(A) \, \log(|A|) \, ,
\end{equation}
where $\on{m}_Q: \,  2^{\cY} \longrightarrow [0,1]$ is the M\"obius inverse of the capacity function $\nu :\, 2^{\cY} \longrightarrow [0,1]$ defined by
\begin{equation}\label{eq:cap}
\nu_Q(A) \defeq \inf_{q \in Q} q(A) 
\end{equation}
for all $A \subseteq \cY$, that is,
$$
\on{m}_Q(A) = \sum_{B \subseteq A} (-1)^{|A \setminus B|} \nu_Q(B) \, .
$$
The measure (\ref{eq:gh}) enjoys several desirable axiomatic properties, and its uniqueness was shown by \cite{klir_ot87}.

The generalization of the Shannon entropy $H$ as a measure of conflict turned out to be more difficult. The upper and lower Shannon entropy play an important role in this regard: 
\begin{equation}\label{eq:gg}
H^*(Q) \defeq \max_{q \in Q} H(q) \, , \quad
H_*(Q) \defeq \min_{q \in Q} H(q)
\end{equation}
Based on these measures, the following disaggregations of total uncertainty (\ref{eq:aggregate}) have been proposed \citep{abel_dt06}:
\begin{align}
H^*(Q) & = \big(H^*(Q) - \on{GH}(Q) \big)  + \on{GH}(Q)  \label{eq:unc1} \\
H^*(Q) & = H_*(Q)  + \big(H^*(Q) - H_*(Q) \big)   \label{eq:unc2}
\end{align}
In both cases, upper entropy serves as a measure of total uncertainty $U(Q)$, which is again justified on an axiomatic basis. In the first case, the generalized Hartley measure is used for quantifying epistemic uncertainty, and aleatoric uncertainty is obtained as the difference between total and epistemic uncertainty. In the second case, epistemic uncertainty is specified in terms of the difference between upper and lower entropy.

\subsubsection{Set-valued prediction}

Credal classifiers are used for ``indeterminate'' prediction, that is, to produce reliable set-valued predictions that are likely to cover the true outcome. In this regard, the notion of \emph{dominance} plays an important role: An outcome $y$ dominates another outcome $y'$ if $y$ is more probable than $y'$ for each distribution in the credal set, that is,
\begin{align} \label{eq:dominance}
\gamma(y,y', \vec{x}) \defeq
\inf_{h \in C} \frac{p( y \given \vec{x}_q, h)}{p(y' \given \vec{x}_q, h)}  > 1 \, ,
\end{align}
where $C$ is a credal set of hypotheses, and $p( y \given \vec{x}, h)$ the probability of the outcome $y$ (for the query $\vec{x}_q$) under hypothesis $h$. 
The set-valued prediction for a query $\vec{x}_q$ is then given by the set of all non-dominated outcomes $y \in \set{Y}$, i.e., those outcomes that are not dominated by any other outcome. Practically, the computations are based on the interval-valued probabilities 
$[\underline{p}(y \given \vec{x}_q), \overline{p}(y \given \vec{x}_q)]$
assigned to each class $y \in \set{Y}$, where 
\begin{align} \label{eq:inter_pro}
\underline{p}(y \given \vec{x}_q) \defeq \inf_{h \in C} p(y \given \vec{x}_q, h)\, , \quad \overline{p}(y \given \vec{x}_q) \defeq \sup_{h \in C} p(y \given \vec{x}_q, h) \, .
\end{align}
Note that the upper probabilities $\overline{p}(y \given \vec{x}_q)$ are very much in the spirit of the plausibility degrees (\ref{eq:plaus}).

Interestingly, uncertainty degrees can be derived on the basis of dominance degrees (\ref{eq:dominance}). For example, the following uncertainty score has been proposed by \citet{antonucci2012} in the context of uncertainty sampling for active learning:
\begin{align} \label{eq:scoreNCC}
s(\vec{x}_q) \defeq - \max \big(\gamma(+1, -1, \vec{x}_q ), \gamma(-1, +1, \vec{x}_q ) \big) \, .
\end{align}
This degree is low if one of the outcomes strongly dominates the other one, but high if this is not the case. It is closely related to the degree of epistemic uncertainty in (\ref{eq:ep}) to be discussed in Section \ref{sec:uqnl}, but also seems to capture some aleatoric uncertainty. In fact, $s(\vec{x}_q)$ increases with both, the width of the interval $[\underline{p}(y \given \vec{x}_q) , \overline{p}(y \given \vec{x}_q)]$ as well as the closeness of its middle point to $\nicefrac{1}{2}$.

\citet{lass_rc20} discusses the modeling of ``credal imprecision'' by means of \emph{sets} of measures in a critical way and contrasts this approach with hierarchical probabilistic (Bayesian) models, i.e., \emph{distributions} of measures (see Fig.\ \ref{fig:setting2}, right side). Among several arguments put forward against sets-of-measures models, one aspect is specifically interesting from a machine learning perspective, namely the problem that modeling a lack of knowledge in a set-based manner may hamper the possibility of inductive inference, up to a point where learning from empirical data is not possible any more. This is closely connected to the insight that learning from data necessarily requires an inductive bias, and learning without assumptions is not possible \citep{wolp_tl96}. To illustrate the point, consider the case of complete ignorance as an example, which could be modeled by the credal set $\mathbb{P}_{all}$ that contains \emph{all} probability measures (on a hypothesis space) as candidate priors. Updating this set by pointwise conditioning on observed data will essentially reproduce the same set as a credal set of posteriors. Roughly speaking, this is because $\mathbb{P}_{all}$ will contain very extreme priors, which will remain extreme even after updating with a sample of finite size. In other words, no finite sample is enough to annihilate a sufficiently extreme prior belief. Replacing $\mathbb{P}_{all}$ by a proper subset of measures may avoid this problem, but any such choice (implying a sharp transition between possible and excluded priors) will appear somewhat arbitrary. According to \citet{lass_rc20}, an appealing alternative is using a probability distribution instead of a set, like in hierarchical Bayesian modeling.

\subsection{Reliable classification}
\label{sec:uqnl}

To the best of our knowledge, \cite{mpub272} were the first to explicitly motivate the distinction between aleatoric and epistemic uncertainty in a machine learning context. Their approach leverages the concept of \emph{normalized likelihood}\mmp{normalized likelihood} (cf.\ Section~\ref{sec:dos}). Moreover, it combines set-based and distributional (probabilistic) inference, and thus can be positioned in-between version space learning and Bayesian inference as discussed in Section \ref{sec:sbvd}. Since the approach contains concepts that might be less known to a machine learning audience, our description is a bit more detailed than for the other methods discussed in this section.

Consider the simplest case of binary classification with classes $\set{Y} \defeq \{-1, +1\}$, which suffices to explain the basic idea (and which has been generalized to the multi-class case by \cite{mpub385}). \cite{mpub272} focus on predictive uncertainty and derive degrees of uncertainty in a prediction in two steps, which we are going to discuss in turn:
\begin{itemize}
\item First, given a query instance $\vec{x}_q$, a degree of ``plausibility'' is derived for each candidate outcome $y \in \set{Y}$. These are degrees in the unit interval, but no probabilities. As they are not constrained by a total mass of 1, they are more apt at capturing a lack of knowledge, and hence epistemic uncertainty (cf.\ Section \ref{sec:rlk}).
\item Second, degrees of aleatoric and epistemic uncertainty are derived from the degrees of plausibility.
\end{itemize}

\subsubsection{Modeling the plausibility of predictions}


Recall that, in version space learning, the plausibility of both hypotheses and outcomes are expressed in a purely bivalent way: according to (\ref{eq:ee1}), a hypotheses is either considered possible/plausible or not ($\llbracket h \in \mathcal{V} \rrbracket$), and an outcome $y$ is either supported or not ($\llbracket h(\vec{x}) = y \rrbracket$). \citet{mpub272} generalize both parts of the prediction from bivalent to \emph{graded} plausibility and support: A hypothesis $h \in \cH$ has a degree of plausibility $\pi_{\cH}(h) \in [0,1]$, and the support of outcomes $y$ is expressed in terms of probabilities $h(\vec{x}) = \prob(y \given \vec{x}) \in [0,1]$.  

More specifically, 
referring to the notion of \emph{normalized likelihood}\mmp{normalized likelihood}, 
a plausibility distribution on $\cH$ (i.e., a plausibility for each hypothesis $h \in \cH$) is defined as follows:
\begin{equation}\label{eq:noli}
\pi_{\cH}(h) \defeq \frac{L(h)}{\sup_{h' \in \cH} L(h')}  = \frac{L(h)}{L(h^{ml})} \enspace ,
\end{equation}
where $L(h)$ is the likelihood of the hypothesis $h$ on the data $\set{D}$ (i.e., the probability of $\set{D}$ under this hypothesis), and $h^{ml} \in \cH$ is the maximum likelihood (ML) estimation. Thus, the plausibility of a hypothesis is in proportion to its likelihood\footnote{In principle, the same idea can of course also be applied in Bayesian inference, namely by defining plausibility in terms of a normalized posterior distribution.}, with the ML estimate having the highest plausibility of 1.

The second step, both in version space and Bayesian learning, consists of translating uncertainty on $\cH$ into uncertainty about the prediction for a query $\vec{x}_q$. To this end, all predictions $h(\vec{x}_{q})$ need to be aggregated, taking into account the plausibility of the hypotheses $h \in \cH$. Due to the problems of the averaging approach (\ref{eq:pd}) in Bayesian inference, a generalization of the ``existential'' aggregation (\ref{eq:ee1}) used in version space learning is adopted:  
\begin{equation}\label{eq:plaus}
\pi(+1 \given \vec{x}_{q}) \defeq \sup_{h \in \cH} \min\big( 
\pi_{\cH}(h) , \pi(+1 \given h, \vec{x}_{q}) \big) ,
\end{equation}
where $\pi(+1 \given h, \vec{x}_{q})$ is the \emph{degree of support} of the positive class provided by $h$\footnote{Indeed, $\pi(\cdot \given h, \vec{x}_{q})$ should not be interpreted as a measure of uncertainty.}. This measure of support, which generalizes the all-or-nothing support $\llbracket h(\vec{x}) = y \rrbracket$ in (\ref{eq:ee1}), is defined as follows: 
\begin{equation}\label{eq:supp}
\pi(+1 \given h, \vec{x}_{q}) \defeq \max \big(2 h(\vec{x}_{q})-1, 0 \big)
\end{equation}
Thus, the support is 0 as long as the probability predicted by $h$ is $\leq 1/2$, and linearly increases afterward, reaching 1 for $h(\vec{x}_{q})=1$. Recalling that $h$ is a probabilistic classifier, this clearly makes sense, since values $h(\vec{x}_{q}) \leq 1/2$ are actually more in favor of the negative class, and therefore no evidence for the positive class. Also, as will be seen further below, this definition assures a maximal degree of aleatoric uncertainty in the case of full certainty about the uniform distribution $h_{1/2}$, wich is a desirable property. 
Since the supremum operator in (\ref{eq:plaus}) can be seen as a generalized existential quantifier, the expression (\ref{eq:plaus}) can be read as follows: The class $+1$ is plausible insofar there exists a hypothesis $h$ that is plausible and that strongly supports $+1$. Analogously, the plausibility for $-1$ is defined as follows:
\begin{equation}\label{eq:plausminus}
\pi(-1 \given \vec{x}_{q}) \defeq \sup_{h \in \cH} \min\big( 
\pi_{\cH}(h) , \pi(-1 \given h, \vec{x}_{q}) \big) ,
\end{equation}
with $\pi(-1 \given h, \vec{x}_{q}) = \max(1 - 2 h(\vec{x}_{q}), 0)$.
An illustration of this kind of ``max-min inference'' and a discussion of how it relates to Bayesian inference (based on ``sum-product aggregation'') can be found in Appendix \ref{sec:maxmin} in the supplementary material.

\subsubsection{From plausibility to aleatoric and epistemic uncertainty}


Given the plausibilities $\pi(+1) = \pi(+1 \given \vec{x}_{q})$ and $\pi(-1) = \pi(-1\given \vec{x}_{q})$ of the positive and negative class, respectively, and having to make a prediction, one would naturally decide in favor of the more plausible class. Perhaps more interestingly, meaningful definitions of epistemic uncertainty $u_e$ and aleatoric uncertainty $u_a$ can be defined on the basis of the two degrees of plausibility: 
\begin{equation}\label{eq:ep}
\begin{split}
u_e & \defeq \min \big( \pi(+1 ) , \pi(-1) \big) \\
u_a &  \defeq  1 - \max \big( \pi(+1) , \pi(-1) \big)
\end{split}
\end{equation}
Thus, $u_e$ is the degree to which both $+1$ and $-1$ are plausible\footnote{The minimum plays the role of a generalized logical conjunction \citep{klem_tn}.}, and $u_a$ the degree to which neither $+1$ nor $-1$ are plausible.
Since these two degrees of uncertainty satisfy $u_a + u_e  \leq 1$,
the total uncertainty (aleatoric $+$ epistemic) is upper-bounded by 1. Strictly speaking, since the degrees $\pi(y)$ should be interpreted as upper bounds on plausibility, which decrease in the course of time (with increasing sample size), the uncertainty degrees (\ref{eq:ep}) should be interpreted as bounds as well. For example, in the very beginning, when no or very little data has been observed, both outcomes are fully plausible ($\pi(+1) = \pi(-1) = 1$), hence $u_e = 1$ and $u_a = 0$. This does not mean, of course, that there is no aleatoric uncertainty involved. Instead, it is simply not yet known or, say, confirmed. Thus, $u_a$ should be seen as a lower bound on the ``true'' aleatoric uncertainty. For instance, when $y$ is the outcome of a fair coin toss, the aleatoric uncertainty will increase over time and reach 1 in the limit, while $u_e$ will decrease and vanish at some point: Eventually, the learner will be fully aware of facing full aleatoric uncertainty.

More generally, the following special cases might be of interest:
\begin{itemize}
\item 
\emph{Full epistemic uncertainty}: $u_e = 1$ requires the existence of at least two fully plausible hypotheses (i.e., both with the highest likelihood), the one fully supporting the positive and the other fully supporting the negative class. This situation is likely to occur (at least approximately) in the case of a small sample size, for which the likelihood is not very peaked. 

\item
\emph{No epistemic uncertainty}: $u_e = 0$ requires either $\pi(+1) = 0$ or $\pi(-1)=0$, which in turn means that $h(\vec{x}_{q}) < 1/2$ for all hypotheses with non-zero plausibility, or $h(\vec{x}_{q}) > 1/2$ for all these hypotheses. In other words, there is no disagreement about which of the two classes should be favored. Specifically, suppose that all plausible hypotheses agree on the same conditional probability distribution $\prob(+1 \given \vec{x}) = \alpha$ and $\prob(-1 \given \vec{x}) = 1-\alpha$, and let $\beta = \max(\alpha , 1- \alpha)$. In this case, $u_e = 0$, and the degree of aleatoric uncertainty $u_a = 2(1- \beta)$ depends on how close $\beta$ is to 1. 

\item 
\emph{Full aleatoric uncertainty}: This is a special case of the previous one, in which $\beta = 1/2$. Indeed, $u_a = 1$ means that all plausible hypotheses assign a probability of $1/2$ to both classes. In other words, there is an agreement that the query instance is a boundary case.

\item
\emph{No uncertainty}: Again, this is a special case of the second one, with $\beta = 1$. A clear preference (close to 1) in favor of one of the two classes means that all plausible hypotheses, i.e., all hypotheses with a high likelihood, provide full support to that class.
\end{itemize}

Although algorithmic aspects are not in the focus of this paper, it is worth to mention that the computation of (\ref{eq:plaus}), and likewise of (\ref{eq:plausminus}), may become rather complex. In fact, the computation of the supremum comes down to solving an optimization problem, the complexity of which strongly depends on the hypothesis space $\cH$. 


\subsection{Conformal prediction}
\label{sec:cp}

Conformal prediction \citep{vovk_al,shaf_at08,bala_cp} is a framework for reliable prediction that is rooted in classical frequentist statistics, more specifically in hypothesis testing. Given a sequence of training observations and a new query $\vec{x}_{N+1}$ (which we denoted by $\vec{x}_{q}$ before) with unknown outcome $y_{N+1}$,
\begin{equation}\label{eq:cpseq}
(\vec{x}_1, y_1), \,  (\vec{x}_2, y_2), \ldots ,  (\vec{x}_N, y_N), \, (\vec{x}_{N+1}, \bullet)
\enspace ,
\end{equation}
the basic idea is to hypothetically replace $\bullet$ by each candidate, i.e., to test the hypothesis $y_{N+1} = y$ for all $y \in \mathcal{Y}$. Only those outcomes $y$ for which this hypothesis can be rejected at a predefined level of confidence are excluded, while those for which the hypothesis cannot be rejected are collected to form the prediction set or \emph{prediction region} $Y^\epsilon \subseteq \mathcal{Y}$. The construction of a set-valued prediction $Y^\epsilon = Y^\epsilon(\vec{x}_{n+1})$ that is guaranteed to cover the true outcome $y_{N+1}$ with a given probability $1- \epsilon$ (for example 95\,\%), instead of producing a point prediction $\haty_{N+1} \in \mathcal{Y}$, is the basic idea of conformal prediction. Here, $\epsilon \in (0,1)$ is a pre-specified level of significance. In the case of classification, $Y^\epsilon$ is a subset of the set of classes $\mathcal{Y} = \{ y_1, \ldots , y_K \}$, whereas in regression, a prediction region is commonly represented in terms of an interval\footnote{Obviously, since $\mathcal{Y} = \mathbb{R}$ is infinite in regression, a hypothesis test cannot be conducted explicitly for each candidate outcome $y$.}. 

Hypothesis testing is done in a nonparametric way: Consider any ``nonconformity'' function $f: \, \mathcal{X} \times \mathcal{Y} \longrightarrow \mathbb{R}$ that assigns scores $\alpha = f(\vec{x}, y)$ to input/output tuples; the latter can be interpreted as a measure of ``strangeness'' of the pattern $(\vec{x}, y)$, i.e., the higher the score, the less the data point $(\vec{x}, y)$ conforms to what one would expect to observe. Applying this function to the sequence (\ref{eq:cpseq}), with a specific (though hypothetical) choice of $y = y_{N+1}$, yields a sequence of scores
$$
\alpha_1, \, \alpha_2, \ldots , \alpha_N , \, \alpha_{N+1} \enspace .
$$
Denote by $\sigma$ the permutation of $\{1, \ldots , N+1\}$ that sorts the scores in increasing order, i.e., such that $\alpha_{\sigma(1)} \leq \ldots \leq \alpha_{\sigma(N+1)}$. Under the assumption that the hypothetical choice of $y_{N+1}$ is in agreement with the true data-generating process, and that this process has the property of exchangeability (which is weaker than the assumption of independence and essentially means that the order of observations is irrelevant), every permutation $\sigma$ has the same probability of occurrence. Consequently, the probability that $\alpha_{N+1}$ is among the $\epsilon$\,\% highest nonconformity scores should be low. This notion can be captured by the $p$-values associated with the candidate $y$, defined as 
$$
p(y) \defeq \frac{\# \{ i \given \alpha_i \geq \alpha_{N+1} \}}{N+1}
$$
According to what we said, the probability that $p(y) < \epsilon$ (i.e., $\alpha_{N+1}$ is among the $\epsilon$\,\% highest $\alpha$-values) is upper-bounded by $\epsilon$. 
Thus, the hypothesis $y_{N+1} = y$ can be rejected for those candidates $y$ for which $p(y) < \epsilon$. 

Conformal prediction as outlined above realizes transductive inference, although inductive variants also exist \citep{papa_ic08}.  The error bounds are 
valid and well calibrated by construction, regardless of the nonconformity function $f$. However, the choice of this function has an important influence on the \emph{efficiency} of conformal prediction, that is, the size of prediction regions: The more suitable the nonconformity function is chosen, the smaller these sets will be. 

Although conformal prediction is mainly concerned with constructing prediction regions, the scores produced in the course of this construction can also be used for quantifying uncertainty. In this regard, the notions of \emph{confidence} and \emph{credibility} \mmp{confidence and credibility} have been introduced \citep{gam_pa02}: Let $p_1, \ldots , p_K$ denote the $p$-values that correspond, respectively, to the candidate outcomes $y_1, \ldots , y_K$ in a classification problem. If a definite choice (point prediction) $\haty$ has to be made, it is natural to pick the $y_i$ with the highest $p$-value. The value $p_i$ itself is then a natural measure of credibility, since the larger (closer to 1) this value, the more likely the prediction is correct. Note that the value also corresponds to the largest significance level $\epsilon$ for which the prediction region $Y^\epsilon$ would be empty (since all candidates would be rejected). In other words, it is a degree to which $y_i$ is indeed a plausible candidate that cannot be excluded. Another question one may ask is to what extent $y_i$ is the unique candidate of that kind, i.e., to what extent other candidates can indeed be excluded. This can be quantified in terms of the greatest $1 - \epsilon$ for which $Y^\epsilon$ is the singleton set $\{ y_i \}$, that is, $1$ minus the second-highest $p$-value. Besides, other methods for quantifying the uncertainty of a point prediction in the context of conformal prediction have been proposed \citep{linu_rc16}. 


With its main concern of constructing valid prediction regions, conformal prediction differs from most other machine learning methods, which produce point predictions $y \in \mathcal{Y}$, whether equipped with a degree of uncertainty or not. In a sense, conformal prediction can even be seen as being orthogonal: It predefines the degree of uncertainty (level of confidence) and adjusts its prediction correspondingly, rather than the other way around. 

\subsection{Set-valued prediction based on utility maximization}
\label{sec:sbus}

In line with classical statistics, but unlike decision theory and machine learning, the setting of conformal prediction does not involve any notion of loss function. In this regard, it differs from methods for set-valued prediction based on utility maximization (or loss minimization), which are primarily used for multi-class classification problems. Similar to conformal prediction, such methods also return a set of classes when the classifier is too uncertain with respect to the class label to predict, but the interpretation of this set is different. Instead of returning a set that contains the true class label with high probability, sets that maximize a set-based utility score are sought. Besides, being based on the conditional distribution $\prob(y\,|\,\vec{x}_q)$ of the outcome $y$ given a query $\vec{x}_q$, most of these methods capture \emph{conditional} uncertainty. 

Let $u(y,\sety)$ be a set-based utility score, where $y$ denotes the ground truth outcome and $\sety$ the predicted set. Then, adopting a decision-theoretic perspective, the Bayes-optimal solution $\sety^{*}_u$ is found by maximizing the following objective:
\begin{equation}
\label{eq:bayesoptimal}
\sety^{*}_u(\vec{x}_q) = \argmax_{\sety \in 2^{\mathcal{Y}}\setminus \{\emptyset\}} \evalue_{p(y\,|\,\vec{x}_q)} \big( u(y,\sety) \big) = \argmax_{\sety \in 2^{\mathcal{Y}}\setminus \{\emptyset\}} \sum_{y \in \mathcal{Y}} u(y,\sety) \, p(y\,|\,\vec{x}_q) \,.
\end{equation}
Solving (\ref{eq:bayesoptimal}) as a brute-force search requires checking all subsets of $\mathcal{Y}$, resulting in an exponential time complexity. However, for many utility scores, the Bayes-optimal prediction can be found more efficiently. Various methods in this direction have been proposed under different names and qualifications of predictions, such as ``indeterminate'' \citep{zaffalon2002}, ``credal'' \citep{cora_lr08}, ``non-deterministic'' \citep{delc_ln09}, and ``cautious'' \citep{Yang2017b}. Although the methods typically differ in the exact shape of the utility function $u: \mathcal{Y} \times 2^{\mathcal{Y}}\setminus \{\emptyset\} \longrightarrow [0,1]$, most functions are specific members of the following family:
\begin{equation}
\label{eq:ufamily}
u(y,\sety) = \left \{ 
\begin{array}{cl}
0 &\quad \mbox{if $y \notin \sety$} \ \\
g(|\sety|)&\quad \mbox{if $y \in \sety$} 
\end{array} \, ,
\right.
\end{equation}
where $|\sety|$ denotes the cardinality of the predicted set $\sety$. 
This family is characterized  by a sequence $(g(1), \ldots,g(K)) \in [0,1]^K$ with $K$ the number of classes. Ideally, $g$ should obey the following properties: 
\begin{enumerate}
\item[(i)] $g(1) = 1$, i.e., the utility $u(y,\sety)$ should be maximal when the classifier returns the true class label as a singleton set.
\item[(ii)] $g(s)$ should be non-increasing, i.e., the utility $u(y,\sety)$ should be higher if the true class is contained in a smaller set of predicted classes.
\item[(iii)] $g(s) \geq 1/s$, i.e., the utility $u(y,\sety)$ of predicting a set containing the true and $s-1$ additional classes should not be lower than the expected utility of randomly guessing one of these $s$ classes. This requirement formalizes the idea of risk-aversion: in the face of uncertainty, abstaining should be preferred to random guessing \citep{Zaffalon2012EvaluatingCC}. 
\end{enumerate}

Many existing set-based utility scores are recovered as special cases of (\ref{eq:ufamily}), including the three classical measures from information retrieval discussed by \citet{delc_ln09}: precision with $g_P(s) = 1/s$, recall with $g_R(s)=1$, and the F$_{1}$-measure with $g_{F1}(s) = 2/(1+s)$.  Other utility functions with specific choices for $g$ are studied in the literature on credal classification \citep{Corani2008NCC,Corani2009LNCC,Zaffalon2012EvaluatingCC,Yang2017b,mpub385}, such as
$$g_{\delta,\gamma}(s) \defeq \frac{\delta}{s} - \frac{\gamma}{s^2} \,, \quad g_{\exp}(s) \defeq 1- \exp{\left(-\frac{\delta}{s}\right)},   \quad g_{log}(s) \defeq \log \left(1 + \frac{1}{s} \right) \,.$$
Especially $g_{\delta,\gamma}(s)$ is commonly used in this community, where $\delta$ and $\gamma$ can only take certain values to guarantee that the utility is in the interval $[0,1]$. Precision (here called discounted accuracy) corresponds to the case $(\delta,\gamma)=(1,0)$. However, typical choices for $(\delta,\gamma$) are $(1.6,0.6)$ and $(2.2,1.2)$ \citep{mpub385}, implementing the idea of risk aversion. The measure $g_{\exp}(s)$ is an exponentiated version of precision, where the parameter $\delta$ also defines the degree of risk aversion. 

Another example appears in the literature on binary or multi-class classification with reject option \citep{herb_cw06,linu_cw18,Ramaswamy2015CAMCRO}. Here, the prediction can only be a singleton or the full set $\mathcal{Y}$ containing $K$ classes. The first case typically gets a reward of one (if the predicted class is correct), while the second case should receive a lower reward, e.g.\ $1-\alpha$.  The latter corresponds to abstaining, i.e., not predicting any class label, and the (user-defined) parameter $\alpha$ specifies a penalty for doing so, with the requirement $0< \alpha < 1-1/K$ to be risk-averse. 


Set-valued predictions are also considered in hierarchical multi-class classification, mostly in the form of internal nodes of the hierarchy  \citep{Freitas_atutorial,Rangwala2017,Yang2017}. Compared to the ``flat'' multi-class case, the prediction space is thus restricted, because only sets of classes that correspond to nodes of the hierarchy can be returned as a prediction. Some of the above utility scores also appear here. For example, \citet{Yang2017} evaluate various members of the $u_{\delta,\gamma}$ family in a framework where hierarchies are considered for computational reasons, while \citet{Oh2017TopKHC} optimizes recall by fixing $|\sety|$ as a user-defined parameter. Popular in hierarchical classification is the tree-distance loss, which could also be interpreted as a way of evaluating set-valued predictions \citep{Bi2015}. This loss is not a member of the family (\ref{eq:ufamily}), however. Besides, it appears to be a less interesting loss function from the perspective of abstention in cases of uncertainty, since by minimizing the tree distance loss, the classifier will almost never predict leaf nodes of the hierarchy. Instead, it will often predict nodes close to the root of the hierarchy, which correspond to sets with many elements\,---\,a behavior that is unfavored if one wants to abstain only in cases of sufficient uncertainty. 

Quite obviously, methods that maximize set-based utility scores are closely connected to the quantification of uncertainty, since the decision about a suitable set of predictions is necessarily derived from information of that kind. The overwhelming majority of the above-mentioned methods depart from conditional class probabilities $p(y \given \vec{x}_q)$ that are estimated in a classical frequentist way, so that uncertainties in decisions are of aleatoric nature. 
Exceptions include \citep{Yang2017} and \citep{mpub385}, who further explore ideas from imprecise probability theory and reliable classification to generate label sets that capture both aleatoric and epistemic uncertainty.

\section{Discussion and conclusion}

The importance to distinguish between different types of uncertainty has recently been recognized in machine learning. The goal of this paper was to sketch existing approaches in this direction, with an emphasis on the quantification of aleatoric and epistemic uncertainty about predictions in supervised learning. Looking at the problem from the perspective of the standard setting of supervised learning, it is natural to associate epistemic uncertainty with the lack of knowledge about the true (or Bayes-optimal) hypothesis $h^*$ within the hypothesis space $\mathcal{H}$, i.e., the uncertainty that is in principle reducible by the learner and could get rid of by additional data. Aleatoric uncertainty, on the other hand, is the irreducible part of the uncertainty in a prediction, which is due to the inherently stochastic nature of the dependency between instances $\vec{x}$ and outcomes $y$.


In a Bayesian setting, epistemic uncertainty is hence reflected by the (posterior) probability $\prob(h \given \cD)$ on $\mathcal{H}$: The less peaked this distribution, the less informed the learner is, and the higher its (epistemic) uncertainty. As we argued, however, important information about this uncertainty might get lost through Bayesian model averaging. In this regard, we also argued that a (graded) set-based representation of epistemic uncertainty could be a viable alternative to a probabilistic representation, especially due to the difficulty of representing ignorance with probability distributions. This idea is perhaps most explicit in version space learning, where epistemic uncertainty is in direct correspondence with the size of the version space. The approach by \citet{mpub272} combines concepts from (generalized) version space learning and Bayesian inference.


There are also other methods for modeling uncertainty, such as conformal prediction, for which the role of aleatoric and epistemic uncertainty is not immediately clear. Currently, the field develops very dynamically and is far from being settled. New proposals for modeling and quantifying uncertainty appear on a regular basis, some of them rather ad hoc and others better justified. Eventually, it would be desirable to ``derive'' a measure of total uncertainty as well as its decomposition into aleatoric and epistemic parts from basic principles in a mathematically  rigorous way, i.e., to develop a kind of axiomatic basis for such a decomposition, comparable to axiomatic justifications of the entropy measure \citep{csis_ac08}. 

In addition to theoretical problems of this kind, there are also many open practical questions. This includes, for example, the question of how to perform an empirical evaluation \mmp{empirical evaluation} of methods for quantifying uncertainty, whether aleatoric, epistemic, or total. In fact, unlike for the prediction of a target variable, the data does normally not contain information about any sort of ``ground truth'' uncertainty. What is often done, therefore, is to evaluate predicted uncertainties \emph{indirectly}, that is, by assessing their usefulness for improved prediction and decision making. As an example, recall the idea of utility maximization discussed in Section \ref{sec:sbus}, where the decision is a set-valued prediction. Another example is \emph{accuracy-rejection curves}\mmp{accuracy-rejection curves}, which depict the accuracy of a predictor as a function of the percentage of rejections \citep{mpub170}: A classifier, which is allowed to abstain on a certain percentage $p$ of predictions, will predict on those $(1-p)$\,\% on which it feels most certain. Being able to quantify its own uncertainty well, it should improve its accuracy with increasing $p$, hence the accuracy-rejection curve should be monotone increasing (unlike a flat curve obtained for random abstention).  

Most approaches so far neglect model uncertainty, in the sense of proceeding from the (perhaps implicit) assumption of a correctly specified hypothesis space $\mathcal{H}$, that is, the assumption that $\mathcal{H}$ contains a (probabilistic) predictor which is coherent with the data.  In \citep{mpub272}, for example, this assumption is reflected by the definition of plausibility in terms of normalized likelihood, which always ensures the existence of at least one fully plausible hypothesis---indeed, (\ref{eq:noli}) is a measure of \emph{relative}, not \emph{absolute} plausibility. On the one side, it is true that model induction, like statistical inference in general, is not possible without underlying assumptions, and that conclusions drawn from data are always conditional to these assumptions. On the other side, misspecification of the model class is a common problem in practice, and should therefore not be ignored. In principle, conflict and inconsistency can be seen as another source of uncertainty, in addition to randomness and a lack of knowledge. Therefore, it would be useful to reflect this source of uncertainty as well (for example in the form of non-normalized plausibility functions $\pi_{\cH}$, in which the gap $1- \sup_h \pi_{\cH}(h)$ serves as a measure of inconsistency between $\cH$ and the data $\cD$, and hence as a measure of model uncertainty). 

Related to this is the ``closed world'' assumption \citep{deng_ge14}, which is often violated in contemporary machine learning applications. Modeling uncertainty and allowing the learner to express ignorance is obviously important in scenarios where new classes may emerge in the course of time, which implies a change of the underlying data-generating process, or the learner might be confronted with out-of-distribution queries. Some first proposals for dealing with this case can already be found in the literature \citep{devr_lc18,lee_as18,mali_pu18,sens_ed18}.

Finally, there are many ways in which other machine learning methodology may benefit from a proper quantification of uncertainty, and in which corresponding measures could be used for ``uncertainty-informed'' decisions. For example, \cite{mpub392} take advantage of the distinction between epistemic and aleatoric uncertainty in active learning, arguing that the former is more relevant as a selection criterion for uncertainty sampling than the latter. Likewise, as already said, a suitable quantification of uncertainty can be very useful in set-valued prediction, where the learner is allowed to predict a subset of outcomes (and hence to partially abstain) in cases of uncertainty.

\subsection*{Acknowledgment}
The authors like to thank Sebastian Destercke, Karlson Pfannschmidt, and Ammar Shaker for helpful remarks and comments on the content of this paper. We are also grateful for the additional suggestions made by the reviewers.  The second author received funding from the Flemish Government under the ``Onderzoeksprogramma Artificiel\"e
Intelligentie (AI) Vlaanderen'' Programme.

\begin{appendix}

\section{Background on uncertainty modeling}
\label{app:unc}

The notion of uncertainty has been studied in various branches of science and scientific disciplines. For a long time, it plays a major role in fields like economics, psychology, and the social sciences, typically in the appearance of applied statistics. Likewise, its importance for artificial intelligence has been recognized very early on\footnote{The ``Annual Conference on Uncertainty in Artificial Intelligence'' (UAI) was launched in the mid 1980s.}, at the latest with the emergence of expert systems, which came along with the need for handling inconsistency, incompleteness, imprecision, and vagueness in knowledge representation \citep{krus_ua}. More recently, the phenomenon of uncertainty has also attracted a lot of attention in engineering, where it is studied under the notion of ``uncertainty quantification'' \citep{owha_ou12}; interestingly, a distinction between aleatoric and epistemic uncertainty, very much in line with our machine learning perspective, is also made there.

\begin{figure}
\begin{center}
\includegraphics[scale=0.45]{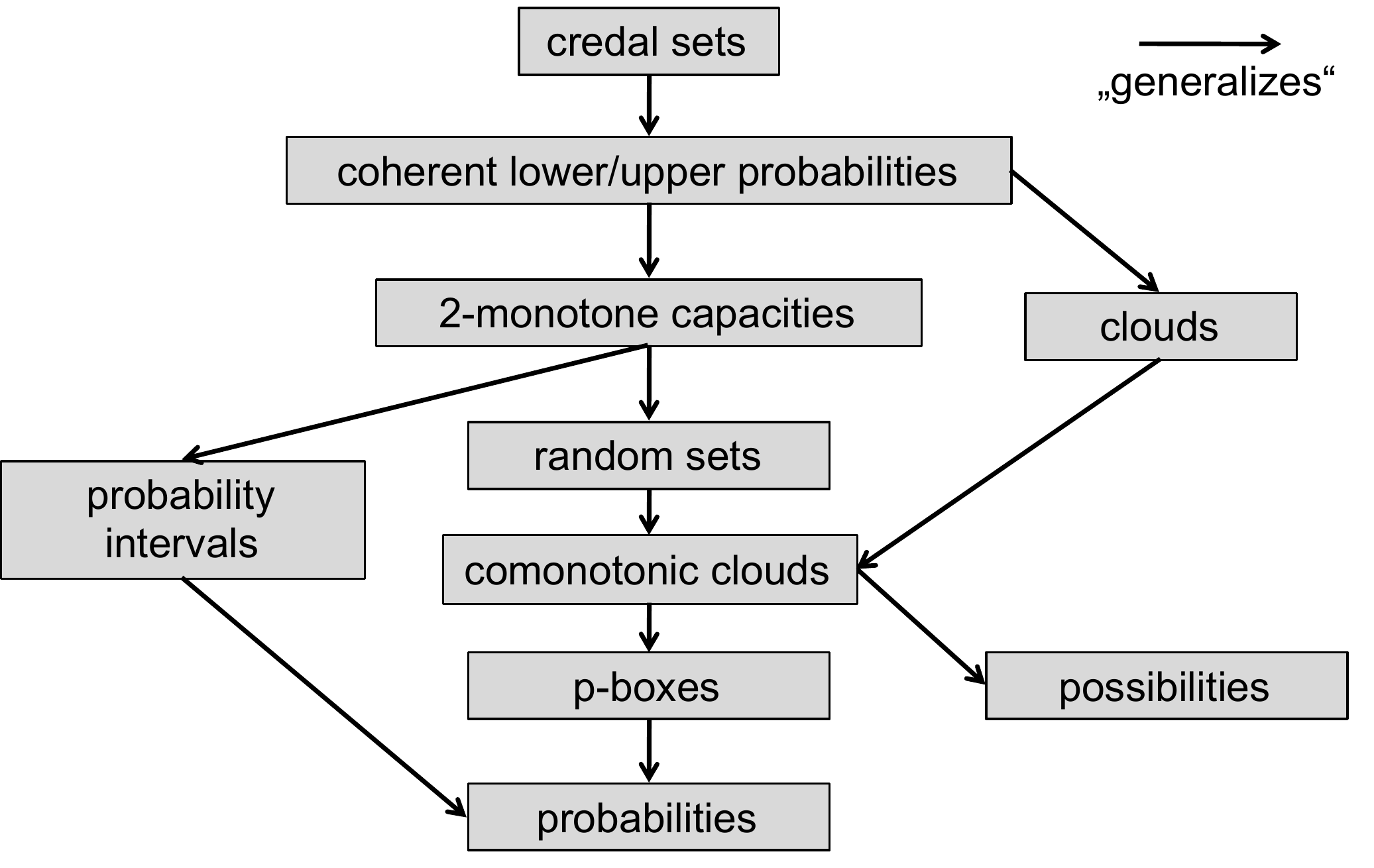} 
\caption{Various uncertainty calculi and common frameworks for uncertainty representation \citep{dest_up08}. Most of these formalisms are generalizations of standard probability theory (an arrow denotes an ``is more general than'' relationship).}
\label{fig:calculi}
\end{center}
\end{figure}

The contemporary literature on uncertainty is rather broad (cf.\ Fig.\ \ref{fig:calculi}). In the following, we give a brief overview, specifically focusing on the distinction between set-based and distributional (probabilistic) representations. Against the background of our discussion about aleatoric and epistemic uncertainty, this distinction is arguably important. Roughly speaking, while aleatoric uncertainty is appropriately modeled in terms of probability distributions, one may argue that a set-based approach is more suitable for modeling ignorance and a lack of knowledge, and hence more apt at capturing epistemic uncertainty.

\subsection{Sets versus distributions}

A generic way for describing situations of uncertainty is to proceed from an underlying reference set $\Omega$, sometimes called the \emph{frame of discernment} \citep{shaf_am}. This set consists of all hypotheses, or pieces of precise information, that ought to be distinguished in the current context.  Thus, the elements $\omega \in \Omega$ are exhaustive and mutually exclusive, and one of them, $\omega^*$, corresponds to the truth. For example, $\Omega = \{ H, T \}$ in the case of coin tossing, $\Omega = \{ \text{win}, \text{loss}, \text{tie} \}$ in predicting the outcome of a football match, or $\Omega = \mathbb{R} \times \mathbb{R}_+$ in the estimation of the parameters (expected value and standard deviation) of a normal distribution from data. For ease of exposition and to avoid measure-theoretic complications, we will subsequently assume that $\Omega$ is a discrete (finite or countable) set. 

As an aside, we note that the assumption of exhaustiveness of $\Omega$ could be relaxed. In a classification problem in machine learning, for example, not all possible classes might be known beforehand, or new classes may emerge in the course of time \citep{hend_ab17,lian_et18,devr_lc18}. In the literature, this is often called the ``open world assumption'', whereas an exhaustive $\Omega$ is considered as a ``closed world'' \citep{deng_ge14}. Although this distinction may look technical at first sight, it has important consequences with regard to the representation and processing of uncertain information, which specifically concern the role of the empty set. While the empty set is logically excluded as a valid piece of information under the closed world assumption, it may suggest that the true state $\omega^*$ is outside $\Omega$ under the open world assumption.

There are two basic ways for expressing uncertain information about $\omega^*$, namely, in terms of \emph{subsets} and in terms of \emph{distributions}. A subset $C \subseteq \Omega$ corresponds to a constraint suggesting that $\omega^* \in C$. Thus, information or knowledge\footnote{We do not distinguish between the notions of information and knowledge in this paper.} expressed in this way distinguishes between values that are (at least provisionally) considered possible and those that are definitely excluded. As suggested by common examples such as specifying incomplete information about a numerical quantity in terms of an interval $C= [l,u]$, a set-based representation is appropriate for capturing uncertainty in the sense of \emph{imprecision}.

Going beyond this rough dichotomy, a distribution assigns a weight $p(\omega)$ to each element $\omega$, which can generally be understood as a degree of belief. At first sight, this appears to be a proper generalization of the set-based approach. Indeed, without any constraints on the weights, each subset $C$ can be characterized in terms of its indicator function $\mathbb{I}_C$ on $\Omega$ (which is a specific distribution assigning a weight of 1 to each $\omega \in C$ and 0 to all $\omega \not\in \Omega$). However, for the specifically important case of probability distributions, this view is actually not valid.


First, probability distributions need to obey a normalization constraint. In particular, a probability distribution requires the weights to be nonnegative and integrate to 1. A corresponding probability measure on $\Omega$ is a set-function $\Prob: \, 2^\Omega \longrightarrow [0,1]$ such that $\Prob(\emptyset) = 0$, $\Prob(\Omega)=1$, and 
\begin{equation}\label{eq:addi}
\Prob(A \cup B) = \Prob(A) + \Prob(B) 
\end{equation}
for all disjoint sets (events) $A, B \subseteq \Omega$. With $\prob(\omega) = \Prob(\{ \omega \})$ for all $\omega \in \Omega$ it follows that $\Prob(A) = \sum_{\omega \in A} \prob(\omega)$, and hence $\sum_{\omega \in \Omega} \prob(\omega)=1$.
Since the set-based approach does not (need to) satify this constraint, it is no longer a special case.

Second, in addition to the question of how information is represented, it is of course important to ask how the information is processed. In this regard, the probabilistic calculus differs fundamentally from constraint-based (set-based) information processing. The characteristic property of probability is its additivity (\ref{eq:addi}), suggesting that the belief in the disjunction (union) of two (disjoint) events $A$ and $B$ is the sum of the belief in either of them. In contrast to this, the set-based approach is more in line with a logical interpretation and calculus. Interpreting a constraint $C$ as a logical proposition $(\omega \in C)$, an event $A \subseteq \Omega$ is possible as soon as $A \cap C \neq \emptyset$ and impossible otherwise. Thus, the information $(\omega \in C)$ can be associated with a set-function $\Pi:\, 2^\Omega \longrightarrow \{ 0, 1 \}$ such that $\Pi(A) = \llbracket A \cap C \neq \emptyset \rrbracket$. Obviously, this set-function satisfies $\Pi(\emptyset) = 0$, $\Pi(\Omega) = 1$, and 
\begin{equation}\label{eq:maxi}
\Pi(A \cup B) = \max \big( \Pi(A) , \Pi(B) \big) 
\end{equation}
Thus, $\Pi$ is ``maxitive'' instead of additive \citep{shil_mm71,dubo_pt06}. Roughly speaking, an event $A$ is evaluated according to its (logical) consistency with a constraint $C$, whereas in probability theory, an event $A$ is evaluated in terms of its probability of occurrence. The latter is reflected by the probability mass assigned to $A$, and requires a comparison of this mass with the mass of other events (since only one outcome $\omega$ is possible, the elementary events compete with each other). Consequently, the calculus of probability, including rules for combination of information, conditioning, etc., is quite different from the corresponding calculus of constraint-based information processing \citep{dubo_pt06}.

\subsection{Representation of ignorance}
\label{sec:aproi}

From the discussion so far, it is clear that a probability distribution is essentially modeling the phenomenon of \emph{chance} rather than \emph{imprecision}. One may wonder, therefore, to what extent it is suitable for representing epistemic uncertainty in the sense of a lack of knowledge. 

In the set-based approach, there is an obvious correspondence between the degree of uncertainty or imprecision and the cardinality of the set $C$: the larger $|C|$, the larger the lack of knowledge\footnote{In information theory, a common uncertainty measure is $\log(|C|)$.}. Consequently, knowledge gets weaker by adding additional elements to $C$. In probability, the total amount of belief is fixed in terms of a unit mass that is distributed among the elements $\omega \in \Omega$, and increasing the weight of one element $\omega$ requires decreasing the weight of another element $\omega'$ by the same amount. Therefore, the knowledge expressed by a probability measure cannot be ``weakened'' in a straightforward way.

Of course, there are also measures of uncertainty for probability distributions, most notably the (Shannon) entropy
$$
H(p) \defeq - \sum_{\omega \in \Omega} p(\omega) \, \log p(\omega) \, .
$$
However, these are primarily capturing the shape of the distribution, namely its ``peakedness'' or non-uniformity \citep{mpub128}, and hence inform about the predictability of the outcome of a random experiment. Seen from this point of view, they are more akin to aleatoric uncertainty, whereas the set-based approach is arguably better suited for capturing epistemic uncertainty. 

For these reasons, it has been argued that probability distributions are less suitable for representing \emph{ignorance} in the sense of a lack of knowledge \citep{dubo_rp96}. 
For example, the case of \emph{complete ignorance} is typically modeled in terms of the uniform distribution $p \equiv 1/|\Omega|$ in probability theory; this is justified by the ``principle of indifference'' invoked by Laplace, or by referring to the principle of maximum entropy\footnote{Obviously, there is a technical problem in defining the uniform distribution in the case where $\Omega$ is not finite.}. Then, however, it is not possible to distinguish between (i) precise (probabilistic) knowledge about a random event, such as tossing a fair coin, and (ii) a complete lack of knowledge, for example due to an incomplete description of the experiment. This was already pointed out by the famous Ronald Fisher, who noted that ``\emph{not knowing the chance of mutually exclusive events and knowing the chance to be equal are two quite different states of knowledge}''.  

Another problem in this regard is caused by the measure-theoretic grounding of probability and its additive nature. For example, the uniform distribution is not invariant under reparametrization (a uniform distribution on a parameter $\omega$ does not translate into a uniform distribution on $1/\omega$, although ignorance about $\omega$ implies ignorance about $1/\omega$). For example, expressing ignorance about the length $x$ of a cube in terms of a uniform distribution on an interval $[l,u]$ does not yield a uniform distribution of $x^3$ on $[l^3 , u^3]$, thereby suggesting some degree of informedness about its volume. Problems of this kind render the use of a uniform prior distribution, often interpreted as representing epistemic uncertainty in Bayesian inference, at least debatable\footnote{This problem is inherited by hierarchical Bayesian modeling. See work on ``non-informative'' priors, however \citep{jeff_46,bern_rp79}.}.

\subsection{Sets of distributions}

Given the complementary nature of sets and distributions, and the observation that both have advantages and disadvantages, one may wonder whether the two could not be combined in a meaningful way. Indeed, the argument that a single (probability) distribution is not enough for representing uncertain knowledge is quite prominent in the literature, and many generalized theories of uncertainty can be considered as a combination of that kind \citep{dubo_pt,wall_sr,shaf_am,smet_tt94}. 

\begin{figure}
\begin{center}
\includegraphics[scale=0.45]{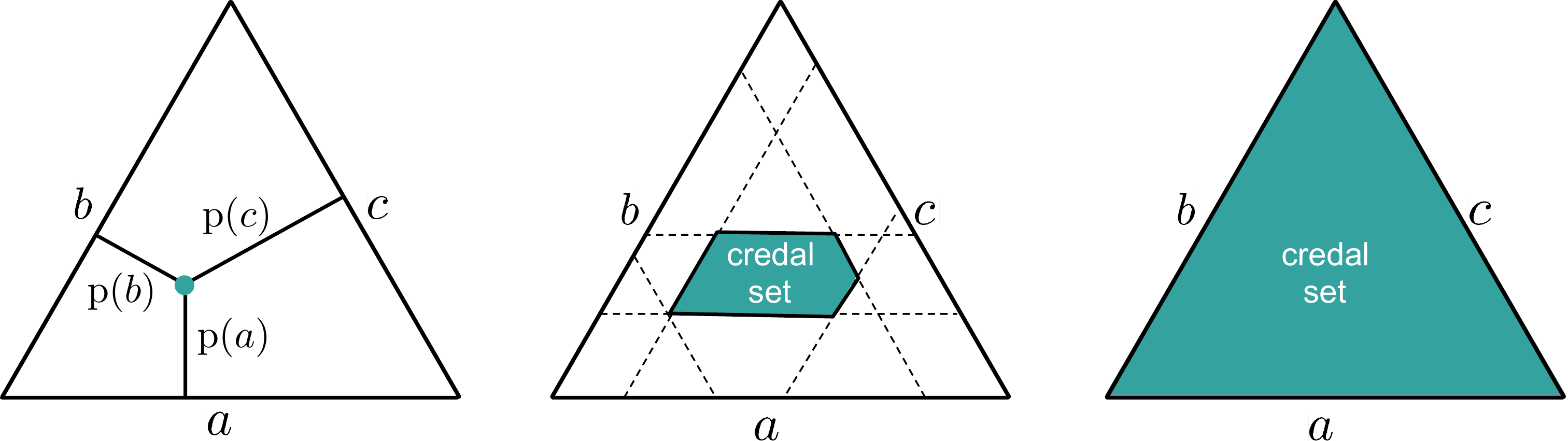}
\caption{Probability distributions on $\Omega = \{ a,b,c \}$ as points in a Barycentric coordinate system: Precise knowledge (left) versus incomplete knowledge (middle) and complete ignorance (right) about the true distribution.}
\label{fig:bary}
\end{center}
\end{figure}

Since we are specifically interested in aleatoric and epistemic uncertainty, and since these two types of uncertainty are reasonably captured in terms of sets and probability distributions, respectively, a natural idea is to consider \emph{sets of probability distributions}. In the literature on imprecise probability, these are also called \emph{credal sets} \citep{cozm_cn00,zaff_tn02}. An illustration is given in Fig.\ \ref{fig:bary}, where probability distributions on $\Omega = \{ a,b,c \}$ are represented as points in a Barycentric coordinate systems. A credal set then corresponds to a subset of such points, suggesting a lack of knowledge about the true distribution but restricting it in terms of a set of possible candidates.  

Credal sets are typically assumed to be convex subsets of the class $\mathbb{P}$ of all probability distributions on $\Omega$. Such sets can be specified in different ways, for example in terms of upper and lower bounds on the probabilities $\Prob(A)$ of events $A \subseteq \Omega$. A specifically simple approach (albeit of limited expressivity) is the use of so-called \emph{possibility distributions} and related \emph{possibility measures} \citep{dubo_pt}. A possibility distribution is a mapping $\pi: \, \Omega \longrightarrow [0,1]$, and the associated measure is given by 
\begin{equation}
\Pi: \, 2^\Omega \longrightarrow [0,1], \, A \mapsto \sup_{\omega \in A} \pi(\omega) \, .
\end{equation}
A measure of that kind can be interpreted as an upper bound, and thus defines a set $\mathcal{P}$ of dominated probability distributions:
\begin{equation}
\mathcal{P} \defeq \big\{ \Prob \in \mathbb{P} \given \Prob(A) \leq \Pi(A) \text{ for all } A \subseteq \Omega \big\}
\end{equation}
Formally, a possibility measure on $\Omega$ satisfies $\Pi(\emptyset)=0$, $\Pi(\Omega)=1$, and $\Pi(A \cup B) = \max(\Pi(A), \Pi(B))$ for all $A, B \subseteq \Omega$. Thus, it generalizes the maxitivity (\ref{eq:maxi}) of sets in the sense that $\Pi$ is not (necessarily) an indicator function, i.e., $\Pi(A)$ is in $[0,1]$ and not restricted to $\{ 0, 1 \}$. A related \emph{necessity measure} is defined as $N(A) = 1 - \Pi(\bar{A})$, where $\bar{A} = \Omega \setminus A$. Thus, an event $A$ is plausible insofar as the complement of $A$ is not necessary. Or, stated differently, an event $A$ necessarily occurs if the complement of $A$ is not possible. 

In a sense, possibility theory combines aspects of both set-based and distributional approaches. In fact, a possibility distribution can be seen as both a generalized set (in which elements can have graded degrees of membership) and a non-additive measure. Just like a probability, it allows for expressing graded degrees of belief (instead of merely distinguishing possible from impossible events), but its calculus is maxitive instead of additive\footnote{For this reason, possibility measures can also be defined on non-numerical, purely ordinal structures.}. 

The dual pair of measures $(N, \Pi)$ allows for expressing ignorance in a proper way, mainly because $A$ can be declared plausible without declaring $\bar{A}$ implausible. In particular, $\Pi(A) \equiv 1$ on $2^\Omega \setminus \emptyset$ models complete ignorance: Everything is fully plausible, and hence nothing is necessary ($N(A) = 1 - \Pi(\bar{A}) = 0$ for all $A$). A probability measure, on the other hand, is self-dual in the sense that $\Prob(A) = 1 - \Prob(\bar{A})$. Thus, a probability measure is playing both roles simultaneously, namely the role of the possibility and the role of the necessity measure. 
Therefore, it is more constrained than a representation $(N, \Pi)$. In a sense, probability and possibility distributions can be seen as two extremes on the scale of uncertainty representations\footnote{Strictly speaking, possibilities are not more expressive than probabilities, since possibility distributions cannot model degenerate probability distributions: $\Pi \neq N$ unless $\Pi(\{ \omega^* \}) = 1$ for some $\omega^* \in \Omega$ and $\Pi(\{ \omega \}) = 0$ otherwise.}.

\subsection{Distributions of sets}
\label{sec:dos}

Sets and distributions can also be combined the other way around, namely in terms of distributions of sets. Formalisms based on this idea include the calculus of \emph{random sets} \citep{math_rs,nguy_or78} as well as the Dempster-Shafer theory of evidence \citep{shaf_am}.

In evidence theory, uncertain information is again expressed in terms of a dual pair of measures on $\Omega$, a measure of \emph{plausibility} and a measure of \emph{belief}. Both can be derived from an underlying \emph{mass function} or \emph{basic belief assignment} $m:\, 2^\Omega \longrightarrow [0,1]$, for which $m(\emptyset) = 0$ and $\sum_{B \subseteq \Omega} m(B) = 1$. Obviously, $m$ defines a probability distribution on the subsets of $\Omega$. Thus, instead of a single set or constraint $C$, like in the set-based approach, we are now dealing with a set of such constraints, each one being assigned a weight $m(C)$. Each $C \subseteq \Omega$ such that $m(C) > 0$ is called a \emph{focal element} and represents a single piece of evidence (in favor of $\omega^* \in C$). Assigning masses to subsets $C$ instead of single elements $\omega$ allows for combining randomness and imprecision. 

\begin{figure}
\begin{center}
\includegraphics[scale=0.75]{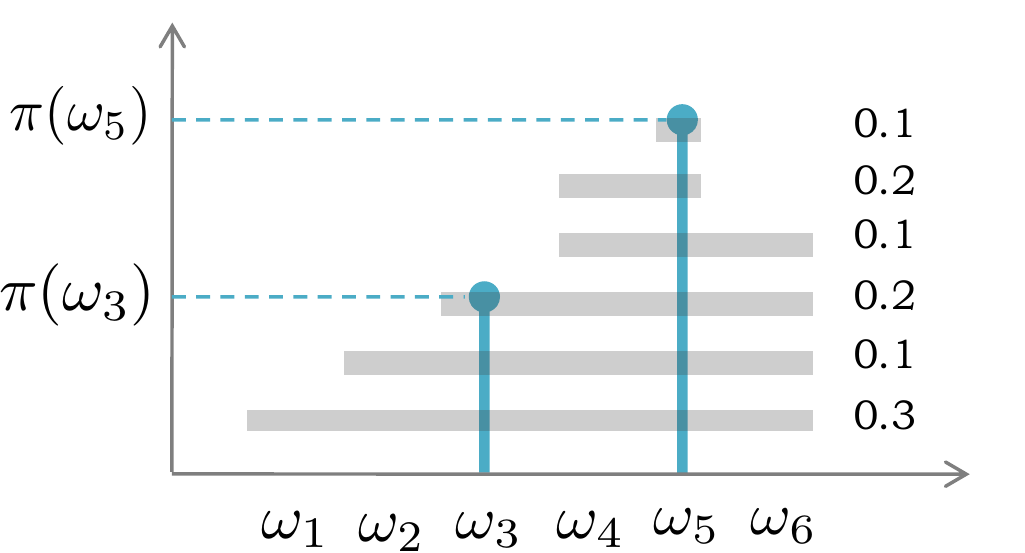}
\caption{Possibility distribution as a contour function of a basic belief assignment $m$ (values assigned to focal sets on the right). In this example, $\pi(\omega_5)=1$, because $\omega_5$ is contained in all focal sets, whereas $\pi(\omega_3)=0.6$.}
\label{fig:contour}
\end{center}
\end{figure}

A plausibility and belief function are derived from a mass function $m$ as follows: 
$$
\on{Pl}(A) \defeq \sum_{B \cap A \neq \emptyset} m(B)  \, ,  \quad
\on{Bel}(A) \defeq \sum_{B \subseteq A } m(B) \, .
$$
Plausibility (or belief) functions indeed generalize both probability and possibility distributions. A probability distribution is obtained in the case where all focal elements are singleton sets. 
A possibility measure is obtained as a special case of a plausibility measure (and, correspondingly, a necessity measure as a special case of a belief measure) for a mass function the focal sets of which are nested, i.e., such that $C_1 \subset C_2 \subset \cdots \subset C_r$. The corresponding possibility distribution is the \emph{contour function} of the plausibility measure: $\pi(\omega) = \on{Pl}(\{ \omega\}) \defeq \sum_{C : \, \omega \in C} m(C)$ for all $\omega \in \Omega$. Thus, $\pi(\omega)$ can be interpreted as the probability that $\omega$ is contained in a subset $C$ that is randomly chosen according to the distribution $m$ (see Fig.\ \ref{fig:contour} for an illustration).

Note that we have obtained a possibility distribution in two ways and for two different interpretations, representing a set of distributions as well as a distribution of sets. One concrete way to define a possibility distribution $\pi$ in a data-driven way, which is specifically interesting in the context of statistical inference, is in terms of the \emph{normalized} or \emph{relative likelihood}. Consider the case where $\omega^*$ is the parameter of a probability distribution, and we are interested in estimating this parameter based on observed data $\mathcal{D}$; in other words, we are interested in identifying the distribution within the family $\{ \Prob_\omega \with \omega \in \Omega \}$ from which the data was generated. The likelihood function is then given by $L(\omega; \mathcal{D}) \defeq \Prob_\omega(\mathcal{D})$, and the normalized likelihood as 
$$
L^{n}(\omega; \mathcal{D}) \defeq \frac{L(\omega; \mathcal{D})}{\sup_{\omega' \in \Omega} L(\omega; \mathcal{D})} \enspace .
$$
This function can be taken as the contour function of a (consonant) plausibility function $\pi$, i.e., $\pi(\omega) = L^{n}(\omega; \mathcal{D})$ for all $\omega \in \Omega$; the focal sets then simply correspond to the confidence intervals that can be extracted from the likelihood function, which are of the form $C_\alpha = \{ \omega \with L^{n}(\omega; \mathcal{D}) \geq \alpha \}$. This is an interesting illustration of the idea of a distribution of sets: A confidence interval can be seen as a constraint, suggesting that the true parameter is located inside that interval. However, a single (deterministic) constraint is not meaningful, since there is a tradeoff between the correctness and the precision of the constraint. Working with a set of constraints---or, say, a flexible constraint---is a viable alternative.

The normalized likelihood was originally introduced by \citet{shaf_am}, and has been justified axiomatically in the context of statistical inference by \citet{wass_bf90}. Further arguments in favor of using the relative likelihood as the contour function of a (consonant) plausibility function are provided by \citet{deno_lb14}, who shows that it can be derived from three basic principles: the likelihood principle, compatibility with Bayes' rule, and the so-called minimal commitment principle. See also \citep{dubo_as97} and \citep{catt_ls05} for a discussion of the normalized likelihood in the context of possibility theory.

\section{Max-min versus sum-product aggregation}
\label{sec:maxmin}

\begin{figure}
\begin{center}
\includegraphics[scale=0.8]{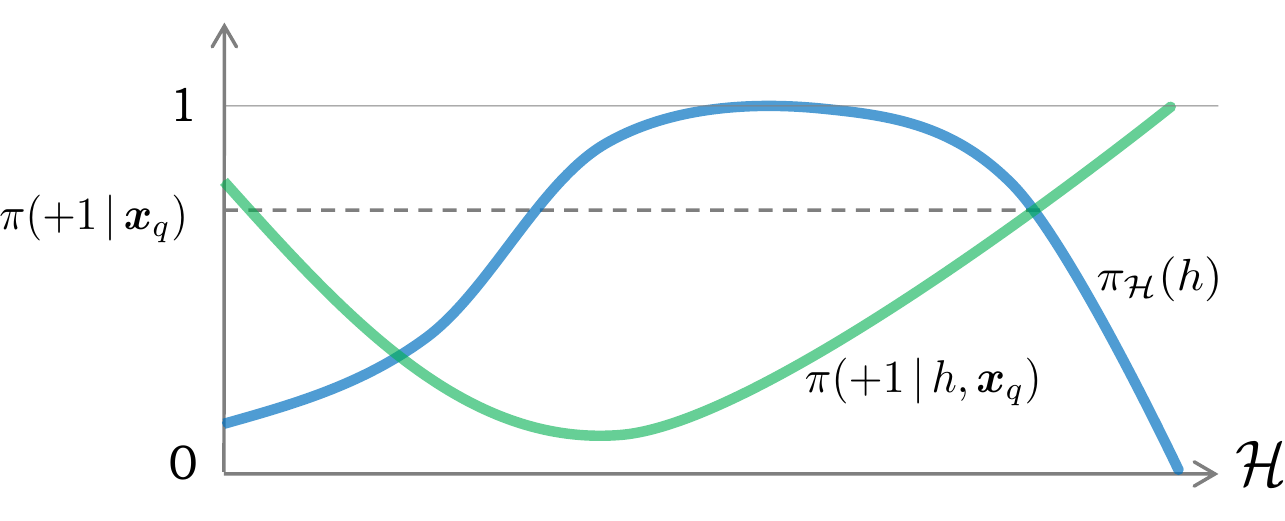} 
\caption{The plausibility $\pi(+1 \given \vec{x}_{q})$ of the positive class is given by the maximum (dashed line) over the pointwise minima of the plausibility of hypotheses $h$ (blue line) and the corresponding plausibility of the positive class given $h$ (green line).}
\label{fig:inf}
\end{center}
\end{figure}

Recall the definition of plausibility degrees $\pi(y \given \vec{x}_{q})$ as introduced in Section \ref{sec:uqnl}.
The computation of $\pi(+1 \given \vec{x}_{q})$ according to (\ref{eq:plaus}) is illustrated in Fig.\ \ref{fig:inf}, where the hypothesis space $\cH$ is shown schematically as one of the axes. In comparison to Bayesian inference (\ref{eq:pd}), two important differences are notable: 
\begin{itemize}
\item
First, evidence of hypotheses is represented in terms of normalized likelihood $\pi_{\cH}(h)$ instead of posterior probabilities $\prob(h \given \cD)$, and support for a class $y$ in terms of $\pi(y \given h, \vec{x}_{q})$ instead of probabilities $h(\vec{x}_{q}) = \prob(y \given \vec{x}_{q})$. 
\item
Second, the ``sum-product aggregation'' in Bayesian inference 
is replaced by a ``max-min aggregation''. 
\end{itemize}
More formally, the meaning of sum-product aggregation is that (\ref{eq:pd}) corresponds to the computation of the standard (Lebesque) integral of the class probability $\prob(y \given \vec{x}_{q})$ with respect to the (posterior) probability distribution $\prob(h \given \cD)$. Here, instead, the definition of $\pi(y \given \vec{x}_{q})$ corresponds to the Sugeno integral \citep{suge_to} of the support $\pi(y \given h, \vec{x}_{q})$ with respect to the possibility measure $\Pi_{\cH}$ induced by the distribution (\ref{eq:noli}) on $\cH$:
\begin{equation}
\pi(y \given \vec{x}_{q}) =  S \!\!\!\!\!\! \int_{\cH} \pi(y \given h, \vec{x}_{q}) \circ \Pi_{\cH}
\end{equation}
In general, given a measurable space $(X,\mathcal{A})$ and an $\mathcal{A}$-measurable function $f:\, X \longrightarrow [0,1]$, the Sugeno integral of $f$ with respect to a monotone measure $g$ (i.e., a measure on $\mathcal{A}$ such that $g(\emptyset) = 0$, $g(X) = 1$, and $g(A) \leq g(B)$ for $A \subseteq B$) is defined as
\begin{equation}
S \!\!\!\!\!\! \int_X f(x) \circ g \defeq \sup_{A \in \mathcal{A}} \left[ \min \left( \min_{x \in A} f(x) , g(A) \right) \right] = \sup_{\alpha \in [0,1]} \Big[ \min \big( \alpha , g(F_\alpha) \big) \Big] \, , 
\end{equation}
where $F_\alpha \defeq \{ x \with f(x) \geq \alpha \}$.

In comparison to sum-product aggregation, max-min aggregation avoids the loss of information due to averaging and is more in line with the ``existential'' aggregation in version space learning. In fact, it can be seen as a graded generalization of (\ref{eq:cbi}). Note that max-min inference requires the two measures $\pi_{\cH}(h)$ and $\pi(+1 \given h, \vec{x}_{q})$ to be commensurable. This is why the normalization of the likelihood according to (\ref{eq:noli}) is important.

Compared to MAP inference (\ref{eq:map}), max-min inference takes more information into account. Indeed, MAP inference only looks at the probability of hypotheses but ignores the probabilities assigned to the classes. In contrast, a class can be considered plausible according to (\ref{eq:plaus}) even if not being strongly supported by the most likely hypothesis $h^{ml}$---this merely requires sufficient support by another hypothesis $h$, which is not much less likely than $h^{ml}$.

\end{appendix}


\begin{thebibliography}{124}
\providecommand{\natexlab}[1]{#1}
\providecommand{\url}[1]{{#1}}
\providecommand{\urlprefix}{URL }
\expandafter\ifx\csname urlstyle\endcsname\relax
  \providecommand{\doi}[1]{DOI~\discretionary{}{}{}#1}\else
  \providecommand{\doi}{DOI~\discretionary{}{}{}\begingroup
  \urlstyle{rm}\Url}\fi
\providecommand{\eprint}[2][]{\url{#2}}

\bibitem[{Abellan and Moral(2000)}]{abel_an00}
Abellan J, Moral S (2000) A non-specificity measure for convex sets of
  probability distributions. International Journal of Uncertainty, Fuzziness
  and Knowledge-Based Systems 8:357--367

\bibitem[{Abellan et~al.(2006)Abellan, Klir, and Moral}]{abel_dt06}
Abellan J, Klir J, Moral S (2006) Disaggregated total uncertainty measure for
  credal sets. International Journal of General Systems 35(1)

\bibitem[{Aggarwal et~al.(2014)Aggarwal, Kong, Gu, Han, and Yu}]{agga_al14}
Aggarwal C, Kong X, Gu Q, Han J, Yu P (2014) Active learning: {A} survey. In:
  Data Classification: Algorithms and Applications, pp. 571--606

\bibitem[{Antonucci et~al.(2012)Antonucci, Corani, and
  Gabaglio}]{antonucci2012}
Antonucci A, Corani G, Gabaglio S (2012) Active learning by the naive credal
  classifier. In: Proceedings of the Sixth European Workshop on Probabilistic
  Graphical Models (PGM), pp. 3--10

\bibitem[{Balasubramanian et~al.(2014)Balasubramanian, Ho, and Vovk}]{bala_cp}
Balasubramanian V, Ho S, Vovk V (eds)  (2014) Conformal Prediction for Reliable
  Machine Learning: Theory, Adaptations and Applications. Morgan Kaufmann

\bibitem[{Barber et~al.(2020)Barber, Candes, Ramdas, and
  Tibshirani}]{barb_tl20}
Barber R, Candes E, Ramdas A, Tibshirani R (2020) The limits of
  distribution-free conditional predictive inference. CoRR abs/1903.04684v2,
  \urlprefix\url{http://arxiv.org/abs/1903.04684v2}

\bibitem[{Bazargami and Mac-Namee(2019)}]{Bazargami2019}
Bazargami M, Mac-Namee B (2019) The elliptical basis function data descriptor
  network: A one-class classification approach for anomaly detection. In:
  European Conference on Machine Learning and Knowledge Discovery in Databases

\bibitem[{Bernardo(1979)}]{bern_rp79}
Bernardo J (1979) Reference posterior distributions for {B}ayesian inference.
  Journal of the Royal Statistical Society, Series B (Methodological)
  41(2):113--147

\bibitem[{Bernardo(2005)}]{bern_ai05}
Bernardo J (2005) An introduction to the imprecise {D}irichlet model for
  multinomial data. International Journal of Approximate Reasoning
  39(2--3):123--150

\bibitem[{Bi and Kwok(2015)}]{Bi2015}
Bi W, Kwok J (2015) Bayes-optimal hierarchical multilabel classification. IEEE
  Transactions on Knowledge and Data Engineering 27:1--1,
  \doi{10.1109/TKDE.2015.2441707}

\bibitem[{Blum and Riedmiller(2013)}]{Blum13}
Blum M, Riedmiller M (2013) Optimization of {G}aussian process hyperparameters
  using {R}prop. In: Proc.\ ESANN, 21st European Symposium on Artificial Neural
  Networks, Bruges, Belgium

\bibitem[{Breiman(2001)}]{brei_rf01}
Breiman L (2001) Random forests. Machine Learning 45(1):5--32

\bibitem[{Cattaneo(2005)}]{catt_ls05}
Cattaneo M (2005) Likelihood-based statistical decisions. In: Proc.\ 4th Int.\
  Symposium on Imprecise Probabilities and their Applications, pp. 107--116

\bibitem[{Chow(1970)}]{chow_oo70}
Chow C (1970) On optimum recognition error and reject tradeoff. {\sc IEEE}
  Transactions on Information Theory IT-16:41--46

\bibitem[{Corani and Zaffalon(2008{\natexlab{a}})}]{cora_lr08}
Corani G, Zaffalon M (2008{\natexlab{a}}) Learning reliable classifiers from
  small or incomplete data sets: {T}he naive credal classifier 2. Journal of
  Machine Learning Research 9:581--621

\bibitem[{Corani and Zaffalon(2008{\natexlab{b}})}]{Corani2008NCC}
Corani G, Zaffalon M (2008{\natexlab{b}}) Learning reliable classifiers from
  small or incomplete data sets: The naive credal classifier 2. Journal of
  Machine Learning Research 9:581--621

\bibitem[{Corani and Zaffalon(2009)}]{Corani2009LNCC}
Corani G, Zaffalon M (2009) Lazy naive credal classifier. In: Proceedings of
  the 1st ACM SIGKDD Workshop on Knowledge Discovery from Uncertain Data, ACM,
  New York, NY, USA, U '09, pp. 30--37, \doi{10.1145/1610555.1610560}

\bibitem[{Cozman(2000)}]{cozm_cn00}
Cozman F (2000) Credal networks. Artificial Intelligence 120(2):199--233

\bibitem[{Csisz\'ar(2008)}]{csis_ac08}
Csisz\'ar I (2008) Axiomatic characterizations of information measures. Entropy
  10:261--273

\bibitem[{{Del Coz} et~al.(2009){Del Coz}, D\'iez, and Bahamonde}]{delc_ln09}
{Del Coz} J, D\'iez J, Bahamonde A (2009) Learning nondeterministic
  classifiers. The Journal of Machine Learning Research 10:2273--2293

\bibitem[{Deng(2014)}]{deng_ge14}
Deng Y (2014) Generalized evidence theory. CoRR abs/404.4801.v1,
  \urlprefix\url{http://arxiv.org/abs/404.4801}

\bibitem[{Denker and LeCun(1991)}]{denk_tn91}
Denker J, LeCun Y (1991) Transforming neural-net output levels to probability
  distributions. In: Proc.\ NIPS, Advances in Neural Information Processing
  Systems

\bibitem[{Denoeux(2014)}]{deno_lb14}
Denoeux T (2014) Likelihood-based belief function: Justification and some
  extensions to low-quality data. International Journal of Approximate
  Reasoning 55(7):1535--1547

\bibitem[{Depeweg et~al.(2018)Depeweg, Hernandez-Lobato, Doshi-Velez, and
  Udluft}]{depe_du18}
Depeweg S, Hernandez-Lobato J, Doshi-Velez F, Udluft S (2018) Decomposition of
  uncertainty in {B}ayesian deep learning for efficient and risk-sensitive
  learning. In: Proc.\ ICML, 35th International Conference on Machine Learning,
  Stockholm, Sweden

\bibitem[{{Der Kiureghian} and Ditlevsen(2009)}]{kiur_ao09}
{Der Kiureghian} A, Ditlevsen O (2009) Aleatory or epistemic? does it matter?
  Structural Safety 31:105--112

\bibitem[{Destercke et~al.(2008)Destercke, Dubois, and Chojnacki}]{dest_up08}
Destercke S, Dubois D, Chojnacki E (2008) Unifying practical uncertainty
  representations: {I}. {G}eneralized p-boxes. International Journal of
  Approximate Reasoning 49:649--663

\bibitem[{DeVries and Taylor(2018)}]{devr_lc18}
DeVries T, Taylor G (2018) Learning confidence for out-of-distribution
  detection in neural networks. CoRR abs/1802.04865,
  \urlprefix\url{http://arxiv.org/abs/1802.04865}

\bibitem[{Dubois(2006)}]{dubo_pt06}
Dubois D (2006) Possibility theory and statistical reasoning. Computational
  Statistics and Data Analysis 51(1):47--69

\bibitem[{Dubois and H\"ullermeier(2007)}]{mpub128}
Dubois D, H\"ullermeier E (2007) Comparing probability measures using
  possibility theory: A notion of relative peakedness. International Journal of
  Approximate Reasoning 45(2):364--385

\bibitem[{Dubois and Prade(1988)}]{dubo_pt}
Dubois D, Prade H (1988) Possibility Theory. Plenum Press

\bibitem[{Dubois et~al.(1996)Dubois, Prade, and Smets}]{dubo_rp96}
Dubois D, Prade H, Smets P (1996) Representing partial ignorance. IEEE
  Transactions on Systems, Man and Cybernetics, Series A 26(3):361--377

\bibitem[{Dubois et~al.(1997)Dubois, Moral, and Prade}]{dubo_as97}
Dubois D, Moral S, Prade H (1997) A semantics for possibility theory based on
  likelihoods. Journal of Mathematical Analysis and Applications
  205(2):359--380

\bibitem[{Endres and Schindelin(2003)}]{endr_an03}
Endres D, Schindelin J (2003) A new metric for probability distributions. IEEE
  Transactions on Information Theory 49(7):1858--1860

\bibitem[{Flach(2017)}]{flac_cc17}
Flach P (2017) Classifier calibration. In: Encyclopedia of Machine Learning and
  Data Mining, Springer, pp. 210--217

\bibitem[{Freitas(2007)}]{Freitas_atutorial}
Freitas AA (2007) A tutorial on hierarchical classification with applications
  in bioinformatics. In: Research and Trends in Data Mining Technologies and
  Applications,, pp. 175--208

\bibitem[{Frieden(2004)}]{frie_sf}
Frieden B (2004) Science from Fisher Information: A Unification. Cambridge
  University Press

\bibitem[{Gal and Ghahramani(2016)}]{gal_bc16}
Gal Y, Ghahramani Z (2016) Bayesian convolutional neural networks with
  {B}ernoulli approximate variational inference. In: Proc.\ of the ICLR
  Workshop Track

\bibitem[{Gama(2012)}]{gama_as12}
Gama J (2012) A survey on learning from data streams: current and future
  trends. Progress in Artificial Intelligence 1(1):45--55

\bibitem[{Gammerman and Vovk(2002)}]{gam_pa02}
Gammerman A, Vovk V (2002) Prediction algorithms and confidence measures based
  on algorithmic randomness theory. Theoretical Computer Science 287:209--217

\bibitem[{Gneiting and Raftery(2005)}]{gnei_sp05}
Gneiting T, Raftery A (2005) Strictly proper scoring rules, prediction, and
  estimation. Tech. Rep. 463R, Department of Statistics, University of
  Washington

\bibitem[{Goodfellow et~al.(2016)Goodfellow, Bengio, and
  Courville}]{Goodfellow2016}
Goodfellow I, Bengio Y, Courville A (2016) Deep Learning. Goodfellow, I.,
  Bengio, Y., Courville, A.: Deep Learning. The MIT Press, Cambridge,
  Massachusetts, London, England

\bibitem[{Graves(2011)}]{grav_pv11}
Graves A (2011) Practical variational inference for neural networks. In: Proc.\
  NIPS, Advances in Neural Information Processing Systems, pp. 2348--2356

\bibitem[{Hartley(1928)}]{hart_to28}
Hartley R (1928) Transmission of information. Bell Syst\ Tech\ Journal
  7(3):535--563

\bibitem[{Hechtlinger et~al.(2019)Hechtlinger, Poczos, and
  Wasserman}]{hech_cd19}
Hechtlinger Y, Poczos B, Wasserman L (2019) Cautious deep learning. CoRR
  abs/1805.09460.v2, \urlprefix\url{http://arxiv.org/abs/1805.09460}

\bibitem[{Hellman(1970)}]{hell_tn70}
Hellman M (1970) The nearest neighbor classification rule with a reject option.
  IEEE Transactions on Systems, Man and Cybernetics SMC-6:179--185

\bibitem[{Hendrycks and Gimpel(2017)}]{hend_ab17}
Hendrycks D, Gimpel K (2017) A baseline for detecting misclassified and
  out-of-distribution examples in neural networks. In: Proc. ICLR, Int.\
  Conference on Learning Representations

\bibitem[{Herbei and Wegkamp(2006)}]{herb_cw06}
Herbei R, Wegkamp M (2006) Classification with reject option. Canadian Journal
  of Statistics 34(4):709--721

\bibitem[{Hora(1996)}]{hora_aa96}
Hora S (1996) Aleatory and epistemic uncertainty in probability elicitation
  with an example from hazardous waste management. Reliability Engineering and
  System Safety 54(2--3):217--223

\bibitem[{H\"uhn and H\"ullermeier(2009)}]{mpub170}
H\"uhn J, H\"ullermeier E (2009) {FR3}: A fuzzy rule learner for inducing
  reliable classifiers. IEEE Transactions on Fuzzy Systems 17(1):138--149

\bibitem[{H\"ullermeier and Brinker(2008)}]{mpub169}
H\"ullermeier E, Brinker K (2008) Learning valued preference structures for
  solving classification problems. Fuzzy Sets and Systems 159(18):2337--2352

\bibitem[{Jeffreys(1946)}]{jeff_46}
Jeffreys H (1946) An invariant form for the prior probability in estimation
  problems. Proceedings of the Royal Society A 186:453--461

\bibitem[{Johansson et~al.(2018)Johansson, L\"ofstr\"om, Sundell, Linusson,
  Gidenstam, and Bostr\"om}]{joha_vp18}
Johansson U, L\"ofstr\"om T, Sundell H, Linusson H, Gidenstam A, Bostr\"om H
  (2018) Venn predictors for well-calibrated probability estimation trees. In:
  Proc.\ COPA, 7th Symposium on Conformal and Probabilistic Prediction and
  Applications, Maastricht, The Netherlands, pp. 3--14

\bibitem[{Jordan et~al.(1999)Jordan, Ghahramani, Jaakkola, and
  Saul}]{jord_ai99}
Jordan M, Ghahramani Z, Jaakkola T, Saul L (1999) An introduction to
  variational methods for graphical models. Machine Learning 37(2):183--233

\bibitem[{Kay(1992)}]{mack_ap92}
Kay DM (1992) A practical {B}ayesian framework for backpropagation networks.
  NeuralComputation 4(3):448--472

\bibitem[{Kendall and Gal(2017)}]{kend_wu17}
Kendall A, Gal Y (2017) What uncertainties do we need in {B}ayesian deep
  learning for computer vision? In: Proc.\ NIPS, Advances in Neural Information
  Processing Systems, pp. 5574--5584

\bibitem[{Khan and Madden(2014)}]{Khan_2014}
Khan SS, Madden MG (2014) One-class classification: taxonomy of study and
  review of techniques. The Knowledge Engineering Review 29(3):345–374,
  \urlprefix\url{http://dx.doi.org/10.1017/S026988891300043X}

\bibitem[{Klement et~al.(2002)Klement, Mesiar, and Pap}]{klem_tn}
Klement E, Mesiar R, Pap E (2002) Triangular Norms. Kluwer Academic Publishers

\bibitem[{Klir(1994)}]{klir_mo94}
Klir G (1994) Measures of uncertainty in the {D}empster-{S}hafer theory of
  evidence. In: Yager R, Fedrizzi M, Kacprzyk J (eds) Advances in the
  Dempster-Shafer theory of evidence, Wiley, New York, pp. 35--49

\bibitem[{Klir and Mariano(1987)}]{klir_ot87}
Klir G, Mariano M (1987) On the uniqueness of possibilistic measure of
  uncertainty and information. Fuzzy Sets and Systems 24(2):197--219

\bibitem[{Kolmogorov(1965)}]{kolm_ta65}
Kolmogorov A (1965) Three approaches to the quantitative definition of
  information. Problems Inform\ Trans 1(1):1--7

\bibitem[{Kruppa et~al.(2014)Kruppa, Liu, Biau, Kohler, K\"onig, Malley, and
  Ziegler}]{krup_pe14}
Kruppa J, Liu Y, Biau G, Kohler M, K\"onig I, Malley J, Ziegler A (2014)
  Probability estimation with machine learning methods for dichotomous and
  multi-category outcome: {T}heory. Biometrical Journal 56(4):534--563

\bibitem[{Kruse et~al.(1991)Kruse, Schwecke, and Heinsohn}]{krus_ua}
Kruse R, Schwecke E, Heinsohn J (1991) Uncertainty and Vagueness in Knowledge
  Based Systems. Springer-Verlag

\bibitem[{Kull and Flach(2014)}]{kull_rm14}
Kull M, Flach P (2014) Reliability maps: {A} tool to enhance probability
  estimates and improve classification accuracy. In: Proc.\ ECML/PKDD, European
  Conference on Machine Learning and Principles and Practice of Knowledge
  Discovery in Databases, Nancy, France, pp. 18--33

\bibitem[{Kull et~al.(2017)Kull, de~Menezes, Filho, and Flach}]{kull_bc17}
Kull M, de~Menezes T, Filho S, Flach P (2017) Beta calibration: a well-founded
  and easily implemented improvement on logistic calibration for binary
  classifiers. In: Proc.\ AISTATS, 20th International Conference on Artificial
  Intelligence and Statistics, Fort Lauderdale, FL, USA, pp. 623--631

\bibitem[{Lakshminarayanan et~al.(2017)Lakshminarayanan, Pritzel, C.\, and
  Blundell}]{laks_sa17}
Lakshminarayanan B, Pritzel A, C.\, Blundell (2017) Simple and scalable
  predictive uncertainty estimation using deep ensembles. In: Proc.\ NeurIPS,
  31st Conference on Neural Information Processing Systems, Long Beach,
  California, USA

\bibitem[{Lambrou et~al.(2011)Lambrou, Papadopoulos, and Gammerman}]{lamb_rc11}
Lambrou A, Papadopoulos H, Gammerman A (2011) Reliable confidence measures for
  medical diagnosis with evolutionary algorithms. IEEE Trans\ on Information
  Technology in Biomedicine 15(1):93--99

\bibitem[{Lassiter(2020)}]{lass_rc20}
Lassiter D (2020) Representing credal imprecision: from sets of measures to
  hierarchical {B}ayesian models. Philosophical Studies Forthcoming

\bibitem[{Lee et~al.(2018{\natexlab{a}})Lee, Bahri, Novak, Schoenholz,
  Pennington, and Sohl-Dickstein}]{lee_dn18}
Lee J, Bahri Y, Novak R, Schoenholz S, Pennington J, Sohl-Dickstein J
  (2018{\natexlab{a}}) Deep neural networks as {G}aussian processes. In: Proc.\
  ICLR, Int.\ Conference on Learning Representations

\bibitem[{Lee et~al.(2018{\natexlab{b}})Lee, Lee, Lee, and Shin}]{lee_as18}
Lee K, Lee K, Lee H, Shin J (2018{\natexlab{b}}) A simple unified framework for
  detecting out-of-distribution samples and adversarial attacks. CoRR
  abs/1807.03888.v2, \urlprefix\url{http://arxiv.org/abs/1807.03888}

\bibitem[{Liang et~al.(2018)Liang, Li, R.\, and Srikant}]{lian_et18}
Liang S, Li Y, R.\, Srikant (2018) Enhancing the reliability of
  out-of-distribution image detection in neural networks. In: Proc. ICLR, Int.\
  Conference on Learning Representations

\bibitem[{Linusson et~al.(2016)Linusson, Johansson, Bostr\"{o}m, and
  L\"ofstr\"om}]{linu_rc16}
Linusson H, Johansson U, Bostr\"{o}m H, L\"ofstr\"om T (2016) Reliable
  confidence predictions using conformal prediction. In: Proc.\ PAKDD, 20th
  Pacific-Asia Conference on Knowledge Discovery and Data Mining, Auckland, New
  Zealand

\bibitem[{Linusson et~al.(2018)Linusson, Johansson, Bostr\"{o}m, and
  L\"ofstr\"om}]{linu_cw18}
Linusson H, Johansson U, Bostr\"{o}m H, L\"ofstr\"om T (2018) Classification
  with reject option using conformal prediction. In: Proc.\ PAKDD, 22nd
  Pacific-Asia Conference on Knowledge Discovery and Data Mining, Melbourne,
  VIC, Australia

\bibitem[{Liu et~al.(2009)Liu, Ting, and hua Zhou}]{Liu_isolationforest}
Liu FT, Ting KM, hua Zhou Z (2009) Isolation forest. In: Proc.\ ICDM 2008,
  Eighth IEEE International Conference on Data Mining, IEEE Computer Society,
  pp. 413--422

\bibitem[{Maau et~al.(2017)Maau, Cozman, Conaty, and de~Campos}]{dera_cs17}
Maau DD, Cozman F, Conaty D, de~Campos CP (2017) Credal sum-product networks.
  In: PMLR: Proceedings of Machine Learning Research (ISIPTA 2017), vol~62, pp.
  205--216

\bibitem[{Malinin and Gales(2018)}]{mali_pu18}
Malinin A, Gales M (2018) Predictive uncertainty estimation via prior networks.
  In: Proc.\ NeurIPS, 32nd Conference on Neural Information Processing Systems,
  Montreal, Canada

\bibitem[{Matheron(1975)}]{math_rs}
Matheron G (1975) Random Sets and Integral Geometry. John Wiley and Sons

\bibitem[{Mitchell(1977)}]{mitc_vs77}
Mitchell T (1977) Version spaces: {A} candidate elimination approach to rule
  learning. In: Proceedings {\sc IJCAI-77}, pp. 305--310

\bibitem[{Mitchell(1980)}]{mitc_tn80}
Mitchell T (1980) The need for biases in learning generalizations. Tech. Rep.
  TR CBM--TR--117, Rutgers University

\bibitem[{Mobiny et~al.(2017)Mobiny, Nguyen, Moulik, Garg, and Wu}]{mobi_dc19}
Mobiny A, Nguyen H, Moulik S, Garg N, Wu C (2017) Drop{C}onnect is effective in
  modeling uncertainty of {B}ayesian networks. CoRR abs/1906.04569,
  \urlprefix\url{http://arxiv.org/abs/1906.04569}

\bibitem[{Neal(2012)}]{neal_bl12}
Neal R (2012) Bayesian learning for neural networks. Springer Science \&
  Business Media 118

\bibitem[{Nguyen(1978)}]{nguy_or78}
Nguyen H (1978) On random sets and belief functions. Journal of Mathematical
  Analysis and Applications 65:531--542

\bibitem[{Nguyen et~al.(2018)Nguyen, Destercke, Masson, and
  H{\"{u}}llermeier}]{mpub385}
Nguyen V, Destercke S, Masson M, H{\"{u}}llermeier E (2018) Reliable
  multi-class classification based on pairwise epistemic and aleatoric
  uncertainty. In: Proceedings {IJCAI} 2018, 27th International Joint
  Conference on Artificial Intelligence, Stockholm, Sweden, pp. 5089--5095

\bibitem[{Nguyen et~al.(2019)Nguyen, Destercke, and H\"ullermeier}]{mpub392}
Nguyen V, Destercke S, H\"ullermeier E (2019) Epistemic uncertainty sampling.
  In: Proc.\ DS 2019, 22nd International Conference on Discovery Science,
  Split, Croatia

\bibitem[{Oh(2017)}]{Oh2017TopKHC}
Oh S (2017) Top-k hierarchical classification. In: {AAAI}, {AAAI} Press, pp.
  2450--2456

\bibitem[{Owhadi et~al.(2012)Owhadi, Sullivan, McKerns, Ortiz, and
  Scovel}]{owha_ou12}
Owhadi H, Sullivan T, McKerns M, Ortiz M, Scovel C (2012) Optimal uncertainty
  quantification. CoRR abs/1009.0679.v3,
  \urlprefix\url{http://arxiv.org/abs/1009.0679}

\bibitem[{Papadopoulos(2008)}]{papa_ic08}
Papadopoulos H (2008) Inductive conformal prediction: Theory and application to
  neural networks. Tools in Artificial Intelligence 18(2):315--330

\bibitem[{Papernot and McDaniel(2018)}]{pape_dk18}
Papernot N, McDaniel P (2018) Deep k-nearest neighbors: {T}owards confident,
  interpretable and robust deep learning. CoRR abs/1803.04765v1,
  \urlprefix\url{http://arxiv.org/abs/1803.04765}

\bibitem[{Perello-Nieto et~al.(2016)Perello-Nieto, Filho, Kull, and
  Flach}]{pere_bc16}
Perello-Nieto M, Filho TS, Kull M, Flach P (2016) Background check: {A} general
  technique to build more reliable and versatile classifiers. In: Proc.\ ICDM,
  International Conference on Data Mining

\bibitem[{Platt(1999)}]{Pla00}
Platt J (1999) Probabilistic outputs for support vector machines and comparison
  to regularized likelihood methods. In: Smola A, Bartlett P, Schoelkopf B,
  Schuurmans D (eds) Advances in Large Margin Classifiers, MIT Press,
  Cambridge, MA, pp. 61--74

\bibitem[{Pukelsheim(2006)}]{puke_od}
Pukelsheim F (2006) Optimal Design of Experiments. SIAM

\bibitem[{Ramaswamy et~al.(2015)Ramaswamy, Tewari, and
  Agarwal}]{Ramaswamy2015CAMCRO}
Ramaswamy HG, Tewari A, Agarwal S (2015) Consistent algorithms for multiclass
  classification with a reject option. CoRR abs/1505.04137

\bibitem[{Rangwala and Naik(2017)}]{Rangwala2017}
Rangwala H, Naik A (2017) Large scale hierarchical classification: foundations,
  algorithms and applications. In: The European Conference on Machine Learning
  and Principles and Practice of Knowledge Discovery in Databases

\bibitem[{R\'enyi(1970)}]{reny_pt}
R\'enyi A (1970) Probability Theory. North-Holland, Amsterdam

\bibitem[{Sato et~al.(2018)Sato, Suzuki, Shindo, and Matsumoto}]{sato_ia18}
Sato M, Suzuki J, Shindo H, Matsumoto Y (2018) Interpretable adversarial
  perturbation in input embedding space for text. In: Proceedings {IJCAI} 2018,
  27th International Joint Conference on Artificial Intelligence, Stockholm,
  Sweden, pp. 4323--4330

\bibitem[{Seeger(2004)}]{seeg_gp04}
Seeger M (2004) Gaussian processes for machine learning. International Journal
  of Neural Systems 14(2):69--104

\bibitem[{Senge et~al.(2014)Senge, B\"osner, Dembczynski, Haasenritter, Hirsch,
  Donner-Banzhoff, and H\"ullermeier}]{mpub272}
Senge R, B\"osner S, Dembczynski K, Haasenritter J, Hirsch O, Donner-Banzhoff
  N, H\"ullermeier E (2014) Reliable classification: Learning classifiers that
  distinguish aleatoric and epistemic uncertainty. Information Sciences
  255:16--29

\bibitem[{Sensoy et~al.(2018)Sensoy, Kaplan, and Kandemir}]{sens_ed18}
Sensoy M, Kaplan L, Kandemir M (2018) Evidential deep learning to quantify
  classification uncertainty. In: Proc.\ NeurIPS, 32nd Conference on Neural
  Information Processing Systems, Montreal, Canada

\bibitem[{Shafer(1976)}]{shaf_am}
Shafer G (1976) A Mathematical Theory of Evidence. Princeton University Press

\bibitem[{Shafer and Vovk(2008)}]{shaf_at08}
Shafer G, Vovk V (2008) A tutorial on conformal prediction. Journal of Machine
  Learning Research pp. 371--421

\bibitem[{Shaker and H\"ullermeier(2020)}]{mpub416}
Shaker M, H\"ullermeier E (2020) Aleatoric and epistemic uncertainty with
  random forests. In: Proc.\ IDA 2020, 18th International Symposium on
  Intelligent Data Analysis, Springer, Konstanz, Germany, LNCS, vol 12080, pp.
  444--456, \doi{10.1007/978-3-030-44584-3\_35}

\bibitem[{Shilkret(1971)}]{shil_mm71}
Shilkret N (1971) Maxitive measure and integration. Nederl\ Akad\ Wetensch\
  Proc\ Ser\ A 74 = Indag\ Math 33:109--116

\bibitem[{Smets and Kennes(1994)}]{smet_tt94}
Smets P, Kennes R (1994) The transferable belief model. Artificial Intelligence
  66:191--234

\bibitem[{Sondhi et~al.(2019)Sondhi, Perry, and Simon}]{feng_sp19}
Sondhi JFA, Perry J, Simon N (2019) Selective prediction-set models with
  coverage guarantees. CoRR abs/1906.05473.v1,
  \urlprefix\url{http://arxiv.org/abs/1906.05473}

\bibitem[{Sourati et~al.(2018)Sourati, Akcakaya, Erdogmus, Leen, and
  Dy}]{sour_ap18}
Sourati J, Akcakaya M, Erdogmus D, Leen T, Dy J (2018) A probabilistic active
  learning algorithm based on {F}isher information ratio. IEEE Transactions on
  Pattern Analysis and Machine Intelligence 40(8)

\bibitem[{Sugeno(1974)}]{suge_to}
Sugeno M (1974) Theory of fuzzy integrals and its application. PhD thesis,
  Tokyo Institute of Technology

\bibitem[{Tan and Le(2019)}]{tan_er19}
Tan M, Le Q (2019) {EfficientNet}: {R}ethinking model scaling for convolutional
  neural networks. In: Proc.\ ICML, 36th Int.\ Conference on Machine Learning,
  Long Beach, California

\bibitem[{Tax and Duin(2004)}]{Tax2004}
Tax DM, Duin RP (2004) Support vector data description. Machine Learning
  54(1):45--66, \doi{10.1023/B:MACH.0000008084.60811.49},
  \urlprefix\url{https://doi.org/10.1023/B:MACH.0000008084.60811.49}

\bibitem[{Vapnik(1998)}]{vapn_sl98}
Vapnik V (1998) Statistical Learning Theory. John Wiley \& Sons

\bibitem[{Varshney(2016)}]{vars_es16}
Varshney K (2016) Engineering safety in machine learning. In: Proc.\ Inf.\
  Theory Appl.\ Workshop, La Jolla, CA

\bibitem[{Varshney and Alemzadeh(2016)}]{vars_ot16}
Varshney K, Alemzadeh H (2016) On the safety of machine learning:
  {C}yber-physical systems, decision sciences, and data products. CoRR
  abs/1610.01256, \urlprefix\url{http://arxiv.org/abs/1610.01256}

\bibitem[{Vovk et~al.(2003)Vovk, Gammerman, and Shafer}]{vovk_al}
Vovk V, Gammerman A, Shafer G (2003) Algorithmic Learning in a Random World.
  Springer-Verlag

\bibitem[{Walley(1991)}]{wall_sr}
Walley P (1991) Statistical Reasoning with Imprecise Probabilities. Chapman and
  Hall

\bibitem[{Wasserman(1990)}]{wass_bf90}
Wasserman L (1990) Belief functions and statistical evidence. The Canadian
  Journal of Statistics 18(3):183--196

\bibitem[{Wolpert(1996)}]{wolp_tl96}
Wolpert D (1996) The lack of a priori distinctions between learning algorithms.
  Neural Computation 8(7):1341--1390

\bibitem[{Yager(1983)}]{yage_ea83}
Yager R (1983) Entropy and specificity in a mathematical theory of evidence.
  International Journal of General Systems 9:249--260

\bibitem[{Yang et~al.(2009)Yang, Wanga, Mi, de~Lin, and Cai}]{yang_ur09}
Yang F, Wanga HZ, Mi H, de~Lin C, Cai WW (2009) Using random forest for
  reliable classification and cost-sensitive learning for medical diagnosis.
  BMC Bioinformatics 10

\bibitem[{Yang et~al.(2017{\natexlab{a}})Yang, Destercke, and
  Masson}]{Yang2017}
Yang G, Destercke S, Masson M (2017{\natexlab{a}}) Cautious classification with
  nested dichotomies and imprecise probabilities. Soft Computing 21:7447--7462

\bibitem[{Yang et~al.(2017{\natexlab{b}})Yang, Destercke, and
  Masson}]{Yang2017b}
Yang G, Destercke S, Masson M (2017{\natexlab{b}}) The costs of indeterminacy:
  How to determine them? IEEE Transactions on Cybernetics 47:4316--4327

\bibitem[{Zadrozny and Elkan(2001)}]{zadr_oc01}
Zadrozny B, Elkan C (2001) Obtaining calibrated probability estimates from
  decision trees and {N}aive {B}ayesian classifiers. In: Proc.\ ICML, Int.\
  Conference on Machine Learning, pp. 609--616

\bibitem[{Zadrozny and Elkan(2002)}]{zadr_tc02}
Zadrozny B, Elkan C (2002) Transforming classifier scores into accurate
  multiclass probability estimates. In: Proc.\ KDD--02, 8th International
  Conference on Knowledge Discovery and Data Mining, Edmonton, Alberta, Canada,
  pp. 694--699

\bibitem[{Zaffalon(2002{\natexlab{a}})}]{zaffalon2002}
Zaffalon M (2002{\natexlab{a}}) The naive credal classifier. Journal of
  Statistical Planning and Inference 105(1):5--21

\bibitem[{Zaffalon(2002{\natexlab{b}})}]{zaff_tn02}
Zaffalon M (2002{\natexlab{b}}) The naive credal classifier. Journal of
  Statistical Planning and Inference 105(1):5--21

\bibitem[{Zaffalon et~al.(2012)Zaffalon, Giorgio, and
  Deratani~Mau{\'a}}]{Zaffalon2012EvaluatingCC}
Zaffalon M, Giorgio C, Deratani~Mau{\'a} D (2012) Evaluating credal classifiers
  by utility-discounted predictive accuracy. Int J Approx Reasoning
  53:1282--1301

\bibitem[{Ziyin et~al.(2019)Ziyin, Wang, Liang, Salakhutdinov, Morency, and
  Ueda}]{ziyin2019deep}
Ziyin L, Wang Z, Liang PP, Salakhutdinov R, Morency LP, Ueda M (2019) Deep
  gamblers: Learning to abstain with portfolio theory. \eprint{1907.00208}

\end{thebibliography}

\end{document}